\let\accentvec\vec
\documentclass[fleqn]{llncs}
\let\spvec\vec
\let\vec\accentvec

\usepackage[latin1]{inputenc}%cesar

\usepackage{multirow}
\usepackage{epsfig}

\newcommand{\ra}{\rightarrow}
\long\def\comment#1{}

\usepackage{amssymb}
\usepackage{amsmath}%para el cases

\usepackage{yhmath} % \widetriangle
\usepackage{tipa}   % many
\usepackage{phonetic}  % \textipa and related symbols
\usepackage{extraipa}  % \subdoublevert
\usepackage{mathabx} % \widecheck  % THIS IS NOT STANDARD. NEEDS TO BE MANUALLY INSTALLED.

\usepackage{stmaryrd} % \Ydown

\usepackage{harpoon} % \overleftharpdown, \overleftharp

\usepackage{mathtools} % \overbracket

\usepackage{accents}  % fabrication of new accents (\accentset, \underaccent)

\newcommand{\textsubcross}[1]{\tipaloweraccent{24}{#1}}

\newcommand{\annl}[1]{\textopencorner{#1}}   %{$\llcorner{}{\mbox{#1}}$}
\newcommand{\annr}[1]{\textcorner{#1}}   %{$\lrcorner{}{\mbox{#1}}$}

\newcommand{\nonl}[1]{$\llcorner{}{\mbox{#1}}$}   %{\textopencorner{#1}}
\newcommand{\nonr}[1]{$\lrcorner{}{\mbox{#1}}$}   %{\textcorner{#1}}
\newcommand{\si}[1]{\textsubcross{#1}}
\newcommand{\st}[1]{\textsyllabic{#1}}
\newcommand{\stst}[1]{\subdoublevert{#1}}
\newcommand{\se}[1]{\textraising{#1}}
\newcommand{\sel}[1]{\textadvancing{#1}}
\newcommand{\ser}[1]{\textretracting{#1}}
\newcommand{\selr}[1]{\sel{}\textsubplus{}}
\newcommand{\serl}[1]{\textsubplus{}\ser{}}
\newcommand{\co}[1]{\textsubring{#1}}
\newcommand{\io}[1]{\d{#1}}
\newcommand{\vo}[1]{\textsubwedge{#1}}
\newcommand{\no}[1]{\textsubcircum{#1}}

\newcommand{\svo}[1]{\textinvsubbridge{#1}}
\newcommand{\sno}[1]{\textsubbridge{#1}}

\newcommand{\hvo}[1]{\b{\textinvsubbridge{#1}}}
\newcommand{\hno}[1]{\b{\textsubbridge{#1}}}

\newcommand{\w}[1]{\textsubw{#1}}
\newcommand{\y}[1]{$\underaccent{\Ydown}{\mbox{#1}}$}
\newcommand{\sch}[1]{$\accentset{\backepsilon}{\mbox{#1\upbar{}}}$}  % package textcomp
\newcommand{\ssch}[1]{$\accentset{\backepsilon}{\mbox{#1}}$}  % package textcomp

\newcommand{\pln}[1]{\textvbaraccent{#1}}
\newcommand{\ppln}[1]{\textvbaraccent{#1}}

\newcommand{\nat}[1]{\~{#1}}
\newcommand{\nnat}[1]{$\widetilde{\mbox{#1}}$}

\newcommand{\brd}[1]{\={#1}}
\newcommand{\bbrd}[1]{$\overline{\mbox{#1}}$}

\newcommand{\rnd}[1]{\r{#1}}
\newcommand{\rrnd}[1]{\r{#1}}

\newcommand{\clr}[1]{\^{#1}}
\newcommand{\cclr}[1]{$\widehat{\mbox{#1}}$}

\newcommand{\opq}[1]{\v{#1}}
\newcommand{\oopq}[1]{$\widecheck{\mbox{#1}}$}

\newcommand{\dip}[1]{$\widetriangle{\mbox{#1}}$}  % Problems
\newcommand{\ddip}[1]{$\widetriangle{\mbox{#1}}$}
\newcommand{\idp}[1]{\`{#1}}
\newcommand{\iidp}[1]{$\overleftharpdown{\mbox{#1}}$}
\newcommand{\udp}[1]{\'{#1}}
\newcommand{\uudp}[1]{$\overleftharp{\mbox{#1}}$}

\newcommand{\iot}[1]{\.{#1}}

\newcommand{\cnt}[1]{\"{#1}}
\newcommand{\ccnt}[1]{\"{#1}}

\newcommand{\red}[1]{\crtilde{#1}}
\newcommand{\rred}[1]{\crtilde{#1}}

\newcommand{\group}[1]{\underline{#1}}
\begin{document}

\title{Annot\st{a}t\iot{e}d \iot{E}nglish}
% http://commons.wikimedia.org/wiki/File:Newton's_tree,_Botanic_Gardens,_Cambridge.JPG

\author{Jos$\acute{\mbox{e}}$ Hern$\acute{\mbox{a}}$ndez-Orallo}
\institute{DSIC, Universitat Polit\`ecnica de Val\`encia, Cam\'{\i} de Vera s/n, 46022 Val\`encia, Spain.
 %{\tt fmartinez@dsic.upv.es}
 \email{jorallo@dsic.upv.es}
}

\date{}
\maketitle

\begin{abstract}

This document presents \annl{}Annot\st{a}t\iot{e}d \iot{E}nglish\annr{}, a system of diacritical symbols which turns English pronunciation into a precise and unambiguous process. The annotations are defined and located in such a way that the original English text is not altered (not even a letter), thus allowing for a consistent reading and learning of the English language with and without annotations. The annotations are based on a set of general rules that makes the frequency of annotations not dramatically high (despite the chaotic orthography of English). This makes the reader easily associate annotations with exceptions, and makes it possible to shape, internalise and consolidate some rules for the English language which otherwise are weakened by the enormous amount of exceptions in English pronunciation.

The advantages of this annotation system are manifold. Any existing text can be annotated without a significant increase in size. This means that we can get an annotated version of any document or book with the same number of pages and fontsize. Since no letter is affected, the text can be perfectly read by a person who does not know the annotation rules, since annotations can be simply ignored. The annotations are based on a set of rules which can be progressively learned and recognised, even in cases where the reader has no access or time to read the rules. This means that a reader can understand most of the annotations after reading a few pages of  \annl{}Annot\st{a}t\iot{e}d \iot{E}nglish\annr{}, and can take advantage from that knowledge for any other annotated document she may read in the future.  

These features pave the way for multiple applications. \annl{}Annot\st{a}t\iot{e}d \iot{E}nglish\annr{} can be used as a tool for teachers and parents when English-speaking children learn English orthography or simply when they start reading. Annotated textbooks, tales, dictionaries and any other material can make English orthography less painful. \annl{}Annot\st{a}t\iot{e}d \iot{E}nglish\annr{} can also be very useful for students of English as a foreign language, since pronunciation is terribly troublesome for them, because they need to learn the meaning, the spelling and the pronunciation for each word they learn, and the pronunciation does not improve at all by reading. This is so because we can only see the spelling and infer the meaning, but books do not provide the correct pronunciation for each word. In fact, incorrect pronunciation (we are not referring here to a bad accent or intonation) persists in people who has lived in an English-speaking country for decades, because the true pronunciation is never shown (unless the IPA transcription is looked up in a dictionary), just heard in different contexts and situations. Finally, \annl{}Annot\st{a}t\iot{e}d \iot{E}nglish\annr{} can also be very practical for any regular user (native or not) of the English language, very especially when facing new (especially technical) words and they have doubts about their pronunciation.

In this document we introduce the rules for annotations and its symbols. This document is not intended for a general audience, and the explanations are focussed to precisely understand how the annotation system works. It is obvious that if the system has to be explained to a final user, a shorter and simpler manual should be issued for that. In any case, for learning \annl{}Annot\st{a}t\iot{e}d \iot{E}nglish\annr{}, the best thing is to take a look at the examples. There is a section in this document with some annotated texts. It is recommended to take a quick look at them before getting into details.
 
{\bf Keywords:} English spelling, English pronunciation, Phonetic rules, Diacritics, Pronunciation without Respelling, Spelling Reform.   

\end{abstract}

\pagenumbering{arabic}

%%%%%%%%%%%%%%%%%%%%%%%%%%%%%%%%%%%%%%%%%%%%%%%%%%%%%%%%%%%%%%%%%%

\section{Introduction}

English is the lingua franca of the world. This means that it is used by many people from different countries and native languages. This also means that it is used with many different proficiency levels. A proper coordination between written and spoken English is only achieved by the few. The main reason for this is the deficient spelling system of the English language. We are not going to discuss on the many causes (the {\em Great Vowel Shift}, the lack of an Academy of the English Language acting as regulator, the nature of English as an amalgamation of words from many different languages). The issue is that it is a fact that English spelling is problematic. But it is also a fact that any spelling reform (shallow or deep) has been stubbornly doomed to failure.

Spanning from several centuries ago, reforms can be distinguished into those which focus on spelling and those which focus on pronunciation.
Many languages in the world based on the Roman alphabet produce one single pronunciation from a given spelling (Spanish, German, Italian, and French, are examples of this), but almost no language produce one single spelling given a pronunciation (Italian is perhaps the closest one to that goal among the major European languages).
Additionally, there have been minor reform proposals (as those undertook by American English in the nineteenth century), major ones (Webster proposed to go further beyond) and drastic reforms. There are also several organisations for English spelling reform\footnote{One of them is the spelling society:  http://www.spellingsociety.org/aboutsss/position.php, which does not currently advocate any specific solution, but supports any study and progress on this issue.}.

If we take a look at some of the most recent proposals, we can get an idea of why they failed. For instance, Basic Roman Spelling \cite{Ivanov2004} preserves the alphabet (``{\em Luk thru dha senta av dha muun hwen it iz bluu}'' $\rightarrow$ ``look through the center of the moon when it is blue"), while it oversimplifies pronunciation, in order to supposedly give one-to-one correspondence between spelling and pronunciation. Similarly, Interspel \cite{Yule1986} preserves the alphabet but it is closer to the original English spelling: ``{\em English will luse its position as the wurld's lingua franca, which it inherited for politicl and comercial reasns. The rest of the wurld has problems lerning two English languajes, the spoken and the written, when it shud be posibl to lern the one from the other}''. The one-to-one correspondence is lost, but pronunciation can be better  determined (but not unambiguously). The first feeling after reading this is that a lot of words should be corrected. We have devoted a lot of effort to learn the correct spelling of each word in order to mangle everything again. Other approaches try to avoid this confusion by using extended languages, such as John Malone's Unifon \cite{Ratz1966}, but going out from the Latin alphabet is a strong bet. Additionally, many of these reforms do not produce good pronunciations in some cases. For instance, most of them ignore or give a secondary role to stress, which is one of the main issues for a good pronunciation of English vowels, and to distinguish between many pairs (verb/noun), such as `a present' and `to present' (which are still different because of the vowel sounds), as well as `an imprint' and `to imprint' (which have exactly the same vowel pronunciation, only differing on the stress).

There are several problems about these proposals. They try to change the language spelling. This requires a broad consensus and an academy or working group to undertake them. Even in languages where minor changes have been made (e.g. Portuguese and German undertook minor reforms in the last decades), this has been difficult and has required almost a generation to settle. Any reform involves that old texts become obsolete and a lot of retraining is required, as well as reprinting textbooks, documents and almost every written thing. But even ignoring all these major problems, there are many others, such as the loss of word etymology. This loss of word origin is not a picky thing for linguists; most people would not care about `solid' being spelled as `sollid', even if it is not etymological, because it comes from Latin `solidus'. The problem is that any reform which would not preserve some basic etymology would make the learning of some words difficult to native speakers and especially foreigners (e.g. `fone' is more difficult to recognise than `phone', `Rusha' than `Russia', `Krischan' than `Christian', `memmeri' than `memory', `langwidge' than `language', `Arkeeopteriks' than `Archaeopteryx', `owshan' than `ocean'). This would obscure the relation among words with the same root. For instance, `sine' (for `sign') would make its relation with `signal' much less evident. In addition, thousands of homographs (`to , `two' and `too'; `male' and `mail'; `eye', `I'; `by', `buy' and `bye'; etc) would be created. Another problem would be the incorporation of new technical (e.g. biological and chemical) words. For instance, `Arkeeopteriks' and  `Akwa' (for `Aqua') would be very awkward. And, finally, one above all: we like the language as it is and we want to learn and work with one language, not with two languages (the old and the new).
And finally, any spelling reform would have a dramatic impact in computational linguistics and thousands of computer applications which process English automatically, such as editors, spell checkers, searching engines, etc.
All these reasons are much more important that the reluctance of users, the lack of consensus, the non-existence of an Academy of the English Language, or the {\em Great Vowel Shift}. English spelling reforms have not succeeded simply because it is not true that everything would be beneficial once the transformation would be completed.

So, is there any way to systematically know how a word in English is pronounced {\em without} a spelling reform? Yes, of course. This is called the {\em phonetic transcription} of a word. For instance, we can look up any word in the dictionary and see its IPA (International Phonetic Alphabet) transcription, where it says that 
`explanation' is pronounced \textipa{[""Ekspl@"neIS@n]}. Therefore, the English pronunciation problem seems to be solved\footnote{In fact, we could write IPA instead (the most drastic and accurate spelling reform one can imagine), and we would have a one-to-one correspondence between spelling and pronunciation.}. 
However, when we learn the language or we read a book in English, we do not check the phonetics for every word. Computer technology makes it possible to add an extra line to each line of text with the IPA transcription. However, this would double book sizes and would make readers switch repeatedly from English to IPA and vice-versa, as having a book in two languages or with subtitles.  

Instead of that, our proposal is to keep English unaltered, but to annotate its pronunciation over the very word, not apart from it. 
We can, of course, find some precedents. For instance, many editions of the Webster's Elementary Spellingbook (e.g. the 1836's edition \cite{webster1836american}) contained a set of diacriticals which helped to learn the pronunciation of each word. The words were not respelled. Instead, ``marked letters'' were used, as in `s\.{o}n' or `{\=g}et'. Some other dictionaries also used this philosophy until the mid XXth century, when respelling became more popular (and the pronunciation of `son' was explained by respelling it like `sun', and `pleasure' was respelled as `plesher'). Of course, things changed dramatically when the IPA popularised, and even though the respelled pronunciation is still common in dictionaries (most especially in America), the idea of pronunciation without respelling (the ``marked letters') faded away. Apart from (or out of) English language textbooks and dictionaries, none of these systems have been used profusely, because the systems have not been designed to mark (or annotate) a whole text, but to clarify the pronunciation of some difficult words. In fact, it is still seen in some editions of the Bible, since most of their readers have difficulties on the pronunciation (and meaning) of words which are not frequent nowadays.
There are also some children books which colour some letters to show their pronunciation, based on the ambitious (and once popular) respelling system known as the {\em Initial Teaching Alphabet} \cite{ITA}.

However, these approaches (including the ones used in old dictionaries) were incomplete (not all possible sounds were covered) and unsystematic (not intended to be used for every word) in such a way that an entire document can be annotated. The reason-why is that these approaches had almost no by-default rules, and using them for a whole document would have rendered it full of annotations, at a ratio of many annotations per word (something similar to what is shown in Table \ref{tab:tradeoff} (left)). For instance, there were marks for words such as `cat' and `cell', even though they look pretty regular to any English speaker. Additionally, they were not considering a possible automation of the process, since these diacritical systems were conceived before computers spread over. And, finally, they were created at a moment when English phonology was not so well studied as it is today, and were typically focussed on one single dialect of English.

Our proposal is not just a modernisation of these early approaches to pronunciation without respelling. It is different in philosophy and basic principles. The basic idea is to develop a set of symbols (marks or annotations, whatever we want to call them), but also a set of {\em rules}, in such a way that we can annotate some words which do not follow the rules, but we keep the parts that are regular free from annotations. With this, we do have the pronunciation, we also have the original English word, and we also see the parts which have a regular pronunciation and which parts do not. The idea is to devise a transformation code, in such a way that using these rules and the annotations we can extract the right pronunciation from a word spelling. The only caveat is that, as we will see, even with a well-established set of annotations and rules, an English word can be annotated in several different ways to get the same pronunciation. The solution, as we will see, is to use the way which minimises the annotations (or some measure of the annotation cost).

As a result, \annl{}Annot\st{a}t\iot{e}d \iot{E}nglish\annr{} is:

\begin{itemize} 
\item A system of diacritical annotations which are placed around (above and below) the words to indicate its true pronunciation. So it follows the old motto ``pronunciation without respelling".
\item A set of rules which are designed to reduce the proportion of required annotations. This keeps the text as clear as possible but, very importantly, helps to understand the spelling rules in English. A reader can focus on the general rules since exceptions are just annotated.
\item While reading \annl{}Annot\st{a}t\iot{e}d \iot{E}nglish\annr{}, you {\em are} reading English. Annotations can be ignored by a proficient reader.
\item Every new word which is learned when reading an annotated book comes with both its English spelling and its pronunciation. When a person reads in \annl{}Annot\st{a}t\iot{e}d \iot{E}nglish\annr{}, she can improve and learn both spelling and pronunciation. Both things at the same time and in the same space.
\item No rule can depend on the meaning of the word, its position or its kind (plural/singular, word origin). For instance, annotations must be necessary different for homographs which are not homophones, as in `a present' and `to present', or `to read' and `have read'. 
\item Because of the previous item, the pronunciation (interpretation) process can of course be automated, so an annotated document can be pronounced automatically by a robotic or a computer system\footnote{There are many voice synthesisers which have disambiguation tools so `I read now' and `I read yesterday' are well pronounced. However, these synthesisers fail in other more complex cases, so being amusing for some users but crossing some others (e.g. blind people). This would not happen for an annotated text since disambiguation would not be needed here. Apart from that, there are intonation issues in voice synthesisers we are not concerned here.}. 
\item The annotation (coding) process can also be automated\footnote{Disambiguation tools would be needed here, but note that a text have to be annotated once but then it can be read thousands of times.}, so electronic documents and books can be converted from plain (traditional) English into \annl{}Annot\st{a}t\iot{e}d \iot{E}nglish\annr{}, to be printed on paper or read over an e-book or a computer screen.
\item It tries to be compatible with most English dialects, while also allowing different annotations in case a word is pronounced differently.
\item Diacritical symbols try to reuse many symbols which are common in other languages or even in English. Additionally, although we introduce diacritics which span over more than one symbol, there is always an equivalence to one-character annotations. This eases typographic issues when editing and printing annotated documents.
\end{itemize}

\noindent Regarding the accents, the annotations must be able to distinguish between sounds which are different in at least one dialect. For instance, the derived pronunciations of `cot' and `caught' must be different, even though some speakers do not distinguish them. Other speakers may merge `caught' and `court', and similarly we may find homophones in some dialects which do not take place for others, such as in father/farther, formally/formerly, tune/toon, Lenin/Lennon, etc. Rhotic accents and their way to pronounce the vowels before an `r' and the `r' itself suffer from many mergers. Our system can distinguish between `merry', `marry', `Mary', `fairy' and `merely', even though some speakers can pronounce `merry', `Mary' and `marry' in the same way. It also distinguishes between `hurry' and `furry'. This may give the impression that we favour Received Pronunciation (RP, the standard pronunciation for English in England) over General American (GA, the standard pronunciation for American English). However, this is just a consequence of RP having more phonemes than GA. In some other cases, the annotation must give preference for one dialect, and we have to decide whether we annotate `glance' with the `a' in `father' or with the `a' in `cat'.

We do not distinguish, however, between strong and weak forms for words such as `the', `a', `to', `and', `of', etc. Even though, in some cases, the weak form is preferred (especially for `the' and `a'), we assume every word is independent and we annotate it as if it were pronounced independently. Consequently, we annotate the word `you' as \annl{}y\si{o}u\annr{} (\textipa{[ju:]}), even though it can be pronounced as \textipa{[j@]}. Nonetheless, our annotation system typically changes the default pronunciation of vowels depending on whether they are stressed or not. Consequently, our system is implicitly consistent with the strong and weak forms for many words. For some very common words such as the word `the', we have incorporated a rule (by the way, a rule that any English speaker knows).

Once explained what our proposal is, it is still convenient to state what \annl{}Annot\st{a}t\iot{e}d \iot{E}nglish\annr{} is {\em not}:

\begin{itemize} 
\item A spelling reform. English spelling is left untouched.
\item A phonetic transcription. Most of the spelling is reused to make up the final picture. Many words do not require annotations.
\item A set of rules for computer applications. Even though an important factor for the success of this proposal is to provide computer programs which automatically annotate books and documents, this will be just a tool (an important one), but not the core of the project.  \annl{}Annot\st{a}t\iot{e}d \iot{E}nglish\annr{} can be used to teach and learn English pronunciation and spelling, without the computer annotator.
\end{itemize}

\noindent We have mentioned the advantages and applications of \annl{}Annot\st{a}t\iot{e}d \iot{E}nglish\annr{} in the abstract. Perhaps the most direct application is for non-native speakers, who struggle to learn the correct pronunciation of many words they have  read many times but only occasionally or carelessly heard. In fact, native speakers have to {\em suffer} the terrible pronunciation of beginners, which are unable to coordinate their written English with their poor spoken English (or vice versa). A great proportion of members in international organisations, conferences, multinational companies, etc., come from non-English speaking countries. Even though they have a good proficiency of written English (in some cases better than many native speakers), their pronunciation is not only incorrect, but full of particularities from their own languages, which makes their cross understanding more difficult (even for the native speakers).

For native speakers, this may be useful as well. According to several studies (e.g. \cite{Thorstad1991}), children take more time to read in English than those in other languages such as Spanish, German or Italian. Explicitly connecting spelling with pronunciations through a set of rules and symbols could be very useful for children\footnote{Phonetic proposals to reading (which were very popular in the 1960s), such as the Initial Teaching Alphabet \cite{ITA} were unsuccessful partly because the system should be abandoned and switched to `real' English at some stage of the process.}. Furthermore, even after years of school, (adult) native speakers still do have difficulties with foreign words and technical words, and some words are pronounced differently depending on their region or their social level. The spelling system in English does not put many constraints on how a word should be pronounced, given its spelling\footnote{In fact, it can be argued whether English spelling is something in between ideographic writing systems and phonetic writing system, with words in English being mere mnemonics.}. This has produced that many new, commercial, foreign and technical words have different pronunciations. For instance, words such as `Linux', `Lego'{$^\copyright$}, `genre', `evolution', `Neanderthal', `amino' have several pronunciations.

One of the key issues in this proposal has been to find a compromise between rules and required annotations. Of course, rules have to be learned (or inferred from annotated text). Thus, in order to avoid this effort, we could just explain the meaning of each annotation and use them profusely (because there would be no rules at all to rely on in order to save up some annotations). This would imply that most words would require several annotations. In fact, every vowel and many consonants would require the annotation at every position. Alternatively, we could make an exhaustive set of rules in order to avoid as many annotations as possible. The problem of this approach is that we may finally make up a much too complex set of rules. This would make the reading of an annotated text a difficult task, apart from the time which would be required to learn the rules. 

The idea is to find a trade-off between the two extremes. We will develop a set of rules, as general as possible, many of which would be recognised by any English speaker, in such a way that they can be learned quickly. This short set of rules could reduce the number of annotations significantly, without causing ambiguities or confusions in the interpretation process.

Table \ref{tab:tradeoff} shows the difference between an annotated text without rules (left) and an annotated text with rules (right). We have to say that it is not only a question of economy and clarity. The version with rules helps the reader assimilate the pronunciation and spelling rules of English, so helping to bridge the gap between written and spoken English.

\begin{table}
\begin{center}
{\small
\begin{tabular}{|| p{6cm} || p{6cm}    ||}
\hline
\bf{Annotated text without rules} & \bf{Annotated text with rules} \\ \hline \hline

\pln{I}t w\opq{a}\vo{s} \vo{th}\cnt{e} b\pln{e}st \pln{o}\vo{f} t\nat{\i}m\si{e}\vo{s}, \pln{\i}t w\opq{a}\vo{s} \vo{th}\cnt{e} w\cnt{o}rst \pln{o}\vo{f} t\nat{\i}m\si{e}\vo{s}, \pln{\i}t w\opq{a}\vo{s} \vo{th}\nat{e} \nat{a}\hvo{g}\si{e} o\vo{f} w\st{\pln{\i}}\vo{s}d\cnt{o}m, 
 \pln{\i}t w\opq{a}\vo{s} \vo{th}\cnt{e} \nat{a}\hvo{g}\si{e} o\vo{f} f\st{\rnd{o}}\si{o}lishn\cnt{e}ss, \pln{\i}t w\opq{a}\vo{s}
 \vo{th}\nat{e} \st{\nat{e}}p\cnt{o}c\si{h} \pln{o}\vo{f} b\iot{e}l\bbrd{\st{\i}e}f, \pln{\i}t w\opq{a}\vo{s}
 \vo{th}\nat{e} \st{\nat{e}}p\cnt{o}c\si{h} \pln{o}\vo{f} \stst{i}ncr\cnt{e}d\st{\nat{u}}lity, \pln{\i}t w\opq{a}\vo{s}
 \vo{th}\cnt{e} s\nnat{\st{e}a}\vo{s}\cnt{o}n \pln{o}\vo{f}  L\nat{\i}\si{g}\si{h}t,  \pln{\i}t w\opq{a}\vo{s}
 \vo{th}\cnt{e} s\nnat{\st{e}a}\vo{s}\cnt{o}n \pln{o}\vo{f} 
 D\brd{a}rkness, \pln{\i}t w\opq{a}\vo{s}
 \vo{th}\cnt{e} spr\pln{\i}\co{n}\si{g} \pln{o}\vo{f}  h\nat{o}p\si{e}, 
\pln{\i}t w\opq{a}\vo{s}
 \vo{th}\cnt{e}  w\st{\pln{\i}}nt\cnt{e}r \pln{o}\vo{f} d\iot{e}sp\st{\nat{a}}\si{i}r, ...
&
It w\opq{a}s the best o\vo{f} times, it w\opq{a}s  the w\cnt{o}rst o\vo{f} times, it w\opq{a}s  the age o\vo{f} wisdom, it w\opq{a}s the age o\vo{f} foolishness, it w\opq{a}s  the epoc\si{h} o\vo{f} b\iot{e}l\bbrd{\st{\i}e}f, it w\opq{a}s the  epoc\si{h} o\vo{f} incred\st{u}lity, it w\opq{a}s  the season o\vo{f} Light, it w\opq{a}s the season o\vo{f} Darkness, it w\opq{a}s  the spring o\vo{f} hope, 
it w\opq{a}s  the winter o\vo{f} d\iot{e}sp\st{a}ir, ... \\ \hline \hline
\end{tabular}
} % small
\vspace{5mm}
\caption{Difference between an annotated text without rules (left) and our current proposal using rules (right)}
\label{tab:tradeoff} 
\end{center}
\end{table}

\begin{figure}
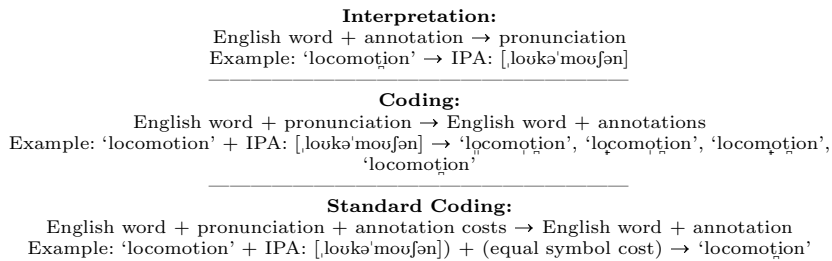

	\centering
		{%\tiny
		  \scriptsize
		  %\footnotesize
		{\bf Interpretation:} 		\\
		English word + annotation  $\rightarrow$ pronunciation \\ 
		Example: `locomo\sno{ti}on'  $\rightarrow$ IPA: \textipa{[""loUk@"moUS@n]} \\
		------------------------------------------------------------ \\
		{\bf Coding:}      \\
		 English word  + pronunciation  $\rightarrow$ English word + annotations \\
		Example: `locomotion' + IPA: \textipa{[""loUk@"moUS@n]}  $\rightarrow$ `l\stst{o}com\st{o}\sno{ti}on', `lo\serl{}com\st{o}\sno{ti}on', `locom\serl{}o\sno{ti}on', `locomo\sno{ti}on' \\
		------------------------------------------------------------ \\
		{\bf Standard Coding:} \\
		 English word  + pronunciation + annotation costs $\rightarrow$ English word + annotation  \\
		Example: `locomotion' +  IPA: \textipa{[""loUk@"moUS@n]}) + (equal symbol cost) $\rightarrow$ `locomo\sno{ti}on'  \\
		}
		\caption{Inputs and outputs when interpreting and annotating texts.}
\label{fig:inteannot} 
\end{figure}

It is important to distinguish between the process of reading (interpreting) an annotated word and the process of annotating (coding) a word, as well as identifying their inputs and outputs, as shown in Figure \ref{fig:inteannot}.
Reading an annotated word can be done very easily and only requires a quick look at the rules and some practice with a few pages of text. Interpreting an annotated text takes the original English word and its annotations  as inputs and produces its pronunciation. This process in unambiguous and has only one possible outcome.
Conversely, coding a word is more cumbersome, because several annotations are possible. The good thing is that this task is not performed by the reader. Given an English word and a pronunciation, an annotator (a person or a computer program) has to choose between the possible annotations. In general, there are not many options, so given a word and its pronunciation, we can establish some criteria or set of scores in order to choose the best (or standard) annotation for the word. In fact, this process can be performed automatically by a computer program if we have access to the IPA transcriptions of all the English words appearing in a text.

This document is organised as follows. In section \ref{sec:symbols} we present the set of symbols we will use and their basic meaning. We also include some IPA notation and briefly explain how the annotation symbols look like and how they can be produced in \LaTeX{} (you can ignore the \LaTeX{} commands if you are not going to use \LaTeX{} as a text editor). In section \ref{sec:rules} we present the rules to follow the annotation, i.e. the interpretation process. In section \ref{sec:annotate} we discuss the problem of annotating a document. Section \ref{sec:examples} includes some annotated texts, which are very helpful to see how \annl{}Annot\st{a}t\iot{e}d \iot{E}nglish\annr{} works in practice. In section \ref{sec:conclusion} we conclude the paper with a discussion about some preliminary statistics, whether the compromise between rules/annotations is correct, dialects, typography, the use of diacritical annotations, and, of course, future work. Appendices include alternative rules we finally did not implement, and some \LaTeX{} stuff.

\section{Symbols}\label{sec:symbols}

Here we introduce the symbols we will use, their meaning and how to use them in \LaTeX{}.

\subsection{Phonetic Symbols}

For phonetic transcriptions, we follow the conventions from the International Phonetics Alphabet (IPA). We will use the symbols shown in table \ref{tab:ipa} (consonants and semiconsonants) and \ref{tab:ipa2} (vowels and diphthongs). The only particular notation which differs from IPA is that
we will use the notation \textipa{[K]} to represent either \textipa{[\*r]} or  \textipa{[@(\*r)]}. This is useful to generally distinguishing between `flower' and `flour', since the first is \textipa{[flaU@\*r]} in GA and \textipa{[flaU@]} in RP, while the second is \textipa{[flaU\*r]} in GA and \textipa{[flaU@]} in RP. With our notation, `flower' is \textipa{[flaU@\*r]} and `flour' \textipa{[flaUK]}. We use \textipa{[\*r]} to represent the special sound of the `r' in English in contrast to other languages. In general, if we are only treating with English, we can use simply \textipa{[r]}.

\begin{table}
\begin{center}
\begin{tabular}{|| c | c | c ||}
\hline
\bf{IPA}&\bf{Example / Explanation} & \LaTeX{} command ({\tt \textbackslash{}textipa\{[...]\}})\\
\hline\hline
\textipa{[p]}	&	{\em p}ill & {\tt [p]} \\\hline
\textipa{[b]}	&	{\em b}it & {\tt [b]} \\\hline
\textipa{[t]}	&	{\em t}en & [t] \\\hline
\textipa{[d]}	&	{\em d}ive & [d] \\\hline
\textipa{[g]}	&	{\em g}o & [g] \\\hline
\textipa{[k]}	&	{\em c}at & [k] \\\hline
\textipa{[m]}	&	{\em m}ine & [m] \\\hline
\textipa{[n]}	&	{\em n}ice & [n] \\\hline
\textipa{[N]}	&	si{\em ng} & [N] \\\hline
\textipa{[r] / [\*r]}	&	{\em r}ay & [r] / [\textbackslash{}*r] \\\hline
\textipa{[f]}	&	{\em f}ine & [f] \\\hline
\textipa{[v]}	&	{\em v}ine & [v] \\\hline
\textipa{[T]}	&	{\em th}in & [T] \\\hline
\textipa{[D]}	&	{\em th}em & [D] \\\hline
\textipa{[s]}	&	{\em s}ong & [s] \\\hline
\textipa{[z]}	&	{\em z}oo & [z] \\\hline
\textipa{[S]}	&	{\em sh}ine & [S] \\\hline
\textipa{[Z]}	&	lei{\em s}ure & [Z] \\\hline
\textipa{[tS]}	&	{\em ch}in & [tS] \\\hline
\textipa{[dZ]}	&	{\em m}agic & [dZ] \\\hline
\textipa{[h]}	&	{\em h}e & [h]\\\hline
\textipa{[x]}	&	lo{\em ch} & [x] \\\hline
\textipa{[j]}	&	{\em y}es & [j] \\\hline
\textipa{[l]}	&	{\em l}ine & [l] \\\hline
\textipa{[w]}	&	{\em w}e & [w] \\\hline
\textipa{[\*w]} & {\em wh}en (some dialects) & [\textbackslash{}*w] \\\hline 
\textipa{[R]}	&	compu{\em t}er (some dialects) & [R] \\\hline

\end{tabular}
\vspace{5mm}
\caption{IPA phonetic symbols (consonants and semiconsonants). Adapted from {\tt http://en.wikipedia.org/wiki/English\_orthography}}
\label{tab:ipa} 
\end{center}
\end{table}

\begin{table}
\begin{center}
\begin{tabular}{|| c | c | c ||}
\hline
\bf{IPA}&\bf{Example / Explanation} & \LaTeX{} command {\tt \textbackslash{}textipa\{[...]\}})\\
\hline\hline

\textipa{["]}& Start of stressed syllable. & ["] \\\hline 
\textipa{[""]} & Start of secondary stressed syllable. & [""] \\\hline

\textipa{[i:]}	&	b{\em e} & [i:] \\\hline
\textipa{[I]}	&	b{\em i}t & [I] \\\hline
\textipa{[u:]}	&	t{\em oo}l & [u:] \\\hline
\textipa{[U]}	&	l{\em oo}k & [U] \\\hline
\textipa{[eI]}	&	r{\em a}te & [eI] \\\hline
\textipa{[@]}	&	anth{\em e}m & [@] \\\hline
\textipa{[oU]}	&	b{\em o}ne & [oU] \\\hline
\textipa{[E]}	&	m{\em e}t & [E]  \\\hline
\textipa{[\ae]}	&	h{\em a}nd & [\textbackslash{}ae] \\\hline

\textipa{[2]}	&	s{\em u}n & [2]  \\\hline
\textipa{[O:]}	&	j{\em aw} & [O:] \\\hline
\textipa{[6] }	&	l{\em o}ck & [6] \\\hline
\textipa{[A:]}	&	f{\em a}ther & [A:] \\\hline
\textipa{[aI]}	&	f{\em i}ne & [aI] \\\hline
\textipa{[OI]}	&	b{\em o}y & [OI] \\\hline
\textipa{[aU]}	&	{\em ou}t & [aU] \\\hline
\textipa{[ju:]}	&	m{\em u}se & [ju:] \\\hline
\textipa{[A\*r]}	&	c{\em ar} & [Ar] \\\hline
\textipa{[3\*r]}	&	f{\em er}n & [3r] \\\hline
\textipa{[O\*r]}	&	d{\em oor} & [Or] \\\hline
\textipa{[E(@)\*r] / [EK]}	&	h{\em air} & [E(@)\*r] / [EK]\\\hline
\textipa{[I(@)\*r] / [IK]}	&	h{\em er}e & [I(@)\*r] / [IK] \\\hline
\textipa{[aI(@)\*r] / [aIK]}	&	f{\em ir}e & [aI(@)\*r] / [aIK] \\\hline
\textipa{[U(@)\*r] / [UK]}	&	m{\em oo}r & [U(@)\*r] / [UK] \\\hline
\textipa{[jU(@)\*r] / [jUK]}	&	p{\em u}re & [jU(@)\*r] / [jUK] \\\hline

\textipa{[j@]}  & fail{\em u}re (reduced unstressed u ) & [j@] \\\hline
\textipa{[8]}	&	es{\em o}teric (reduced unstressed o) & [8] \\\hline
\textipa{[0]}	&	beautif{\em u}l (reduced unstressed u) & [0] \\\hline
\textipa{[1]}	&	lent{\em i}l (reduced unstressed i) & [1] \\\hline

\end{tabular}
\vspace{5mm}
\caption{IPA phonetic symbols for vowels and diphthongs. Adapted from {\tt http://en.wikipedia.org/wiki/English\_orthography}}
\label{tab:ipa2} 
\end{center}
\end{table}

Using IPA, we can describe the pronunciation of a word as shown with the following example:

{\em explanation} $\rightarrow$ \textipa{[""Ekspl@"neIS@n]}  

\noindent Some words have different pronunciations depending on the accent. The previous table is not exactly phonetic but pseudophonetic (they try to generalise several English accents), and it is custom for many English dictionaries, which only need to include different pronunciations when they are not inferrable from the accent (e.g, the two different pronunciations for the word `either').

We incorporate a slight modification in the way stress is coded in IPA. Instead of placing the stress at the beginning of each syllable, we place the stress at the first vowel in the syllable. This avoids the elicitation of where syllables start or end, an issue which is not strictly necessary for pronunciation, especially when the word is read at a normal pace. The following example shows this modification:

{\em explanation} $\rightarrow$ \textipa{[""Ekspl@n"eIS@n]} (instead of \textipa{[""Ekspl@"neIS@n]})

\noindent Note that this does not mean that we have to split the word into `explan-ation'. It is just that we place the stress on the vowels, since each syllable has only one vowel group.

\subsection{Annotated and Non-annotated Text}

It is generally obvious to tell annotated from non-annotated text and vice-versa, since we see some symbols over and beneath the letters in one case and none in the other.
Nevertheless, in some cases, a word or a part of a document can be left non-annotated. In order to avoid confusion, we need to make that clear. That can be the case with words that have more than one possible pronunciation (and the annotator does not want to choose among them). Another case is when the annotator (which might be an automated system) does not know the annotation for a word. Finally, another case is for foreign words, especially for places and names.

\begin{center}
    \begin{tabular}[h!]{|p{0.9\textwidth}|}    \hline
  \noindent {\bf Non-annotated text indicator}: We use two symbols for opening ({\tt \textbackslash{nonl}\{\}}: \nonl{}) and closing ({\tt \textbackslash{nonr}\{\}}: \nonr{}).
  \\ \hline
\indent Examples: The first day in Spain, we stayed in a sm\rnd{a}ll vill\iot{a}ge c\rnd{a}lled \nonl{}Záragüe\nonr{}, near \nonl{}Setías\nonr{}.
    \\\hline
    \end{tabular}
\end{center}

\noindent Similarly, in a non-annotated text, we could annotate some difficult or technical words.

\begin{center}
    \begin{tabular}[h!]{|p{0.9\textwidth}|}    \hline
  \noindent {\bf Annotated text indicator}: We use two symbols for opening ({\tt \textbackslash{annl}\{\}}): \annl{}) and closing ({\tt \textbackslash{annr}\{\}}: \annr{}).
  \\ \hline
\indent Examples: The word `catastrophe' is pronounced \annl{}cat\st{a}stroph\iot{e}\annr{}.
  \\\hline
    \end{tabular}
\end{center}

\subsection{Silent Letters, Stress and Separators}

One of the most useful annotations is to make a letter silent.

\begin{center}
    \begin{tabular}[h!]{|p{0.9\textwidth}|}    \hline
  \noindent {\bf Silent}: Any letter can be eliminated (it becomes silent) with the symbol ({\tt \textbackslash{si\{a\}}}): \si{a}.
  \\ \hline
\indent Examples: g\si{h}\nat{o}st, wh\brd{e}r\si{e}\st{a}s.
   \\\hline
    \end{tabular}
\end{center}

\noindent A stress symbol can be used to locate stresses for words with more than one syllable.

\begin{center}
    \begin{tabular}[h!]{|p{0.9\textwidth}|}    \hline
  \noindent {\bf Stresses}: Main stress ({\tt \textbackslash{st\{a\}}}): \st{a}. Secondary stress ({\tt \textbackslash{stst\{a\}}}): \stst{a},
  \\ \hline
\indent Examples: ag\st{o}, f\stst{u}ndam\st{e}ntal, \pln{a}nth\st{\pln{o}}logy
   \\\hline
    \end{tabular}
\end{center}

\noindent We have five different separators, which are useful to break digraphs and to separate compound words. They may also have an effect on stress.

\begin{center}
    \begin{tabular}[h!]{|p{0.9\textwidth}|}    \hline
  \noindent {\bf Unstressed separator}: The word is broken but stress is not modified: ({\tt a\textbackslash{se}\{\}e}): a\se{}e, and we can use more than one in a word. The rules for the stress do still see the word as a whole. The main use of this is to break digraphs.
  \\ \hline
\indent Examples: parent\se{}h\oopq{oo}d.
  \\\hline
    \noindent {\bf One-stress separator}: Only the left part is stressed ({\tt a\textbackslash{sel}\{\}e}): a\sel{}e. Only the right part is stressed ({\tt a\textbackslash{ser}\{\}e}): a\ser{}e. 
  \\ \hline
\indent Examples: go\sel{}ing, a\ser{}way.
  \\\hline
    \noindent {\bf Two-stress separator}: Both parts are stressed, but the main stress goes to the left part ({\tt a\textbackslash{selr}\{\}e}): a\selr{}e.
Both parts are stressed, but the main stress goes to the right part ({\tt a\textbackslash{serl}\{\}e}): a\serl{}e.
  \\ \hline
\indent Examples: north\selr{}bound.
  \\\hline\hline
  More Examples: fat\selr{}he\si{a}d, home\selr{}l\"{e}ss/h\nat{o}m\si{e}less/home\se{}less/home\sel{}less,  side\selr{}car/s\nat{\i}d\si{e}c\brd{a}r, a\ser{}while, \rnd{a}l\serl{}tho\si{u}gh, out\serl{}go\se{}ing.
   \\\hline
    \end{tabular}
\end{center}

\noindent Occasionally, me may need to introduce a schwa (an unstressed central vowel)

\begin{center}
    \begin{tabular}[h!]{|p{0.9\textwidth}|}    \hline
  \noindent {\bf Schwa}: ({\tt a\textbackslash{sch\{\}}b}): a\sch{}b. This can be done without introducing the tiny space between letters as ({\tt \textbackslash{ssch\{ab\}}}): \ssch{ab}.
  \\ \hline
\indent Examples: aut\st{i}s\sch{}m / aut\st{i}\ssch{sm}  (only some accents), met\sch{}re / me\ssch{tr}e (this is not needed since there is a rule which fixes this) 
   \\\hline
    \end{tabular}
\end{center}

\subsection{Vowels}

Vowel pronunciation requires many annotations given the reduced number of vowel letters in the Roman alphabet ({\em a,e,i,o,u}) plus the use of ({\em w} and {\em y}), compared to the huge number of vowel phonemes in English. 

English has 17 stressed vowel phonemes (including \textipa{[wA:]} and excluding the rhotic sounds, i.e., when an `r' follows) that can be originated from one or two written vowels.
Fortunately, we do not need 17 different symbols, since one single letter can {\em only} have at most 10 different sounds (excluding the effect of a following `r'). For instance, `o' sounds differently in these words: pot, note, bought, love, Goethe, women, do, woman, now, foie.

\begin{center}
    \begin{tabular}[h!]{|p{0.9\textwidth}|}    \hline
  \noindent {\bf Plain}: Single letter ({\tt \textbackslash{pln\{a\}}}): \pln{a}. Double letter ({\tt \textbackslash{ppln\{ae\}}}): \pln{ae} is not used. Before {\tt i}, it goes like this: ({\tt \textbackslash{pln\{\textbackslash{}i\}}}): \pln{\i}.
  \\ \hline
\indent Examples: \pln{a}nth\st{\pln{o}}logy, s\pln{e}ven, \pln{I}taly, \pln{u}gly.
   \\\hline
    \end{tabular}
\end{center}

\begin{center}
    \begin{tabular}[h!]{|p{0.9\textwidth}|}    \hline
  \noindent {\bf  Natural}:  Single letter ({\tt \textbackslash{nat\{a\}}}): \nat{a}. Double letter ({\tt \textbackslash{nnat\{ae\}}}): \nnat{ae}. Before {\tt i}, it goes like this: ({\tt \textbackslash{nat\{\textbackslash{}i\}}}): \nat{\i}.
  \\ \hline
\indent  Examples: r\nat{a}nge, \nat{e}qual, g\si{h}\nat{o}st, b\nat{\i}nd,  regg\nat{a}e, n\nnat{e\i}ther (GA), \nat{u}n\st{i}te.
   \\\hline
    \end{tabular}
\end{center}

\begin{center}
    \begin{tabular}[h!]{|p{0.9\textwidth}|}    \hline
  \noindent {\bf Broad}: Single letter ({\tt \textbackslash{brd\{a\}}}): \brd{a}. Double letter ({\tt \textbackslash{bbrd\{ae\}}}): \bbrd{ae}. But, ({\tt \textbackslash{bbrd\{a\}}}): \bbrd{a} is incorrect. Before {\tt i}, it goes like this: ({\tt \textbackslash{brd\{\textbackslash{}i\}}}): \brd{\i}.
.
  \\ \hline
\indent Examples: f\brd{a}ther, caf\st{\brd{e}}, ma\sno{ch}\st{\brd{\i}}ne, b\bbrd{ou}ght, f\bbrd{\i{}e}ld, s\brd{u}per. 
   \\\hline
    \end{tabular}
\end{center}

\begin{center}
    \begin{tabular}[h!]{|p{0.9\textwidth}|}    \hline
  \noindent {\bf  Clear}: Single letter ({\tt \textbackslash{clr\{a\}}}): \clr{a}. Double letter ({\tt \textbackslash{cclr\{ae\}}}): \cclr{ae}. But, ({\tt \textbackslash{cclr\{a\}}}): \cclr{a} is incorrect. Before {\tt i}, it goes like this: ({\tt \textbackslash{clr\{\textbackslash{}i\}}}): \clr{\i}.
  \\ \hline
\indent  Examples: s\clr{o}me, l\clr{o}ve, bl\cclr{oo}d, s\clr{e}rgent, M\clr{a}greb, l\clr{\i}n\svo{g}erie.
      \\\hline 
       \end{tabular}
\end{center}

\begin{center}
    \begin{tabular}[h!]{|p{0.9\textwidth}|}    \hline
  \noindent {\bf Central}: Single letter ({\tt \textbackslash{ctr\{a\}}}): \cnt{a}. Double letter ({\tt \textbackslash{cctr\{ae\}}}): \ccnt{ae} is not used. Before {\tt i}, it goes like this: ({\tt \textbackslash{ctr\{\textbackslash{}i\}}}): \cnt{\i}.
  \\ \hline
\indent  Examples: \cnt{a}ny, b\cnt{u}\co{r}y, w\cnt{o}rd, terr\cnt{\i}ble.
   \\\hline
    \end{tabular}
\end{center}

\begin{center}
    \begin{tabular}[h!]{|p{0.9\textwidth}|}    \hline
  \noindent {\bf `Ioted'}: Single letter ({\tt \textbackslash{iot\{a\}}}): \iot{a}.
  \\ \hline
\indent  Examples: \iot{E}nglish, end\iot{e}d, w\iot{o}m\iot{e}n, chick\iot{e}n, b\iot{u}sy, mess\iot{a}ge
   \\\hline
    \end{tabular}
\end{center}

\begin{center}
    \begin{tabular}[h!]{|p{0.9\textwidth}|}    \hline
  \noindent {\bf Round}: Single letter ({\tt \textbackslash{rnd\{a\}}}): \rnd{a}. Double letter ({\tt \textbackslash{rrnd\{ae\}}}): \rrnd{ae} is not used.  Before {\tt i}, it goes like this: ({\tt \textbackslash{rnd\{\textbackslash{}i\}}}): \rnd{\i}. \\ \hline
\indent  Examples: \rnd{a}ll, d\rnd{o}, s\rnd{e}\si{w}, m\si{a}\rnd{u}ve
   \\\hline 
    \end{tabular}
\end{center}

\begin{center}
    \begin{tabular}[h!]{|p{0.9\textwidth}|}    \hline
  \noindent {\bf Opaque}:  Single letter ({\tt \textbackslash{opq\{a\}}}): \opq{a}. Double letter ({\tt \textbackslash{oopq\{ae\}}}): \oopq{ae}. But, ({\tt \textbackslash{oopq\{a\}}}): \oopq{a} is incorrect. Before {\tt i}, it goes like this: ({\tt \textbackslash{opq\{\textbackslash{}i\}}}): \opq{\i}.
  \\ \hline
\indent  Examples: w\opq{a}tch, w\opq{o}man, p\opq{u}t, b\oopq{oo}k, w\oopq{ou}\si{l}d, \opq{e}nvelope
   \\\hline
    \end{tabular}
\end{center}

\begin{center}
    \begin{tabular}[h!]{|p{0.9\textwidth}|}    \hline
  \noindent {\bf `i-diphthong'}: Single letter ({\tt \textbackslash{idp\{a\}}}): \idp{a}. Double letter ({\tt \textbackslash{iidp\{ae\}}}): \iidp{ae}. But, ({\tt \textbackslash{iidp\{a\}}}): \iidp{a} is incorrect. Note that ({\tt \textbackslash{idp\{ae\}}}):\idp{ae} is incorrect.
  \\ \hline
\indent Examples: \iidp{ey}e, \iidp{Ei}nst\iidp{e\i}n, p\idp{a}\brd{e}lla,  f\iidp{o\i}e.
   \\\hline
    \end{tabular}
\end{center}

\begin{center}
    \begin{tabular}[h!]{|p{0.9\textwidth}|}    \hline
  \noindent {\bf `u-diphthong'}: Single letter ({\tt \textbackslash{udp\{a\}}}): \udp{a}. Double letter ({\tt \textbackslash{uudp\{ae\}}}): \uudp{ae}.
  But, ({\tt \textbackslash{uudp\{a\}}}): \uudp{a} is incorrect. Note that ({\tt \textbackslash{udp\{ae\}}}):\udp{ae} is incorrect.
  \\ \hline
\indent Examples: n\uudp{ow}, L\uudp{ao}s, Fr\uudp{eu}dian.
   \\\hline
    \end{tabular}
\end{center}

\subsection{Semiconsonants}

Semiconsonant annotations are special in the way that they can convert any letter (or group of letters) into the (semi-)consonants `y' and `w'.
Additionally, if they are placed between letters, they {\em introduce} the sound.

\begin{center}
    \begin{tabular}[h!]{|p{0.9\textwidth}|}    \hline
  \noindent {\bf  Semiconsonant $w$ }:  Single letter substitution ({\tt \textbackslash{w\{a\}}}): \w{a}. Introduction ({\tt \textbackslash{w\{\}}}): a\w{}b.
  \\ \hline
\indent  Examples: s\w{u}\brd{e}de (substitution), \w{\group{ou}}\brd{\i}ja (substitution), \w{}\clr{o}ne (introduction), \w{}\clr{o}nce (introduction)
   \\\hline
   \noindent {\bf  Semiconsonant $y$ }:  Single letter substitution ({\tt \textbackslash{y\{a\}}}): \y{a}. Introduction ({\tt \textbackslash{y\{\}}}): a\y{}b.
  \\ \hline
\indent  Examples:  h\stst{a}llel\st{u}\y{j}ah (substitution), tort\st{\brd{\i}}\y{\group{ll}}a (substitution), \clr{o}n\y{i}on (substitution), mill\y{i}on  (substitution), fail\y{}ure (introduction).
   \\\hline
    \end{tabular}
\end{center}

\subsection{Consonants}

Consonants are more regular and require a fewer number of symbols than vowels.

\begin{center}
    \begin{tabular}[h!]{|p{0.9\textwidth}|}    \hline
  \noindent {\bf Common change}: ({\tt \textbackslash{co}}): \co{g}.   \\ \hline
\indent  Examples: \co{g}ill, tar\co{g}et, \co{c}eltic,  fa\co{j}\brd{\i}ta, loc\co{h} (loc\si{h} in most accents), LaTe\co{X}
   \\\hline
   
     \noindent {\bf Voiced and voiceless }: ({\tt \textbackslash{vo}}, {\tt \textbackslash{no}}): \vo{s}, \no{si} \\ \hline
\indent  Examples: hou\no{s}e, \no{th}ing, sci\vo{ss}ors, toppe\no{d}
   \\\hline
   
     \noindent {\bf  Soft voiced and voiceless}: ({\tt \textbackslash{svo}}, {\tt \textbackslash{sno}}): \svo{x}, \sno{si}    \\ \hline
\indent  Examples: no\sno{ti}on, v\pln{\i}\svo{si}on, lu\sno{x}ury, mi\si{s}\sno{si}on,  mis\sno{si}on (but it would be better as mi\sno{\group{ssi}}on)
   \\\hline
   
     \noindent {\bf Hard voiced and voiceless }: ({\tt \textbackslash{hvo}}, {\tt \textbackslash{hno}}): \hvo{g}, \hno{ti}    \\ \hline
\indent   Examples: \hno{c}elo, na\hno{t}ure, ques\hno{ti}on, s\nat{o}l\hvo{di}er
   \\\hline
    \end{tabular}
\end{center}

\noindent In some cases, especially when grouping more than two letters with the same annotation, the graphical representation of the annotation may not span all over the letters. In this case, we can use a `group', which is not an annotation, but a graphical/typographical sign to better understand that the annotation affects the group of letters.

\begin{center}
    \begin{tabular}[h!]{|p{0.9\textwidth}|}    \hline
  \noindent {\bf Grouping}: Useful to group two or more letters, before applying the annotation. It is generally applied for consonant annotations, but it can also be applied to the `silent' symbol `\si{}'.
 ({\tt \textbackslash{group}}): \group{abc}.   \\ \hline
\indent  Examples: mi\sno{\group{ssi}}on, \iot{e}x\st{a}\hvo{\group{gg}}er\stst{a}te, tort\brd{\i}\y{\group{ll}}a, y\opq{a}\si{\group{ch}}t, lo\si{\group{ug}}\co{h}.
   \\\hline
    \end{tabular}
\end{center}

\noindent These are the symbols we will use to convert any English text into an annotated text such that its pronunciation can be derived unambiguously.
We have not included here an exhaustive explanation of the sounds of every annotation. Doing so would allow the annotation of any text. But, without any rule, this would yield verbose annotations as shown on the left of Table \ref{tab:tradeoff}. In the following section we present a set of rules to dramatically reduce the number of annotations. Thus, the exact IPA phonetic of each combination of letter (or digraph) + annotation will be precisely defined next.

\section{Rules of Pronunciation for Annotated English}\label{sec:rules}

The rules we present in this section will allow a reader to unambiguously determine the pronunciation of an annotated word.
The basic idea is that the words (or parts of words) who have no annotations will be pronounced as the rule says. The words (or parts of words) with annotations will highlight an exception to the rules, and the annotation will clarify the irregular sound it corresponds to.
For instance, the word `that' requires no annotation. First, the digraph `th' here is voiced, which follows a rule saying that it must be so when appearing before a vowel. Second, `a' is in a plain position (followed by a consonant at the end of the word). Third, `t' is regularly pronounced. Quite differently, the word `hou\no{s}e' requires an annotation, because `h', `ou' and the final `e' sound as the rules indicate, but the consonant `s' must be annotated as voiceless since this case breaks the rule that an `s' is voiced between vowels. 

The rules must be applied in a given order. We arrange the process in 12 steps, as follows:

\begin{enumerate}
\item Discard (crossed-out) letters.
\item Identify word segments by hyphens, apostrophes and separation annotations
\item Parse critical digraphs (`ng', `n\co{g}', `gh', \vo{gh}, `wh', `wr', `qu', `gu'.).
\item Distinguish vowel and consonant groups. Handling semiconsonants 
\item Distinguish vowel units in groups using digraphs (aa, ae, ai, ay, au, aw, ea, ee, ei, ey, eu, ew, oa, oi, oy, oo, ou, ow).
\item Place the stresses for each vowel unit.
\item Categorise vowel ocurrences (stressed/unstressed, natural/plain and rhotic/non-rhotic). 
\item Evaluate vowel units. 
\item Distinguish consonant units in groups using digraphs (kn, pn, gn, cn, ph, ch, sh, ps, rh, pt, th, gg, ss).
\item Evaluate consonant units.
\item Recompose the words.
\item Postprocess the IPA result (elimination of repeated consonants and inclusion of schwa into final -\#r).
\end{enumerate}

\begin{figure}
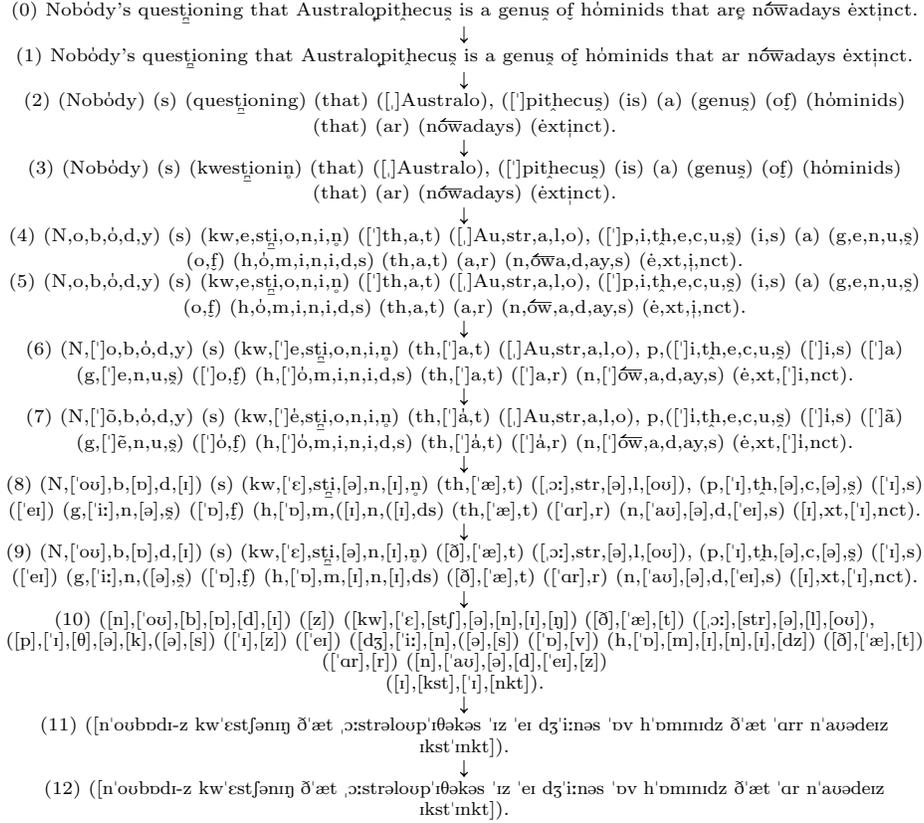

	\centering
		{%\tiny
		  \scriptsize
		  %\footnotesize
		(0)  Nob\pln{o}dy's ques\hno{ti}oning that Australo\serl{}pi\no{th}ecu\no{s} is a genu\no{s} o\vo{f} h\pln{o}minids that ar\si{e} n\uudp{ow}adays \iot{e}xt\st{i}nct. \\
		  $\downarrow$ \\
 		(1)  Nob\pln{o}dy's ques\hno{ti}oning that Australo\serl{}pi\no{th}ecu\no{s} is a genu\no{s} o\vo{f} h\pln{o}minids that ar n\uudp{ow}adays \iot{e}xt\st{i}nct. \\
		  $\downarrow$ \\
		(2) (Nob\pln{o}dy) (s) (ques\hno{ti}oning) (that) (\textipa{[""]}Australo), (\textipa{["]}pi\no{th}ecu\no{s}) (is) (a) (genu\no{s}) (o\vo{f}) (h\pln{o}minids) (that) (ar) (n\uudp{ow}adays) (\iot{e}xt\st{i}nct). \\
		  $\downarrow$ \\
		(3) (Nob\pln{o}dy) (s) (kwes\hno{ti}oni\co{n}) (that) (\textipa{[""]}Australo), (\textipa{["]}pi\no{th}ecu\no{s}) (is) (a) (genu\no{s}) (o\vo{f}) (h\pln{o}minids) (that) (ar) (n\uudp{ow}adays) (\iot{e}xt\st{i}nct). \\
		    $\downarrow$ \\
		(4) (N,o,b,\pln{o},d,y) (s) (kw,e,s\hno{ti},o,n,i,\co{n}) (\textipa{["]}th,a,t) (\textipa{[""]}Au,str,a,l,o), (\textipa{["]}p,i,\no{th},e,c,u,\no{s}) (i,s) (a) (g,e,n,u,\no{s}) (o,\vo{f}) (h,\pln{o},m,i,n,i,d,s) (th,a,t) (a,r) (n,\uudp{ow}a,d,ay,s) (\iot{e},xt,\st{i},nct). \\
		(5) (N,o,b,\pln{o},d,y) (s) (kw,e,s\hno{ti},o,n,i,\co{n}) (\textipa{["]}th,a,t) (\textipa{[""]}Au,str,a,l,o), (\textipa{["]}p,i,\no{th},e,c,u,\no{s}) (i,s) (a) (g,e,n,u,\no{s}) (o,\vo{f}) (h,\pln{o},m,i,n,i,d,s) (th,a,t) (a,r) (n,\uudp{ow},a,d,ay,s) (\iot{e},xt,\st{i},nct). \\
		    $\downarrow$ \\
		(6) (N,\textipa{["]}o,b,\pln{o},d,y) (s) (kw,\textipa{["]}e,s\hno{ti},o,n,i,\co{n}) (th,\textipa{["]}a,t) (\textipa{[""]}Au,str,a,l,o), p,(\textipa{["]}i,\no{th},e,c,u,\no{s}) (\textipa{["]}i,s) (\textipa{["]}a) (g,\textipa{["]}e,n,u,\no{s}) (\textipa{["]}o,\vo{f}) (h,\textipa{["]}\pln{o},m,i,n,i,d,s) (th,\textipa{["]}a,t) (\textipa{["]}a,r) (n,\textipa{["]}\uudp{ow},a,d,ay,s) (\iot{e},xt,\textipa{["]}i,nct). \\
		    $\downarrow$ \\
		(7) (N,\textipa{["]}\nat{o},b,\pln{o},d,y) (s) (kw,\textipa{["]}\pln{e},s\hno{ti},o,n,i,\co{n}) (th,\textipa{["]}\pln{a},t) (\textipa{[""]}Au,str,a,l,o), p,(\textipa{["]}\pln{\i},\no{th},e,c,u,\no{s}) (\textipa{["]}\pln{\i},s) (\textipa{["]}\nat{a}) (g,\textipa{["]}\nat{e},n,u,\no{s}) (\textipa{["]}\pln{o},\vo{f}) (h,\textipa{["]}\pln{o},m,i,n,i,d,s) (th,\textipa{["]}\pln{a},t) (\textipa{["]}\pln{a},r) (n,\textipa{["]}\uudp{ow},a,d,ay,s) (\iot{e},xt,\textipa{["]}\pln{\i},nct). \\
		    $\downarrow$ \\
		   (8) (N,\textipa{["oU]},b,\textipa{[6]},d,\textipa{[I]}) (s) (kw,\textipa{["E]},s\hno{ti},\textipa{[@]},n,\textipa{[I]},\co{n}) (th,\textipa{["\ae]},t) (\textipa{[""O:]},str,\textipa{[@]},l,\textipa{[oU]}), (p,\textipa{["I]},\no{th},\textipa{[@]},c,\textipa{[@]},\no{s}) (\textipa{["I]},s) (\textipa{["eI]}) (g,\textipa{["i:]},n,\textipa{[@]},\no{s}) (\textipa{["6]},\vo{f}) (h,\textipa{["6]},m,(\textipa{[I]},n,(\textipa{[I]},ds) (th,\textipa{["\ae]},t) (\textipa{["Ar]},r) (n,\textipa{["aU]},\textipa{[@]},d,\textipa{["eI]},s) (\textipa{[I]},xt,\textipa{["I]},nct). \\
		   		    $\downarrow$ \\
		   (9) (N,\textipa{["oU]},b,\textipa{[6]},d,\textipa{[I]}) (s) (kw,\textipa{["E]},s\hno{ti},\textipa{[@]},n,\textipa{[I]},\co{n}) (\textipa{[D]},\textipa{["\ae]},t) (\textipa{[""O:]},str,\textipa{[@]},l,\textipa{[oU]}), (p,\textipa{["I]},\no{th},\textipa{[@]},c,\textipa{[@]},\no{s}) (\textipa{["I]},s) (\textipa{["eI]}) (g,\textipa{["i:]},n,(\textipa{[@]},\no{s}) (\textipa{["6]},\vo{f}) (h,\textipa{["6]},m,\textipa{[I]},n,\textipa{[I]},ds) (\textipa{[D]},\textipa{["\ae]},t) (\textipa{["Ar]},r) (n,\textipa{["aU]},\textipa{[@]},d,\textipa{["eI]},s) (\textipa{[I]},xt,\textipa{["I]},nct). \\
		    $\downarrow$ \\
		   (10) (\textipa{[n]},\textipa{["oU]},\textipa{[b]},\textipa{[6]},\textipa{[d]},\textipa{[I]}) (\textipa{[z]}) (\textipa{[kw]},\textipa{["E]},\textipa{[stS]},\textipa{[@]},\textipa{[n]},\textipa{[I]},\textipa{[N]}) (\textipa{[D]},\textipa{["\ae]},\textipa{[t]}) (\textipa{[""O:]},\textipa{[str]},\textipa{[@]},\textipa{[l]},\textipa{[oU]}), (\textipa{[p]},\textipa{["I]},\textipa{[T]},\textipa{[@]},\textipa{[k]},(\textipa{[@]},\textipa{[s]}) (\textipa{["I]},\textipa{[z]}) (\textipa{["eI]}) (\textipa{[dZ]},\textipa{["i:]},\textipa{[n]},(\textipa{[@]},\textipa{[s]}) (\textipa{["6]},\textipa{[v]}) (h,\textipa{["6]},\textipa{[m]},\textipa{[I]},\textipa{[n]},\textipa{[I]},\textipa{[dz]}) (\textipa{[D]},\textipa{["\ae]},\textipa{[t]}) (\textipa{["Ar]},\textipa{[r]}) (\textipa{[n]},\textipa{["aU]},\textipa{[@]},\textipa{[d]},\textipa{["eI]},\textipa{[z]})
		   
		    (\textipa{[I]},\textipa{[kst]},\textipa{["I]},\textipa{[nkt]}). \\
		    $\downarrow$ \\		   
		   (11) (\textipa{[n"oUb6dI-z kw"EstS@nIN D"\ae{}t  ""O:str@loUp"IT@k@s "Iz "eI dZ"i:n@s "6v h"6mInIdz D"\ae{}t "Arr n"aU@deIz Ikst"Inkt]}). \\
		    $\downarrow$ \\	
		   (12) (\textipa{[n"oUb6dI-z kw"EstS@nIN D"\ae{}t  ""O:str@loUp"IT@k@s "Iz "eI dZ"i:n@s "6v h"6mInIdz D"\ae{}t "Ar n"aU@deIz Ikst"Inkt]}). \\
		}
		\caption{An example of the process of interpretation following the rules.}
\label{fig:process} 
\end{figure}

\noindent The example in Figure \ref{fig:process} shows how an annotated sentence is converted into IPA following the previous process.

Next, each step of the process is explained in one subsection each.

\subsection{Discard (crossed-out) Letters}

The first step of the process is easy. As we saw, the symbol `\si{ }' indicates that the letter is crossed out (and hence becomes silent), which means it is deleted and discarded for any subsequent rule. For instance, `pot' and `pot\si{e}' are considered equal.

Examples: c\pln{u}\si{p}b\si{o}ard (\textipa{[k2b@rd]}), dum\si{b} (\textipa{[d2m]}), ar\si{e} (\textipa{[Ar]}).

\noindent Once silent letters are removed we proceed to identify groups and segments.

\subsection{Identify Word Segments. Hyphens, Apostrophes and Separators}

We use the notions of groups and segments, instead of a notion of syllable, which is not used in our annotation system.
Segments are independent parts of a word. Words only have one segment, unless they are broken by a written hyphen (e.g. t\bbrd{wo}-way), an apostrophe (e.g. it's) or by a separation annotation (`\se{}', `\sel{}', `\ser{}', `\selr{}', `\serl{}').  

When an apostrophe has a right-hand side starting with an `r' (y\si{o}u're, they're, etc.), only one segment is created, in order to consider the possible effect of the `r' over the vowels on the left-hand-side.

Examples: t\bbrd{wo}-way (t\bbrd{wo}-way), it's (it-s), Tom's (Tom-s), I'll (I-ll), y\si{o}u're (y\si{o}ure), Ire\sel{}land (Ire-land), incl\st{u}ding (incl\st{u}ding), p\pln{o}licy\selr{}maker (p\pln{o}licy-maker), go\sel{}ing (go-ing), under\serl{}go (under-go), fat\selr{}he\si{a}d (fat-he\si{a}d).

Segments are very important to break possible digraphs or to break up compound words, as we have seen in some examples above.

\subsection{Parse Critical Digraphs}

There are some consonant digraphs that deserve special rules at this early moment of the process, because they become silent or partly silent, or they involve a semiconsonant (and we need to distinguish between consonants and vowels the sooner the better). The rules are as follows and are applied in this order. 

\begin{enumerate}
\item `ng', `n\co{g}'. The digraph `ng' becomes `\co{n}' when it is at the end of a segment, or before any consonant except `l' and `r'. In other words, we have a softer \textipa{[N]} before any combination which makes impossible to articulate the `g' sound. Examples: `sing', `y\si{o}ungster', `Sing\si{h}'. In all the other positions, the digraph `ng' becomes  `nj' (\textipa{[ndZ]}) if it is followed by `e', `i' or `y' (e.g., ginger, \nat{a}ngel, sp\clr{o}ngy) and `\co{n}\co{g}' (\textipa{[Ng]}) in all the other cases (`bingo', `finger', `\iot{E}nglish', `single', `angry'). Note how the words `sing\sel{}er', `bring\sel{}ing' and `cling\sel{}y' are separated into two segments in order to get the proper sound. When `g' is annotated as `\co{g}', the digraph `n\co{g}' is always converted into `\co{n}\co{g}' (\textipa{[Ng]}) (e.g. `an\co{g}er'). When `g' after `n' is annotated as `\hvo{g}' (i.e. `n\hvo{g}'), it is always (\textipa{[ndZ]}) (e.g. `dun\hvo{ge}on'), and when `g' is annotated as `\svo{g}', it is always pronounced \textipa{[nZ]} as in `l\clr{i}n\svo{g}erie'. In these two cases (`n\hvo{g}' and `n\svo{g}') we do not need to perform any action at this moment of the process.

 \item `gh', \oopq{gh}. The digraph `gh' becomes silent everywhere. For instance, it becomes silent for words such as `high', `higher', `tho\si{u}gh', `\no{th}\bbrd{ou}ght', `highness', `neighbour', while it does not for `t\uudp{ou}\co{gh}',  `big\serl{}h\pln{e}\si{a}d\iot{e}d'. At this moment of the process, we also apply the annotation \oopq{gh}, which makes `gh' convert into `\schwa', as in `\pln{E}dinbur\oopq{gh}'. In other cases, where the digraph has been previously destroyed, there is no need to apply this rule, such as in `g\si{h}\nat{o}st' or `M\clr{a}g\si{h}reb'. The annotations \co{gh}, \no{gh} are processed later in the process.

\item `wh'. The digraph `wh' becomes `\co{w}' (that will be finally transcribed into (\textipa{[w]} or \textipa{[\*w]} depending on the dialect) at the beginning of a segment or when it follows a consonant. Examples: `handwheel', `when', `a\ser{}while, w\cnt{o}rthwhile, no\sel{}wh\brd{e}re, \pln{e}v\si{e}ry\sel{}wh\brd{e}re, s\clr{o}m\si{e}wh\brd{e}re, but not with `c\uudp{ow}\selr{}hide' (because it follows a vowel and also because it is annotated). When annotated as `\co{wh}', this is left for later steps of the process (and will be finally transcribed into (\textipa{[\*w]})). Of course we do not have to care when the digraph has been previously destroyed, as in `\si{w}h\rnd{o}' or `\si{w}hole'.

\item `wr'. The digraph `wr' becomes `r' except when it comes after a vowel {\em a,e,i,o,u}. Examples: `write', `wrong', `ship\selr{}wreck', `re\serl{}wr\st{i}te' (because of the separation), `a\ser{}wry' (because of the separation). It does not disappear in `Jewry', `arrowroot', d\uudp{ow}ry.

\item `qu'. The digraph `qu' is always understood as `kw' before a,e,i,o,u,y, and does not need any annotation for the `u' to sound as a semiconsonant, as in `qu\opq{a}lity', `quo', `quest', `\nat{e}qual', `liquor', `aqua', `barb\iot{e}q\stst{u}e' (better `barb\iot{e}c\stst{u}e'. If the `u' does not sound, it has to be crossed out (`\nat{o}p\st{a}q\si{u}e', `crit\brd{\i}q\si{u}e)', `q\nnat{ue}\si{u}e', `q\si{u}\brd{e}\no{s}ad\st{\brd{\i}}\y{\group{ll}}a').

\item `gu'. The digraphs `gu' is always understood as `\co{g}w' before a,o,u and as `\co{g}' before e,i,y.  For instance, the `u' is converted into the semiconsonant sound `w' in `guat', `Gu\stst{a}tem\st{a}la', `gu\brd{a}no', `Pa\co{r}agu\iidp{ay}', `langu\iot{a}ge'.  It is silent in `guess', `guest', `b\pln{a}gu\st{e}tte' `gu\pln{\i}n\iot{e}\si{a}', `disgu\st{i}se', `guide', `guilty', `guit\st{a}r', `guy' (which becomes `\co{g}y'), `Port\udp{u}gu\nat{e}se', coll\st{e}ague, \pln{a}nal\pln{o}gue (note that the `u' becomes silent at this stage of the process, while the `e' will become silent by a rule we will explain below), vogue, t\clr{o}ngue, fat\st{\brd{\i}}gue. The `u' is kept as a vowel when it is followed by a consonant or it is annotated, as in `gun', `lang\si{u}ur', `g\brd{u}bernat\st{o}rial', `\stst{a}mb\st{\pln{\i}}g\nat{u}ous', `\stst{a}mb\st{\pln{\i}}g\nat{u}ity', `peng\w{u}in', `dist\st{i}ng\w{u}ish', `arg\nat{u}e'. The `u' can also become silent before an `a', but in this case, following the rules, it requires a cross-out annotation, such as in `g\si{u}ard', `g\si{u}\stst{a}\co{r}ant\st{e}e'.

\item `ah', `eh', `ih', `oh', `uh', `yh', `wh'. These digraph make the 'h' silent before a consonant. In other words, `h' is silent between vowel and consonant. Examples: \brd{a}h, \brd{e}h, oh, ooh, \pln{u}h, h\pln{u}h, Phar\si{a}oh, Phar\si{a}ohs, Sh\brd{a}h, Y\brd{a}hw\brd{e}h (note that `Y\brd{a}\si{h}w\brd{e}h' would create the digraph `aw'), y\pln{e}ah, Sarah, C\nat{o}hn, K\brd{a}hn, John, K\brd{u}hn, N\brd{a}hcolite, F\brd{a}hd, D\brd{a}hl, K\nat{o}hl, b\brd{u}hl, L\nat{e}hman, \nat{o}hm, B\brd{a}hr\st{a}in, Bohr, R\brd{u}hr. 

\end{enumerate}

\noindent It is interesting to remark (again) that the previous rules are applied in the order they appear. For instance, the word `bi\serl{}lingual' is converted by the `ng' rule into (bi-li\co{n}\co{g}ual). The `gu' rule finally converts it into (bi-li\co{n}\co{g}wal)
There are more digraph rules for vowels and consonants, but they will be seen when vowels and consonants are processed.
We have placed some of them here, because it is very important to consider `w' a semiconsonant and not a vowel in some situations, and the `ng' rule must come before the `gu'. The `gh' rule is given now to consider the introduced vowel \oopq{gh}, and to make the natural sound for `i' in `high'.

\subsection{Distinguish Vowel and Consonant Groups. Handling Semiconsonants}

After the previous process, we are closer to distinguish between vowels and consonants, but there are still some details about `y', `w' and other vowels becoming consonants (and viceversa) that we need to consider.

It is traditionally understood that:

\begin{itemize}
\item Vowels in English are $a, e, i, o, u$.
\item Consonants in English are $b$, $c$, $d$, $f$, $g$, $h$, $j$, $k$, $l$, $m$, $n$, $p$, $q$, $r$, $s$, $t$, $x$, $z$.
\item The letters $w$ and $y$ are considered vowels in some cases and (semi)consonants in other cases.
\end{itemize}

\noindent Apart from clarifying $w$ and $y$, we find cases where $u$ becomes a (semi)consonant in some special cases (apart from `qu', `gua', `guo', `guu' and eventually when annotated after other consonants), and some other annotations converting vowels into consonants and vice-versa. 

We start analysing when $w$ and $y$ are vowels and when they are consonants.
The rule is as follows. The letters $w$ and $y$ are vowels if they are before a consonant, at the end of the group, they are part of a vowel digraph ($ay$, $ey$, $oy$, $aw$, $ew$, $ow$), or they are affected by a vowel annotation. Examples: `cwm', `lynx', `my', `cry', `ray', `d\iot{e}l\st{a}yed', `boy\selr{}fr\si{i}end', `b\si{u}ying', `flowing', `\idp{ey}e' are vowels, while `yell', `lawyer', `when' (remember that this arrives as `\co{w}en' and not `w-hen' at this moment of the process'), `will', `k\brd{\i}w\brd{\i}', `b\iot{e}\ser{}ware' are consonants. Note that if $w$ or $y$ are crossed out, there is no need to decide, such as in `\si{w}hole'. Also, $y$ is considered a vowel if found after one or more consonants. For instance, $y$ is a vowel in `bye', `cyan', `L\pln{\i}byan', `By\pln{e}lor\st{u}\sno{ssi}an', etc. This is not applied for $w$ (`twine'). We can combine $y$ and $w$ as in `W\nat{y}\st{o}ming', where we have that `w' is a consonant and `y' is a vowel. 

There are several annotations that may convert a vowel into a consonant.
If the letter `u' is annotated as `\w{u}', it is transformed into a (semi)consonant `w'. Examples: the `u' in  `peng\w{u}in', s\w{u}\brd{a}ve, s\w{u}\nat{e}de, p\w{u}erto becomes the semiconsonant `w' (remember we do not have to annotate  Pa\co{r}agu\iidp{ay}, but the `u' also becomes `w'), while the `u' in `gun', `g\brd{u}r\brd{u}', `dual', `fuel', `duo' is clearly a vowel. Note that we do not have to care about cases where the `u' has been previously deleted, such as `g\si{u}\st{a}rant\stst{ee}'.
Remember that the digraph `qu' is always understood as `qw' before a,e,i,o,u,y, and does not need any annotation, as in `qu\opq{a}lity', `quo', `quest', `queen', `quas\nat{\i}', `\nat{e}qual', `liquor', `aqua', `colloquy'. But note that in `Q\rnd{u}r\st{\brd{a}}n', `Q\opq{u}m' (a city), etc., it is considered a vowel. Some words taken through Spanish or French may require annotations since the `u' is silent, such as `Q\si{u}ech\w{u}a',  `liq\si{u}\si{e}\st{\nat{u}}r' (or perhaps better `l\pln{\i}q\nnat{\st{u}e}\si{u}r'), `burl\st{e}sq\si{u}e', `\nat{o}p\st{a}q\si{u}e', `crit\brd{\i}q\si{u}e)', `q\si{u}\brd{e}\no{s}ad\st{\brd{\i}}\y{\group{ll}}a' or a formal pronunciation of `Q\si{u}eb\st{e}c'. Some other strange spellings may also require this, such as `q\si{u}\si{a}\brd{y}', `q\nnat{ue}\si{u}e'. As we have also seen, `gue', `gui' and `guy' make the `u' silent, while `gua', `guo' and `guu' convert it into `w', unless annotated, such as in `g\si{u}\st{a}rant\stst{ee}' and `peng\w{u}in'.

The groups `o'  and `ou' in some words may become `w' as well. This is also denoted by `\w{}'. If the sound is completely replaced, as in `\si{O}\w{u}\brd{\i}ja', `v\w{o}\nat{y}e\si{u}r' (RP), `c\si{h}\w{o}\nat{\i}r' we use `\w{}' just below the letter (or the group of letters, `\w{Ou}\brd{\i}ja', although this may be confusing if we do not use the grouping: `\w{\group{Ou}}\brd{\i}ja'). If the vowel sound is preserved and the `w' appears before the vowel, we denote that by \w{} but it is placed between the previous letter and the vowel, as in `\w{}\clr{o}ne',  `\cnt{a}ny\w{}\clr{o}ne' or Q\w{}\opq{a}ntas. Note that we introduce the `w' sound in our decomposition, so `\w{}\clr{o}ne' becomes (w,\clr{o},n,e). 

The letters `i' and `y' can be transformed into or forced into a (semi)consonant `y', by using the annotation `\y{}'. For instance, `\clr{o}n\y{i}on', `bun\y{i}on' (note that `n\y{i}' is now considered a complex consonant), \si{H}\y{i}erro (the Canary island), Ken\y{y}a (note that even it is a `y' it would be considered a vowel if not annotated), C\pln{a}lif\st{o}rn\y{i}a, can\y{i}on, mil\y{i}\st{\cnt{e}}\si{u}, mill\y{i}on. But note that `R\opq{o}m\st{a}nian', `L\pln{\i}bya', \nat{I}b\st{e}rian, s\pln{\i}mian, etc., maintain the vowel (this also depends on the accent).
The same symbol `\y{}' can also be used to annotate the introduction of the phoneme \textipa{[j]} between the letters where it must be placed. Examples: `fail\y{}ure' (which is better annotated as fail\idp{u}re), `la\no{s}\st{a}\si{g}n\y{}a' (the introduction of a new consonant makes the group complex).

Finally, there is a bizarre \no{u} sound which becomes f, as in l\si{i}\pln{e}\no{u}t\st{\pln{e}}nant.
Some other vowels may become consonants in conjunction with other consonants if so annotated. For instance, \hvo{di}, \hvo{de}, \sno{ti}, ... as in s\nat{o}l\hvo{di}er or no\sno{ti}on.

After the previous cases, it may seem difficult to tell between consonants and vowels. But it is not so if we summarise and put everything together. The following set of items clarify (and summarise) this:

\begin{itemize}
\item Vowels in Annotated English are $a, e, i, o, u$ and any group of characters which have an annotation above the letter (e.g., \nat{a}, \uudp{ow}, \oopq{gh}).
The letters $w$ and $y$ are vowels if they are before a consonant, at the end of the group or they are part of a vowel digraph ($ay$, $ey$, $oy$, $aw$, $ew$, $ow$). Also, $y$ is considered a vowel if found after a consonant.
\item Consonants in Annotated English are $b$, $c$, $d$, $f$, $g$, $h$, $j$, $k$, $l$, $m$, $n$, $p$, $q$, $r$, $s$, $t$, $x$, $z$ and any (possibly empty) group of characters which have an annotation below the letter (e.g., \vo{f}, \y{}, \w{ou}, \no{u}). The letters $w$ and $y$ are considered consonants when they are not considered vowels. The letter $u$ is considered a consonant in the unannotated digraph `qu', and in the unannotated groups `gua', `guo', `guu'.
\end{itemize}

\noindent We will use the notation @ for a vowel and \# for a consonant (this is notation for this explanation, not part of the annotations).

Only after the previous rules, we are able to decompose each word segment into vowel and consonant groups.

For each segment, we have to distinguish its groups.
We use the term vowel group for a series of consecutive vowels in a segment. Similarly, a consonant group is a series of consecutive consonants.
Note that a vowel group is not necessarily a unit (a single vowel sound or a diphthong). For instance, `ie' in `science' is a vowel group with two units. Similarly, `d\rnd{o}ing' has a vowel group `\rnd{o}i' which is composed of two units. Some other two-vowel groups can have one unit, such as `ow' in `grow'.
Note that groups cannot be made across consecutive segments. For instance, the groups in go\sel{}ing preclude $o$ and $i$ from being together.

Examples: cons\st{\pln{e}}c\udp{u}tive (c,o,n,s,\st{\pln{e}},c,\nat{u},t,i,v,e), science (sc,ie,nc,e), de\si{b}t (d,e,t), f\brd{a}s\si{t}en (f,\brd{a},s,e,n), go\sel{}ing (g,o - i,ng), diph\no{th}ong (d,i,ph\no{th},o,\co{n}), `cry' (cr,y), 'boy\selr{}fr\si{i}end' (b,oy,fr,end), `b\si{u}ying' (b,yi,ng), `flowing' (fl,owi,\co{n}), `\iidp{ey}e' (\iidp{ey}e), when (\co{w},e,n), `k\brd{\i}w\brd{\i}' (k,\brd{\i},w,\brd{\i}), `b\iot{e}\ser{}ware' (b,\iot{e},w,a,r,e), `cyan' (c,ya,n), `twine' (tw,i,n,e), `Gu\brd{a}tem\st{\brd{a}}la' (Gw,\brd{a},t,e,m,\brd{a},l,a), 'g\brd{u}\co{r}\brd{u}' (g,\brd{u},\co{r},\brd{u})
, `g\si{h}\nat{o}st' (g-\nat{o}-st),  `ship\selr{}wreck' (sh,i,p,r,e,ck), `peng\w{u}in' (p,e,ng\w{u},i,n), `\co{g}\si{u}est' (\co{g},e,st), 
`qu\opq{a}lity' (kw,\opq{a},l,i,t,y),  `liquor' (l,i,kw,o,r), `\w{\group{Ou}}\brd{\i}ja' / `\si{O}\w{u}\brd{\i}ja' (\w{\group{Ou}},\brd{\i},ja) / (\w{u},\brd{\i},ja), `a\ser{}way' (a,w,ay).

\subsection{Distinguish Vowel Units}

The first thing to be done when a vowel group is to be pronounced is to determine the vowel digraphs (in case) in order to determine the vowel units. Vowel units are also useful to place the stress, since stress is always on vowel units (a syllabe has one and only one vowel unit).

The vowel pairs which are considered digraphs and those which are not are shown below:

\begin{itemize}
\item Digraphs: $aa$, $ae$, $ai$, $ay$, $au$, $aw$, $ea$, $ee$, $ei$, $ey$, $eu$, $ew$, $oa$, $oi$, $oy$, $oo$, $ou$, $ow$. 
\item Non-digraphs: $ao$, $eo$, $oe$, $i@$ ($ia$, $ie$, $ii$, $iy$, $io$, $iu$, $iw$), and $u@$ ($ua$, $ue$, $ui$, $uy$, $uo$, $uu$, $uw$).
\end{itemize}

\noindent There are no vowel trigraphs. A vowel group with only one vowel is a vowel unit. When two or more vowels are found together in a group, we form digraphs from left to right (so, `eei' becomes (ee,i) and not (e, ei)). The parts which are not grouped by a digraph are left as single units. Note that any deleted vowel is not considered for digraphs, since they are deleted by a previous rule.

For instance, `bait' applies the digraph and `ai' is maintained as a unit (b,ai,t). The word `science' becomes (sc,i,e,nc,e) because `ie' is not a digraph. If a vowel in a group has an annotation, this breaks any possible digraph. For instance, `y\pln{e}ah' breaks the possible `digraph' $ea$ and then becomes (y,\pln{e},a,h). If the second vowel has a stress annotation it also breaks the unit (\nat{o}\st{a}\no{s}i\no{s}). When an annotation spans over two vowels, it means that these two vowels make a unit, as in `f\bbrd{\i{}e}ld' (f,\bbrd{\i{}e},ld).
When a stress annotation is placed over the second letter of a vowel group, than the vowel group is split into two units (e.g., s\brd{u}pra\st{i}ng\w{u}inal (s,\brd{u},pr,a,\st{i},ng\w{u},i,n,a,l). A silent annotation between two vowels put them together (e.g., `v\nat{e}\si{h}icle' (v,\nat{e},i,cl,e) must annotate the first `e' to avoid the interpretation (v,ei,cl,e). 
It is the same for the silent digraph `gh', but luckily in words such as `higher' we have the cluster `ie' which is not a digraph (h,i,e,r). And, of course, separators break units (e.g. a\ser{}way)

Examples: `go\sel{}ing' (g,o - i,ng), `b\si{u}ying' (b,\si{u}y,i,ng), `flowing' (fl,ow,i,ng), `seeing' (s,ee,i,ng) `lion' (l,i,o,n), `new' (n,ew), `\iidp{ey}e' (\iidp{ey},e), `algae' (a,l,g,ae), `m\brd{\i}\si{a}{\cclr{\st{o}w}}' (m,\brd{\i},{\cclr{\st{o}w}}), `Hallow\st{e}en' (H,a,ll,ow,\st{e}e,n), `me\uudp{\st{o}w}' (m,e,\uudp{\st{o}w}), `sh\rnd{o}e' (sh, \rnd{o}, e), `coc\si{h}l\iot{e}ae' (c,o,c,l,\iot{e},ae), `ae\st{o}lian' (ae,\st{o},l,i,a,n), Ha\se{}w\iidp{\st{a}\i}i (H,a,w,\iidp{\st{a}\i},i) (note how the cluster `aw' is broken), `\pln{E}dinbur\oopq{gh}' (\pln{E},d,i,nb,u,r,\oopq{gh}), `gluey' (gl,u, ey), `gluier' (gl, u, i, e, r).

Following the previous rules, we realise that we can only have vowel units with one or two vowels, never more.

\subsection{Place the Stresses for Each Vowel Unit}

Stress applies to vowel groups and ultimately to vowel units. Here we account to placing the stress over the vowel groups.
% In subsequent sections, we will deal about cases where a vowel group has more than one vowel unit (such as `science').
There are stressed segments (which can be either primary or secondary) and unstressed segments. We follow a set of rules, in this order:

\begin{enumerate}
\item If a word is not separated into segments, it is a primary-stressed segment (e.g., `hou\no{s}e', `difference', `ag\st{o}'). 
\item If a word is separated by a hyphen, only the first segment is considered primary-stressed and the second is unstressed (e.g., `yo-yo')).  
\item If a segment is separated by an apostrophe, only the first segment is considered primary-stressed and the second is unstressed (e.g., `it's'). This rule can be broken if the apostrophe is annotated: o\ser{'}clock.  
\item The segment `the' is special. It is considered unstressed if it is followed by another segment (word) starting with a consonant and stressed elsewhere.
\end{enumerate}

\noindent From these principles, more segments can be created by further separations due to annotations, which also have consequences on the stress.

\begin{enumerate}
\item If we see the separator `\se{}', the two new segments do not modify their stresses. Examples: parent\se{}h\oopq{oo}d (only the first segment is stressed).
\item If we see the separator `\sel{}', the first segment becomes primary-stressed and the second unstressed. Examples: go\sel{}ing, home\sel{}less.
\item If we see the separator `\ser{}', the first segment becomes unstressed and the second primary-stressed. Examples: a\ser{}bound.
\item If we see the separator `\selr{}', the first segment becomes primary-stressed and the second secondary-unstressed. Examples: fat\selr{}he\si{a}d, north\selr{}bound, side\selr{}car.
\item If we see the separator `\serl{}', the first segment becomes secondary-stressed and the second primary-unstressed. Examples: \rnd{a}lth\st{o}\si{u}gh.
\end{enumerate}

\noindent The two first separators may seem to have the same effect, but they are different. For instance, the following annotations are not equivalent: out\se{}go\sel{}ing and out\sel{}go\sel{}ing (the first only gives primary stress to the first vowel group while the second gives primary stress to the two first vowel groups). In any case, the correct annotation for this word would be: out\serl{}go\se{}ing (or better, \stst{o}utg\st{o}\se{}ing).

It is interesting to see examples of unstressed segments: Ire\sel{}land (land), go\sel{}ing (ing), \iot{e}n\ser{}gage (\iot{e}n),  Birming\sel{}\si{h}am (\si{h}am), It's ('s), They've ('ve), The wind (The).

Now we concentrate on placing the stress over the vowel units in each segment.

Unless annotated otherwise, all primary stressed segments are assumed to have the primary stress on the first vowel unit and no other stressed unit. 
Unless annotated otherwise, all secondary stressed segments are assumed to have the secondary stress on the first vowel unit and no other stressed unit. 

Further stressed units in each segment are indicated with the symbol `\st{a}' for primary stress and `\stst{a}' for secondary stress under the first letter of the vowel group. The use of the symbol `st{}' removes any other primary stress in the segment (only one primary stress per segment, unless all of them are annotated). For instance, ag\st{o} we have that `a' is unstressed while `o' is stressed. The annotation for the secondary stress `\stst{a}' does not remove any primary stress we could have previously or any other annotated secondary stress. For instance, `complic\stst{a}te' introduces a secondary stress for the `a' but does not remove the primary stress for the `o'. That means that we do not need to annotate that like `c\st{o}mplic\stst{a}te'. We can have several secondary stresses, such as in `\iot{e}l\stst{e}ctr\nat{o}\pln{e}nc\st{e}phalogr\stst{a}m'

Examples of primary stressed vowel units are: plain (ai), ass\st{u}me (u), difference (i), science (i), r\iot{e}l\st{a}t\iot{e}d (a), ag\st{o} (o), pol\st{\brd{\i}}ce\sel{}man (i), cons\st{\pln{e}}c\udp{u}tive (first `e').

Examples of secondary stressed units: claustroph\stst{o}bic (primary: o, secondary: au), kangar\stst{o}o (primary: oo, secondary: a), softw\stst{a}re (primary: o, secondary: a), t\pln{e}leph\stst{o}ne (primary: e, secondary: o), 

When there are at least two syllables (vowel units) on the left of a primary stress, and there are no annotated secondary stresss, we assume a secondary stress two syllable (vowel units) on the left.

Examples of inferred secondary stress unit: `pa\co{r}ab\st{\pln{o}}lic' (primary: o, secondary: first a), underg\st{o} (primary: o, secondary: u), incred\st{u}lity (primary: u, secondary: first i), `incons\st{i}stent' (primary: second i, secondary: first i), `co\iot{e}x\st{i}st' (primary: i, secondary: o), reconstr\st{u}ct (primary: u, secondary: e), Europ\st{\nat{e}}an (primary: second e, secondary: eu), servi\st{e}tte (primary: secon e, secondary: first e). This is not applied if there are less than two syllables on the left (e.g. ag\st{o} where a is not stressed, \stst{u}nd\st{\rnd{o}}).  If we place the secondary stress elsewhere, the rule is broken: `\iot{e}l\stst{e}ctr\pln{o}l\cnt{y}\no{s}i\no{s}' (primary: o, secondary: second e), att\stst{e}nd\st{e}e (primary: second e, secondary: first e). Note that if we want to indicate two secondary stresses, the rule is broken and we need to mark everything, as in `\stst{u}n\stst{a}mb\st{\pln{\i}}guous'.

Finally, any vowel unit before a consonant group which contains the annotations `\sno{}', `\hno{}', `\svo{}', or `\hvo{}' is assumed to have a primary stress, unless the primary stress is annotated elsewhere. If there are more than one consonant group with these annotation, the rule is only applied for the rightmost.

Examples of inferred primary stress unit: `compre\sno{\group{ssi}}on' (primary: e), `tr\pln{a}nsmi\sno{\group{ssi}}on' (primary: first i), `illu\svo{si}on'. Note that any primary stress at any other location breaks the rule: `ma\sno{ch}\st{\brd{\i}}ne', \vo{W}\st{i}ttgen\sno{s}t\iidp{e\i}n.

Note that the inferred primary stress and the inferred secondary stress can be combined.

Examples of inferred primary stress unit and inferred secondary stress unit: `t\pln{e}l\iot{e}v\pln{\i}\svo{si}on' (primary: first i, secondary: first e), `informa\sno{ti}on' (primary: a, secondary: first i), `d\pln{e}fin\pln{\i}\sno{ti}on' (primary: second i, secondary: first e), `declara\sno{ti}on' (primary: second a, secondary: e), pr\pln{e}senta\sno{ti}on (primary: a, secondary: first e), `communica\sno{ti}on' (primary: a, secondary: u), `pronuncia\sno{ti}on' (primary: a, secondary: u),  `ini\sno{t}ia\sno{ti}on' (primary: a, secondary: second i), `varia\sno{ti}on' (primary: second a, secondary: first a). Note how important it is in these two last examples to determine the number of vowel units since the units in one word are (i,n,i,\sno{t},i,a,\sno{ti},o,n', so it is the second `i' which is two syllables on the left from the stressed vowel unit `a', and the units in the the other word are (v,a,r,i,a,\sno{ti},o,n).

Note that if the secondary stress rule is broken, the primary stress rule can be maintained: r\stst{\pln{e}}pr\iot{e}senta\sno{ti}on (primary: a, secondary: first e), d\stst{i}ffer\pln{e}n\sno{t}ia\sno{ti}on (primary: a, secondary: first i), p\stst{u}rifica\sno{ti}on.

Finally, these rules can be applied with the simple separator `\se{}', but not with the other separators affecting stress.

\subsection{Categorise Vowel Occurrences}

The good thing about this moment of the process is that we can analyse each vowel unit independently. The only contextual information we need to know is whether or not the vowel unit is stressed, whether it is at the end of a word segment, whether it is followed by an `r' and whether it is followed by a simple or complex consonant group (note that we talk about groups and not units for consonants, since consonant groups have not been treated yet) and whether the vowel is followed or not by another vowel group.

These situations will allow us to categorise vowel occurrence under three dimensions: unstressed/stressed, natural/plain (only for stressed single vowels) and rhotic/non-rhotic.

The categorisation between {\bf unstressed} and {\bf stressed} vowel units just inherits the rules for stress processed before. 
Primary and secondary stressed units are considered stressed units. All the other units become unstressed. For instance, in `science', we have a stressed unit `i' and an unstressed unit `e'. Other examples where we show the stressed units: `nucl\iot{e}ar' (u), `f\bbrd{\i{}e}ld' (ie), `s\brd{u}pra\st{i}ng\w{u}inal' (first `u' and first `i'), `v\nat{e}\si{h}icle' (\nat{e}), `Hallow\st{e}en' (first e and a), `claustroph\st{o}bic' (o, au), `k\st{a}ngar\stst{o}o' (oo, a), softw\nat{a}re (o), t\st{\pln{e}}l\iot{e}ph\stst{o}ne (e, o).

The next issue is to distinguish between {\em natural} and {\em plain} occurrences. This only applies to stressed vowel units of one single vowel, i.e. (a, e, i/y, o, u/w), either annotated or not. In fact, this applies to positions, rather than to vowels.
The rules are basically the same as those which are typically learned when spelling English, known as the ``double-consonant rule''.

The first thing to do is  to distinguish between single and complex consonant groups.
Single consonants groups are  $b$, $c$, $d$, $f$, $g$, $h$, $j$, $k$, $l$, $m$, $n$, $p$, $r$, $s$, $t$, $z$ (all single consonants except $x$)  and the groups $\#l$ and $\#r$ (i.e., consonant + $l$/$r$). It is also single when $y$ and $w$ are consonants (e.g. `yoyo', `k\brd{\i}w\brd{\i}'). The annotated double `r' (\co{r}), as in `pha\co{r}ynx' or `pa\co{r}ab\st{\pln{o}}lic', is considered a complex consonant. Note that $q$ is typically followed by a semiconsonantic $u$ which makes it complex, as well as groups `gua', `guo', `guu', but not `gue', `gui' unless the `u' is  annotated as `\w{u}'. Some examples of single consonants are `pod` (p, d), `ale` (l), `hole` (h, l), `able' (bl), `metre' (m, tr), `m\pln{a}cro' (m, cr), `micro' (m, cr), Apr\cnt{\i}l (pr, l), `vogue'  (g), '\nat{o}p\st{a}q\si{u}e' (p,q) (in both cases the u is silent). Note that this also applies when a consonant has been removed, as in `Mic\si{h}a\si{e}l' (c\si{h}) (or `pl\pln{u}m\si{b}er', where we need to annotate to avoid the `plume' sound), since consonant groups are formed after the application of `\si{ }'. All the other consonant groups are considered complex ($th$ as in `then', $ck$ as in `check', $st$ as in `lost', $qu$ as in `aqua',  $phth$ as in `diph\no{th}ong'). Some examples are `temple' (mpl), `checking' (ch, ck, ng), tr\nat{o}phy (ph), graphic (ph), `b\nat{a}the' (th), `lossy' (ss), (`\nat{e}qual') (qu, which is kw at this moment of the process). Note that $gh$ is always considered complex when it is not silent `r\si{o}u\co{gh}'. When it is silent, it is not considered simple or double, but has a very especial implication. As we will see, it automatically makes the previous vowel (if single) a natural vowel, so we consider `hi', `high', `night' and `h\si{e}ight' equivalent. Consequently, we do not annotate `tighten'.

Other annotations may affect the appraisal of simple or complex consonant groups. For instance, semiconsonants $y$ and $w$ are always reckoned as a consonant, so `n\y{i}' in `bun\y{i}on' is now considered a complex consonant). Similarly, `n\y{}' in
`la\no{s}\st{a}\si{g}n\y{}a' is a complex group.
The consonant groups which include  \hvo{di}, \hvo{de}, \sno{ti}, etc., as in proc\st{e}\hvo{d}ure or no\sno{ti}on, are counted as the consonant letters they have inside. For instance, no\sno{ti}on is single, while mi\sno{ssi}on is complex.

From the notion of single/complex consonant group, we can easily derive the notions of {\em natural} and {\em plain} position.
A {\bf natural} position of a stressed single vowel is given when: (1) it appears at the end of a word segment, (2) it is followed by another vowel unit, (3) it is followed by a single consonant group plus a vowel group, or (4) it is followed by a silent `gh' at any position. Examples: s{\em o}, m{\em e}, m{\em y}, h{\em i}gh, n{\em i}ght (note that `gh' disappeared in previous steps of the process), t{\em o}e, g{\em i}ant, p{\em o}\iot{e}t, m{\em i}cro, pl{\em a}te, ph{\em o}ny, d{\em i}no\no{s}aur, {\em a}ble, {\em i}dle, n{\em u}cl\iot{e}ar, v{\em a}gue. All the other stressed positions are {\bf plain} positions. Examples: s{\em a}ffron, l{\em i}ttle, tw{\em e}nty, {\em a}qua, t{\em e}mple, ch{\em e}cking, gr{\em a}phic (ph), pha\co{r}ynx, t\si{o}u\co{gh}er. 
Note that for non-stressed vowels, this does not apply, so we have to annotate `fortn\nat{\i}ght', `\nat{\i}d\st{e}ntity', etc.
In general, we will use the terms ``natural" and ``plain" vowels to refer to vowels in natural and plain positions respectively.

Finally, (single or complex) vowel units can be {\bf rhotic} or {\bf non-rhotic}\footnote{Note that these terms are related but different to their application for dialects, as rhotic and non-rhotic accents.}.
If the vowel is not followed by an `r' then it is clearly non-rhotic, as in `fate', `fat' or `hungry'. 
Unstressed single vowels are considered non-rhotic (`winner', `\iot{e}r\st{a}dic\nat{a}te', `\sno{s}ugar', `lantern', `player'),
For the rest of combinations (stressed single vowels and stressed/unstressed complex vowel units), the rule is as follows: a vowel unit is rhotic if it is followed by a single (non-doubled) `r' (in the segment)  (`fair', `laird', `fairy', 'ear\sel{}ring', `card', `far', `norm', `first', `her', `fir\sel{}ry',
`pray\si{e}r', `may\si{o}r', but not `player') or double `rr'  not followed by a vowel (`err', `whirr'). It is non-rhotic elsewhere. Explicitly, it is non-rhotic when it is not followed by an `r' (mail, now, glade, mode, pad, fitted), is followed by a double rr + consonant (carry, berry, lorry, hurry) or it is followed by `\co{r}' (mi\co{r}\iot{a}cle).

Since unstressed vowel units are always non-rhotic, there can be secondary stress annotations to force a vowel to become rhotic, as in `\stst{i}r\st{\nat{o}}nic', `\stst{u}rb\st{a}ne', `no\sel{}wh\brd{e}re' even though these words are not typically transcribed in IPA with the secondary stress.

Note that the pronunciation varies (and it becomes rhotic or not) for RP and GA. For instance, the (non-annotated) words `carry', `berry', `stirring', `lorry', `furry', `hurry') are only pronounced similarly for `berry', `stirring' and `furry', with annotations `berry', `stir\se{}ring' and `fur\se{}ry'. In some American accents, it also happens for `carry'. In other case, both the pronunciation and the annotation are different, as in `hurry' (RP) and `hur\se{}ry' (GA), and `lorry' (RP) and `lor\se{}ry'. Other cases the same annotation is useful for both RP and GA, even though the pronunciation is different. For instance, laureate is annotated as `l\oopq{au}\co{r}\iot{e}ate', which in both RP and GA distinguishes from the sound of `for'.
Table \ref{tab:rhotic} summarises the possible combinations (the annotations will be better understood in the following section).

\subsection{Evaluate Vowel Units}

Evaluating vowel units is the most complex part of the process, given the great variety of vowel sounds in English. Even though we have to consider every possible case, most cases can simply be handled by a subset of the following set of rules and correspondences.

We will first evaluate vowel units with one vowel (both stressed and unstressed, rhotic and non-rhotic) and then we will evaluate vowel units with two vowels (both stressed and unstressed, rhotic and non-rhotic).

\subsubsection{Vowel Units with One Vowel}.

First we will deal with vowel units with only one vowel (i.e, $a$, $e$, $i$, $y$, $o$, $u$, and $w$). We distinguish three cases: unstressed, natural position and plain position.

Non-annotated unstressed vowel units with only one vowel are pronounced as shown in Table \ref{tab:unstressedvowel}. Note that the sound is independent of whether it is followed by an `r', because non-annotated unstressed vowels are always non-rhotic. That means that `ir\pln{\i}dium' makes a \textipa{[Ir-]} sound for the first `i' (not \textipa{[3r-]} or \textipa{[I@r-]}). This is handy in GA, since `menhir' would make \textipa{[-Ir]}, while we need to annotate it for RP as `menh\brd{\i}r' to make \textipa{[-I@r]}, which also works for GA. Note that we have to annotate \iot{e}l\pln{\i}x\cnt{\i}r (or \iot{e}l\pln{\i}x\si{\i}r), and kaff\cnt{i}r to make the sound \textipa{[-@r]}. For s\rnd{o}\si{u}ven\stst{\brd{\i}}r, it also requires an annotation because it is stressed (and then it is rhotic and it sounds \textipa{[-I@r]}.

\begin{table}
\begin{center}
{\small
\begin{tabular}{|| c | c | p{3cm} | p{3cm}  | p{3cm}  ||}
\hline
\bf{Unit} & \bf{IPA} & \bf{Examples} & \bf{Negative annotated cases} & \bf{Comments} \\ \hline \hline

$a$ & \textipa{[@]}	 & {\em a}b\clr{o}ve, \nat{e}qu{\em a}l, Port\nat{u}g{\em a}l, li{\em a}r, {\em a}rabi{\em a}n, b\opq{u}lw{\em a}rk, burgl{\em a}r & {\em \rnd{a}}lth\nnat{\st{o}u}gh, {\em \brd{a}}rt\st{i}stic, mess{\em \iot{a}}ge, \pln{a}dequ{\em a}te & \\\hline

final $e$ & silent	& mal{\em e}, ti{\em e}, pal{\em e}\sel{}ness, \iidp{ey}{\em e}, they'v{\em e}, metr{\em e}, \pln{a}dequat{\em e}  & cat\st{a}stroph{\em \iot{e}} & Note how compound words are segmented to apply the rule \\\hline

$e$ before final $d$ & silent	& robb{\em e}d, stuff{\em e}\no{d} & nak{\em \iot{e}}d, overf\stst{{\em e}}d & \\\hline

$e$ before final $s$ & silent	& rul{\em e}s, ladi{\em e}s & coach{\em \iot{e}}s, thes{\em \nat{e}}s & \\\hline

$e$ (elsewhere) & \textipa{[@]}	 & di{\em e}t, \iot{e}l\st{\pln{e}}v{\em e}n,  b\pln{a}rr{\em e}l, ladd{\em e}r, sil{\em e}nce, B{\em e}rl\st{i}n, \pln{a}d{\em e}quate  & b{\em \iot{e}}l\st{o}w, {\em \iot{e}}l\st{\pln{e}}ven, chick{\em \iot{e}}n, r\stst{{\em e}}m\st{a}ke, {\em \iot{e}}r\st{a}dic\stst{a}te  & \\\hline

$i$, $y$ & \textipa{[i]}	 & ind{\em i}go, pr\iot{e}tt{\em y}, spag\si{h}\st{e}tti, {\em i}mpl\st{y}, sod{\em i}um, insp{\em i}r\st{a}tion & al\st{u}mn{\em \nat{\i}}, d{\em \nat{\i}}gr\st{e}ss, fak{\em\brd{\i}}r, menh\brd{\em \i}r, \iot{e}l\st{i}x\cnt{\em \i}r, kaff\cnt{\em \i}r  & \\\hline

final $o$ & \textipa{[oU]}	& carg{\em o}, potat{\em o}, radi{\em o}, turb{\em o}\sel{}j\pln{e}t / turb{\em o}\selr{}jet, eur{\em o} & - & Note how compound words can be segmented to apply the rule \\\hline

$o$ before final $s$/$es$ & \textipa{[oU]}	& tang{\em o}s, potat{\em o}es, eur{\em o}s & c\si{h}a\pln{o}\no{s}   & \\\hline

$o$ (elsewhere) & \textipa{[@]}	 & \pln{a}t{\em o}m, c{\em o}mpl\st{y}, saffr{\em o}n, tort{\em o}\si{i}\no{s}e, lab{\em o}r, m{\em o}r\st{\pln{a}}lity (RP), tedi{\em o}us (this is addressed by the -ou- rule), Europ\st{\nat{e}}an & i{\em \pln{o}}n, c{\em \nat{o}}\st{e}rce, m{\em \brd{o}}rt\st{\pln{a}}lity, m{\em \brd{o}}r\st{\pln{a}}lity (GA) &  \\\hline

$u$ & \textipa{[@]}	 & circ{\em u}mstance, dict{\em u}m, lem{\em u}r, murm{\em u}r,  sulfph{\em u}rous, f\st{i}g{\em u}re (RP), tedio{\em u}s (this is addressed by the -ou- rule), s{\em u}sp\st{e}ct, sp\pln{a}\hno{t}{\em u}la & Bant{\em \brd{u}}, {\em \nat{u}}n\st{\brd{\i}}q\si{u}e, Port{\em \udp{u}}gal, {\pln {\em u}}nd\st{\rnd{o}}, {\em \udp{u}}r\st{a}nium, {\em \stst{u}}rb\st{a}ne, fail\y{}{\em u}re / fail{\em \idp{u}}re, t\pln{e}n{\em \idp{u}}re, f\st{\pln{\i}}g\y{}{\em u}re (GA) / f\st{\pln{\i}}g{\em \udp{u}}re (GA), j{\em \opq{u}}r\st{a}ssic  &   \\\hline

\end{tabular}
} % small
\vspace{5mm}
\caption{Pronunciation of non-annotated unstressed vowel units with only one vowel}
\label{tab:unstressedvowel} 
\end{center}
\end{table}

Since \ref{tab:unstressedvowel} introduces some silent sounds for some vowels, it is only at this moment of the process that  we know how many syllables a word has. It is just the number of vowel units which are not silent.

From here we can define the sounds of stressed and/or annotated vowel units with only one vowel as Table \ref{tab:stressedvowelA} and Table \ref{tab:stressedvowelB} show\footnote{The appearance of the symbol `j' in the pronunciation of the vowel `u' depends on the previous consonant and on the English accent. This will be explained below.}. The annotations apply to a single letter, but most of them can be extended to more than one letter with the following rule: if an annotated vowel is followed by a vowel which has been crossed out, then we can equivalently write the annotation spreading over both vowel. For instance: 

br\nat{o}\si{o}ch = br\nnat{oo}ch, br\brd{e}\si{a}k = br\bbrd{ea}k, bl\clr{o}\si{o}d = bl\cclr{oo}d

\begin{table}
\begin{center}
{\small
\begin{tabular}{|| c | p{2cm} | c | c | p{7cm}   ||}
\hline
\bf{Class} & \bf{Unit} & \bf{IPA} & rhotic & \bf{Examples}  \\ \hline \hline
\multirow{5}{*}{Natural} & 

{\em \nat{a}, \nnat{a\textasteriskcentered}}, natural {\em a }             & \textipa{[eI]} & \textipa{[EK]} & m{\em a}te, \sno{ch}amp{\st{\em a}}\si{g}ne, h{\em \nat{a}}\no{s}\si{t}en, r{\em \nat{a}}nge, g{\em a}\si{u}ge, b{\em \nat{a}}ss, C{\em \nat{a}}mbridge, {\em a}\co{c}\si{h}e, membr{\em a}ne , f{\em a}re, v{\em a}ria\sno{ti}on \\\cline{2-5}

& {\em \nat{e}, \nnat{e\textasteriskcentered}}, natural {\em e }             & \textipa{[i:]} & \textipa{[IK]} & b{\em e}, m{\em e}, sc{\em e}ne, k{\em e}\si{y}, h{\em e}'s, prot{\em \nnat{e\i}}n, \nat{e}qual, m{\em e}re  \\\cline{2-5}

& {\em \nat{\i}/\nat{y}, \nnat{\i\textasteriskcentered}/\nnat{y\textasteriskcentered}}, natural {\em i/y } & \textipa{[aI]} & \textipa{[aIK]} & b{\em i}te, m{\em y}, gu{\em y}, h{\em i}, t{\em i}ghten, h{\em i}gh, n{\em i}ght, m{\em \nat{\i}}nd, {\em I}'m, {\em \nat{\i}}d\st{e}ntity, \brd{a}rthr{\em \st{i}}ti\no{s}, w{\em i}-f{\em i}, f{\em i}re, mob{\em \nat{\i}}le (RP), fortn\nat{\i}ght   \\\cline{2-5}

& {\em \nat{o}, \nnat{o\textasteriskcentered}}, natural {\em o }             & \textipa{[oU]} & \textipa{[O\*r]} & s{\em o}, g{\em o}, t{\em o}ne, d{\em \nat{o}}n't, m{\em \nat{o}}st, br{\em \nat{o}}\si{o}ch = br\nnat{{\em oo}}ch, pr{\em o}t\nnat{e\i}n, c{\em \nat{o}}\st{e}rce, m{\em o}re, y{\em o}\si{u}'re (RP), tho\si{u}gh  \\\cline{2-5}

& {\em \nat{u}/\nat{w}, \nnat{u\textasteriskcentered}/\nnat{w\textasteriskcentered}}, natural {\em u/w }             & \textipa{[(j)u:]} & \textipa{[(j)UK]} & y\si{o}{\em u}, m{\em u}te, {\em \nat{u}}n\st{\brd{\i}}q\si{u}e, Nept{\em u}ne, c{\em u}re, r{\em u}le, fr{\em u}\si{i}t, y\si{o}{\em u}'re (GA), tr\si{o}\nat{u}gh\st{o}ut   \\\cline{2-5}

\hline

\multirow{5}{*}{Plain} & 
{\em \pln{a}, \ppln{a\textasteriskcentered}}, plain {\em a }              & \textipa{[\ae]} & \textipa{[A\*r]} & c{\em a}t, t{\em a}xi, b{\em \pln{a}}lance,  f{\em a}r, h{\em a}v\si{e}, n{\em \pln{a}}\sno{ti}onal, Sp{\em \pln{a}}nish, {\em a}r\si{e}  \\\cline{2-5}

& {\em \pln{e}, \ppln{e\textasteriskcentered}}, plain {\em e }              & \textipa{[E]} & \textipa{[3\*r]} & p{\em e}t, m{\em e}ss, sc{\em e}nt, pr{\em \pln{e}}sent, y\pln{\em e}a\si{h}, h{\em e}r, w{\em e}r\si{e}   \\\cline{2-5}

& {\em \pln{\i}/\pln{y}, \ppln{\i\textasteriskcentered}/\ppln{y\textasteriskcentered}}, plain {\em i/y }  & \textipa{[I]} & \textipa{[3\*r]} & p{\em i}t, l{\em y}nx, p{\em i}cking, m{\em i}rror, m{\em i}\co{r}\iot{a}cle, p{\em {\pln{\i}}}ty, s{\em {\pln{\i}}}\si{e}ve, f{\em i}rst, wh{\em i}r\si{r}  \\\cline{2-5}

& {\em \pln{o}, \ppln{o\textasteriskcentered}}, plain {\em o }              & \textipa{[6]} & \textipa{[O\*r]} & p{\em o}t, b{\em o}ttle, b{\em o}nd,  g{\em o}n\si{e}, r{\em \pln{o}}bin, co\si{u}\no{gh}, f{\em o}r, s{\em o}rt, y{\em o}\si{u}r (RP)
  \\\cline{2-5}

& {\em \pln{u}/\pln{w}, \ppln{u\textasteriskcentered}/\ppln{w\textasteriskcentered}}, plain {\em u/w }   & \textipa{[2]} & \textipa{[3\*r]} & c{\em u}t, b{\em u}bble, d{\em u}st, t\si{o}{\em u}ch, c\si{o}{\em \pln{u}}s\cnt{i}n, {\em \pln{u}}gly, f{\em u}r, \stst{\em u}rb\st{a}ne   \\\cline{2-5}

\hline

\multirow{5}{*}{Broad} & 
{\em \brd{a}, \bbrd{a\textasteriskcentered}}  & \textipa{[A:]}   & \textipa{[A\*r]} & f{\em \brd{a}}ther, l{\em \brd{a}}ma, {\em \brd{a}}h, Sha{\em \brd{a}}h  \\\cline{2-5}

& {\em \brd{e}, \bbrd{e\textasteriskcentered}}  & \textipa{[eI]}   & \textipa{[3K]} & c\pln{a}f{\em \brd{e}}, br{\em \brd{e}}\si{a}k = br{\em \bbrd{ea}}k, {\em \brd{e}}h, th{\em \brd{e}}re, w{\em \bbrd{ea}}r, d\pln{o}ssi{\em \brd{e}}\si{r}  \\\cline{2-5}

& {\em \brd{\i}/\brd{y}, \bbrd{\i\textasteriskcentered}/\bbrd{y\textasteriskcentered}}  & \textipa{[i:]} & \textipa{[IK]} & ma\sno{ch}{\st{\brd{\em \i}}}ne, p{\brd{\em \i}}\hno{zz}a, 
sk\brd{\em\i}ing, \nat{u}n{\st{\brd{\em \i}}}q\si{u}e, t\bbrd{{\em \i}e}r, \pln{e}m\st{\brd{\em \i}}r, q\si{u}\si{a}\brd{\em y},  p\bbrd{{\em \i}e}r, menh\brd{\em \i}r, Z\brd{a}{\st{\brd{\em \i}}}re  \\\cline{2-5}

& {\em \brd{o}, \bbrd{o\textasteriskcentered}}  & \textipa{[O:]}   & \textipa{[O\*r]} & b\bbrd{ou}ght, br\bbrd{oa}d  \\\cline{2-5}

& {\em \brd{u}/\brd{w}, \bbrd{u\textasteriskcentered}/\bbrd{w\textasteriskcentered}}  & \textipa{[u:]} & \textipa{[UK]} & s{\em \brd{u}}per, g{\em \brd{u}}r{\em \brd{u}}, Bant{\em \brd{u}}, r{\em \brd{u}}\si{h}r, c{\em \brd{w}}m, t{\em \bbrd{wo}}  \\\cline{2-5}

\hline

\multirow{5}{*}{i-Diphthong} & 
{\em \idp{a}, \iidp{a\textasteriskcentered}}  & \textipa{[aI]}  & \textipa{[aIK]}  & p\idp{a}\brd{e}lla, m\iidp{ae}stro, D\pln{a}l\iidp{a\i}, D\brd{u}b{\iidp{\st{a}\i}}  \\\cline{2-5}

& {\em \idp{e}, \iidp{e\textasteriskcentered}}  & \textipa{[aI]}  & \textipa{[aIK]}  & \iidp{ey}e, h\iidp{e\i}ght (but h\si{e}ight is preferred), \iidp{e\i}ther (RP),  \iidp{E\i}nst\iidp{e\i}n, Fa\si{h}\co{r}enh\iidp{e\i}t,   \\\cline{2-5}

& {\em \idp{\i}, \iidp{\i\textasteriskcentered}}  &               &                    & {\em Not used}   \\\cline{2-5}

& {\em \idp{o}, \iidp{o\textasteriskcentered}}  & \textipa{[wA:]}  & \textipa{[wA:\*r]} & f\iidp{o\i}e, p\pln{a}t\iidp{o\i}\si{s}, mem\iidp{o\i}r   \\\cline{2-5}

& {\em \idp{u}, \iidp{u\textasteriskcentered}}  & \textipa{[j@]}     &  \textipa{[j@\*r]}                  & fail\idp{u}re, ten\idp{u}re, amm\idp{u}n\pln{\i}\sno{ti}on (GA), Merc\idp{u}ry (GA), cell\idp{u}lar (GA), tub\idp{u}lar (GA), curr\st{\pln{\i}}c\idp{u}lum (GA), occ\idp{u}p\stst{y} {GA}     \\\cline{2-5}

\hline

\multirow{5}{*}{u-Diphthong} & 
{\em \udp{a}, \uudp{a\textasteriskcentered}}  & \textipa{[aU]}  & \textipa{[aUK]}  & t\uudp{au}, L\uudp{ao}s  \\\cline{2-5}

& {\em \udp{e}, \uudp{e\textasteriskcentered}}  & \textipa{[OI]}  & \textipa{[OIK]} & Fr\uudp{eu}dian  \\\cline{2-5}

& {\em \udp{\i}, \uudp{\i\textasteriskcentered}}  &               &                   & {\em Not used}   \\\cline{2-5}

& {\em \udp{o}, \uudp{o\textasteriskcentered}}  & \textipa{[aU]}  & \textipa{[aUK]}  &  n\uudp{ow}, northb\uudp{ou}nd   \\\cline{2-5}

& {\em \udp{u}, \uudp{u\textasteriskcentered}}  & \textipa{[jU]}  & \textipa{[jU\*r]} &  \udp{u}r\st{a}nium, f\pln{\i}g\udp{u}re (GA), s\pln{\i}t\udp{u}a\sno{ti}on (RP) (it can also be annotated as s\pln{\i}t\y{}\opq{u}a\sno{ti}on (RP), note that in GA it is `s\pln{\i}\hno{t}\opq{u}a\sno{ti}on'), sens\udp{u}al (RP) (in GA it is `sen\svo{s}\opq{u}al'), amm\udp{u}n\pln{\i}\sno{ti}on (RP), Merc\udp{u}ry (RP), cell\udp{u}lar (RP), tub\udp{u}lar (RP), curr\st{\pln{\i}}cu\udp{u}lum (RP), occ\udp{u}p\stst{y} {RP}    \\\cline{2-5}
\hline
\end{tabular}
} %small
\vspace{5mm}
\caption{Pronunciation of stressed and/or annotated vowel units with one vowel (Part 1/2)}
\label{tab:stressedvowelA} 
\end{center}
\end{table}

\begin{table}
\begin{center}
{\small
\begin{tabular}{|| c | c | c | c | p{7cm}   ||}
\hline
\bf{Class} & \bf{Unit} & \bf{IPA} & rhotic & \bf{Examples}  \\ \hline \hline
\multirow{5}{*}{Clear} & 
{\em \clr{a}, \cclr{a\textasteriskcentered}}  & \textipa{[2]}   &  & M{\em \clr{a}}g\si{h}reb \\\cline{2-5}
 & {\em \clr{e}, \cclr{e\textasteriskcentered}}  & \textipa{[\ae]}   & \textipa{[A\*r]} & s{\em \clr{e}}rge\si{a}nt, cl{\em \clr{e}}rk [RP] \\\cline{2-5}
 & {\em \clr{\i}, \cclr{\i\textasteriskcentered}}  & \textipa{[\ae]} &  & mer{\em \st{\clr{\i}}}n\co{g}\si{u}e, l{\em \clr{\i}}n\svo{g}erie \\\cline{2-5}
 & {\em \clr{o}, \cclr{o\textasteriskcentered}}  & \textipa{[2]}   &  & s{\em \clr{o}}me, l{\em \clr{o}}ve, d{\em \clr{o}}es, m{\em \clr{o}}ney, bl{\em \clr{o}}\si{o}d = bl{\em \cclr{oo}}d, b{\em \clr{o}}rough (RP) \\\cline{2-5}
 & {\em \clr{u}, \cclr{u\textasteriskcentered}}  &  &  & {\em Not used} \\\cline{2-5}

\hline

\multirow{5}{*}{Central} & 
{\em \cnt{a}}  & \textipa{[E]}   &   & {\em \cnt{a}}ny, m{\em \cnt{a}}ny, ag{\em \st{\cnt{a}}}\si{i}n, s{\em \cnt{a}}\si{i}d \\\cline{2-5}
 & {\em \cnt{e}}  & \textipa{[@]}   &  & G\cnt{o}\si{e}t\si{h}{\em \cnt{e}}, \svo{g}\opq{e}nr{\em \cnt{e}}  \\\cline{2-5}
 & {\em \cnt{\i}}  & \textipa{[@]} & \textipa{[@\*r]} & ev{\em \cnt{\i}}l, penc{\em \cnt{\i}}l, c\si{o}\pln{u}s{\em \cnt{i}}n,  \iot{e}l\st{\pln{\i}}x{\em \cnt{\i}}r, kaff{\em \cnt{\i}}r, terr{\em \cnt{\i}}ble, mo{\em \cnt{\i}}le    \\\cline{2-5}
 & {\em \cnt{o}}  & \textipa{[3]}   & \textipa{[3\*r]} & M{\em \cnt{o}}bius, G{\em \cnt{o}}\si{e}t\si{h}\cnt{e}, w{\em \cnt{o}}rd, w{\em \cnt{o}}rst,   b{\em \cnt{o}}r\nat{ou}gh (GA) \\\cline{2-5}
 & {\em \cnt{u}}  & \textipa{[E]} &  & b{\em \cnt{u}}\co{r}y \\\cline{2-5}

\hline

\multirow{5}{*}{Iotted} & 
{\em \iot{a}}  & \textipa{[I]}   &  & o\co{r}{\em \iot{a}}nge, mess{\em \iot{a}}ge, cabb{\em \iot{a}}ge \\\cline{2-5}
 & {\em \iot{e}}  & \textipa{[I]}   &  & {\em \iot{E}}nglish, pr{\em \iot{e}}tty, end{\em \iot{e}}d, w\iot{o}m{\em \iot{e}}n, Ka\co{r}at{\em \iot{e}}  \\\cline{2-5}
 & {\em \iot{\i}}  &    &  & {\em undistinguishable from i} \\\cline{2-5}
 & {\em \iot{o}}  & \textipa{[I]}   &  & w{\em \iot{o}}m\iot{e}n \\\cline{2-5}
 & {\em \iot{u}}  & \textipa{[I]}   &  & b{\em \iot{u}}sy, m\pln{\i}n{\em \iot{u}}te, lett{\em \iot{u}}ce \\\cline{2-5}

\hline

\multirow{5}{*}{Rounded} &
{\em \rnd{a}}  & \textipa{[O:]}  & \textipa{[O\*r]}  & {\em \rnd{a}}ll, c{\em \rnd{a}}ll, w{\em \rnd{a}}\si{l}k, {\em \rnd{a}}lso, w{\em \rnd{a}}ter, w{\em \rnd{a}}r, Ut{\em \rnd{a}}\si{h} \\\cline{2-5}
 & {\em \rnd{e}}  & \textipa{[oU]}  &   &  s{\em \rnd{e}}\si{w}, s{\em \rnd{e}}\si{w}n \\\cline{2-5}
 & {\em \rnd{\i}}  &  &   &  {\em Not used} \\\cline{2-5}
 & {\em \rnd{o}}  & \textipa{[u:]}  &  &  t{\em \rnd{o}}, d{\em \rnd{o}}, sh{\em \rnd{o}}e, l{\em \rnd{o}}se, r{\em \rnd{o}}\si{u}te \\\cline{2-5}
 & {\em \rnd{u}}  & \textipa{[oU]}  & \textipa{[O\*r]}  &  m\si{a}{\em \rnd{u}}ve, \si{a}{\em \rnd{u}}ber\svo{g}\brd{\i}ne, \sno{ch}\si{a}\rnd{u}ffe\si{u}r, bur\si{e}\si{a}\rnd{u}\vo{x}, \sno{s}\rnd{u}re (RP) \\\cline{2-5}

\hline

\multirow{5}{*}{Opaque} &
{\em \opq{a}, \oopq{a\textasteriskcentered}}  & \textipa{[O]}  &   & w{\em \opq{a}}tch, wh{\em \opq{a}}t, w{\em \opq{a}}s, {\em \oopq{au}}\vo{ss}ie, y{\em \opq{a}}\si{c}\si{h}t, w\opq{a}rrant, l\oopq{au}\co{r}\iot{e}ate, resta\si{u}r\opq{a}nt (GA, formal), resta\si{u}r\opq{a}\co{n}\si{t} (RP, formal) \\\cline{2-5}
 & {\em \opq{e}, \oopq{e\textasteriskcentered}}  & \textipa{[O]}  &   & {\em \st{\opq{e}}}nvel\stst{o}pe, {\em \opq{e}}n r{\rnd{o}\si{u}}te, \svo{g}{\em \opq{e}}nr\cnt{e} \\\cline{2-5}
 & {\em \opq{\i}, \oopq{\i\textasteriskcentered}}  & \textipa{[O]} &   & l{\em \opq{\i}}n\svo{g}er\si{i}\brd{e} (GA)  \\\cline{2-5}
 & {\em \opq{o}, \oopq{o\textasteriskcentered}}  & \textipa{[U]}  & \textipa{[UK]} & w{\em \opq{o}}man, w{\em \opq{o}}lf, g{\em \oopq{oo}}d, w{\em \oopq{ou}}\si{l}d, c{\em \oopq{ou}}\si{l}d, t{\em \oopq{ou}}r, y{\em \oopq{ou}}r (GA)  \\\cline{2-5}
 & {\em \opq{u}, \oopq{u\textasteriskcentered}}  & \textipa{[U]}  &   & p{\em \opq{u}}t, p{\em \opq{u}}sh, Q{\em \opq{u}}r\st{\brd{a}}n, J{\em \opq{u}}r\st{a}ssic  \\\hline

\end{tabular}
} %small
\vspace{5mm}
\caption{Pronunciation of stressed vowel and/or annotated units with one vowel (Part 2/2)}
\label{tab:stressedvowelB} 
\end{center}
\end{table}

\noindent Many of the combinations with one letter are not very frequently used, many others (especially those with two letters or in rhotic positions) are never used. They are left blank in the tables.

\subsubsection{Vowel Units with Two Vowels}.

Now we address the sounds of vowel units with two vowels (for both stressed and unstressed).

The rule seen above that an annotation which spreads over two vowels is equivalent to the annotation of the first vowel and the deletion of the second vowel is very useful here. This is especially useful for diphthongs: \iidp{ae}, \iidp{ay}, \uudp{ow}, \uudp{eu}, etc., and other digraphs \oopq{oo}, which can be used to break the by-default rule.

In this way there is no need to determine the sounds of all the pairs with all the annotations. According to the previous rule, this {\em was} done in tables \ref{tab:stressedvowelA} and \ref{tab:stressedvowelB}.
It is only necessary to give the information that Table \ref{tab:vowels} provides\footnote{The appearance of the symbol `j' in eu/ew depends on the previous consonant and on the English accent. This will be explained below.}, which is the default sound of the English digraphs which make a unit. For completeness (this is not necessary), we include some examples of the rest of digraphs which do not make up a unit in Table \ref{tab:vowelsnotunit}.

\begin{table}
\begin{center}
{\small
\begin{tabular}{|| p{1.5cm} |  p{1.5cm} | p{1.5cm} | p{4cm} | p{4cm} ||}
\hline
\bf{Pairs} & \bf{IPA} (Stressed)   & \bf{IPA} (Unstressed) & \bf{Examples} & \bf{Annot. Exceptions} \\ \hline \hline

aa     & \textipa{[A:]}, \textipa{[A\*r]}        & \textipa{[A:]}, \textipa{[A\*r]}   & Maastri\co{ch}t, baz\st{a}ar  & \stst{i}ntra\se{}ax\st{i}llary\\\hline

ae     & \multirow{3}{*}{\textipa{[i:]}, \textipa{[IK]}}       & \multirow{2}{*}{\textipa{[i:]}, \textipa{[IK]}}   & Daedalus, an\st{a}emia, conc\si{h}ae, am\si{o}\st{e}bae,  algae, ae\st{o}lian, \co{c}\si{h}\nat{\i}m\st{a}era, coc\si{h}l\iot{e}\stst{ae}  & Mic\si{h}a\si{e}l, c\si{a}\iot{e}s\st{a}rean, \si{a}esth\st{\pln{e}}tic, \st{a}\si{e}ropl\stst{a}ne, \stst{a}\si{e}r\st{o}bics, m\iidp{ae}stro, Fa\si{e}roes \\\cline{4-5}

ee  &         &   & see, seeing, deer, dund\st{e}e, att\stst{e}nd\st{e}e, chimp\pln{a}nz\st{e}e, reindeer  & re\iot{e}x\st{a}mine\\\cline{3-5}

ea  &        & \textipa{[I@]}, \textipa{[I@\*r]}   & mean, seal, fear, rear, ann\st{e}al, fish\selr{}meal, laureate, Boolean, cereal, nuclear, l\pln{\i}near, area, cornea, coc\si{h}lea &  r\iot{e}al, \nat{\i}d\st{\iot{e}}a, y\pln{e}a\si{h}, Europ\st{\nat{e}}an, G\si{u}\pln{\i}n\nnat{ea}, spre\si{a}d, e\si{a}rly, gr\bbrd{ea}t, br\bbrd{ea}k, br\si{e}aking, cr\iot{e}\st{a}tion, w\bbrd{ea}r, O\sno{ce}an, C\iot{e}t\st{a}\sno{ce}an, coc\si{h}l\iot{e}\stst{ae} \\\hline

ai/ay  &  \multirow{2}{*}{\textipa{[eI]}, \textipa{[EK]}}       & \textipa{[eI]}, \textipa{[EK]}   & bay, bait, fair, highway, \rnd{a}lways, Friday, funfair, cocktail   &  s\cnt{a}\si{y}s, D\pln{a}l\iidp{\st{a}\i}, certa\si{i}n \\\cline{3-5}

ei/ey   &         & \textipa{[I]}, \textipa{[EK]}  & they, eight, neighbour, j\si{o}urney, h\clr{o}ney, s\pln{o}v\si{e}rei\si{g}n, forfeit, \si{h}eir, their, fo\co{r}eign &  be\sel{}ing,  b\bbrd{e\i}\co{r}\st{\opq{u}}t, prot\nat{e\i}n,  \nat{a}\no{th}\iot{e}ist, v\nat{e}\si{h}icle, s\iidp{e\i}sm\st{\pln{o}}logy, h\si{e}ight,  conc\nnat{\st{e}\i}t, perc\nnat{\st{e}\i}ve (although perc\st{e}\si{i}ve is preferred), wastew\nnat{e\i}r, w\nnat{e\i}rd  \\\hline

au/aw  & \textipa{[O:]}, \textipa{[O\*r]}        & \textipa{[O:]}, \textipa{[O\*r]}   & jaw, fawn, autom\st{\pln{a}}tic, laureate, centaur, dinosaur, Austr\st{a}lia  & m\si{a}\rnd{u}ve, ga\si{u}ge, Ha\ser{}w\iidp{a\i}i (note how the cluster `aw' is broken), \oopq{Au}str\st{a}lia (some accents), l\oopq{au}\co{r}eate (a short non-rhotic `o' sound, like in `lorry')\\\hline

eu/ew  & \textipa{[(j)u:]}, \textipa{[(j)UK]} & \textipa{[(j)u:]}, \textipa{[(j)UK]} & new, few, nephew, feudal, be\si{a}uty, euph\st{o}\co{r}ic, Euro, Europe, eur\st{e}ka  & Fr\uudp{eu}dian, clo\no{s}e\selr{}up  \\\hline

oa     &  \multirow{2}{*}{\textipa{[oU]}, \textipa{[O\*r]}}        & \multirow{2}{*}{\textipa{[oU]}, \textipa{[O\*r]}}   & loaf, cocoa, w\cnt{o}rkload, coarse, bezoar  & But: o\ser{}a\no{s}i\no{s}, br\bbrd{oa}d, N\nat{o}ah \\\cline{4-5}

ow     &        &    & know, owner, show, glow, elbow, window, blown  & n\uudp{ow}, fl\uudp{ow}er, br\uudp{ow}n \\\hline

ou     & \textipa{[aU]}, \textipa{[aUK]}      & \textipa{[@]}, \textipa{[@\*r]}    & hound, stout, flour, j\pln{e}\si{a}lous, c\pln{a}moufl\brd{a}\vo{g}e, labour,
c\clr{o}lour.   &  co\si{u}rt, s\nnat{ou}l, p\nnat{ou}ltry,  t\oopq{ou}r, r\si{o}u\co{gh}, c\si{o}\pln{u}s\cnt{i}n, c\si{o}untry, j\si{o}urney, c\si{o}u\co{r}\iot{a}ge, w\oopq{ou}\si{l}d, r\rnd{o}\si{u}te\\\hline  

oi/oy  & \textipa{[OI]}, \textipa{[OIK]}       & \textipa{[OI]}, \textipa{[OIK]}  & Boy, voice, turqu\pln{o}ise, alloy (noun), all\st{o}y (verb), moi\co{r}a, coir  &   go\sel{}ing, \co{c}\si{o}\nat{y}\st{o}t\iot{e}, \stst{e}\nat{o}z\nat{o}ic, mem\iidp{o\i}r, ch\hvo{o}\nat{\i}r.\\\hline

oo     & \textipa{[u:]}, \textipa{[UK]}       & \textipa{[u:]}, \textipa{[UK]}  & moon, moor.   & g\oopq{oo}d, bl\cclr{oo}d, do\si{o}r \\\hline

\end{tabular}
} %small
\vspace{5mm}
\caption{Pronunciation of vowel units with two vowels.}
\label{tab:vowels} 
\end{center}
\end{table}

\begin{table}
\begin{center}
{\small
\begin{tabular}{|| p{1.5cm} | p{10cm} ||}
\hline
\bf{Pairs} & \bf{Examples}  \\ \hline \hline

ao                                & kaolin, cha\pln{o}\no{s}, L\uudp{ao}\no{s}, Mac\st{\uudp{ao}}, M\st{\uudp{ao}}ism, ch\nat{a}\st{\pln{o}}tic, \nat{a}\st{o}rta, \sno{g}\nat{a}\si{o}l  \\\hline   

eo                                    & Cleop\st{\pln{a}}tra, L\st{e}ot\stst{a}rd, l\pln{e}\si{o}pard, pe\si{o}ple, e\pln{o}n,  \co{C}\si{h}e\pln{o}ps, G\iot{e}\st{o}logy, ae\st{o}lian, e\nat{o}z\st{\nat{o}}ic, Leon\st{a}rdo, \iot{e}rr\st{o}n\iot{e}ous, B\brd{e}ow\opq{u}lf, hom\iot{e}\st{\pln{o}}pathy  \\\hline

oe                                     & goes, potatoes, d\rnd{o}er, am\si{o}\st{e}bae, Bo\si{e}r, d\clr{o}es, Boe\se{}ing, co\iot{e}x\st{i}st   \\\hline

ia, ya, ie, ye, ii, yi, iy, yy, io, yo, iu, yu, iw, yw                                 & giant, via, cyan, diagr\nat{a}m, \nat{I}b\se{e}rian, Bel\hno{gi}an, phobia, science, quiet, diet, dyer, flyer, society, ambience, berries, married, f\bbrd{\i{}e}ld, b\iot{e}l\bbrd{\i{}e}ve, s\pln{\i}\si{e}ve (or si\si{e}v\si{e}), fr\si{i}end, t\bbrd{\i{}e}r / t\pln{\i}er, dying, crying, bio, v\stst{i}ol\st{i}n, sk\brd{\i}ing, W\bbrd{\i\i}, Haw\st{\iidp{a\i}}i, r\st{a}di\stst{i}, lion, na\sno{ti}on, cesium, Bel\hno{gi}um, di\udp{u}r\st{\pln{e}}tic (note that the `i' is stressed), han\si{d}kerchi\si{e}f, servi\st{e}tte  \\\hline

ua, ue, ui, uo, uu, uw, uy                                   & dual, ann\nat{u}al, u\svo{su}al, nu\si{i}sance, fluent, fuel, blue, arg\nat{u}e, b\si{u}ild, gluier (from gluey), b\si{u}y, guy, Guy\st{\pln{a}}na  \\\hline
%%uy     & Yes?????????? & \textipa{[aI]}, \textipa{[aI@r]}      & \textipa{[aI]}, \textipa{[aI@r]} &    \\\hline    

\end{tabular}
} %small
\vspace{5mm}
\caption{Examples of vowel pairs which are not units.}
\label{tab:vowelsnotunit} 
\end{center}
\end{table}

The appearance of the symbol \textipa{[j]} in the combination eu/ew in Table \ref{tab:vowels} follows the same pattern as the pronunciation of natural `u' in Table \ref{tab:stressedvowelA} and Table \ref{tab:stressedvowelB}. While `fume' and `few' always perform the sound \textipa{[ju:]} in all accents, and `june' and `jew' always perform the sound \textipa{[u:]} in all the accents, there are differences with other phonemes. Our coding assumes \textipa{[ju:]} everywhere and will try to reduce that in the postprocess stage, depending on the accent which is desired.

Finally, table \ref{tab:rhotic} summarises the behaviour of several cases which are rhotic and non-rhotic, and their final pronunciation.
We do not distinguish between North, sort, warm and force, tore, boar, port, but they do sound differently in Ireland, Scotland and Wales, and parts of the United States. Precisely, it is not a question of being rhotic or not, since all these dialects pronounce the `r'. The difference is more about openness. If we want to highlight the difference we could use the following annotation for the second group: f\brd{o}rce, t\brd{o}re, b\bbrd{oa}r.
We do not distinguish either between fern, fir and fur, even though they may sound (or used to sound) differently in Scotland and parts of Ireland. In any case, no annotation would be needed here, since the letter (e, i, u) can be used to guess the specific pronunciation in each case.

\begin{table}
\begin{center}
{\small
\begin{tabular}{|| p{4cm} | p{2.5cm} | p{6cm}  ||}
\hline
\bf{Units}                       & \bf{IPA}      & \bf{Examples}             \\ \hline \hline
Unstressed single vowels  (equivalent for rhotic and non-rhotic) & \textipa{[@], [i]}  & Color, honors, ir\pln{\i}dium, ir\st{a}scible, farm{\em e}r, ar\st{a}c\si{h}nid, irr\st{a}\sno{ti}onal, \iot{e}r\st{a}\no{s}e, \udp{u}r\st{a}nium, pattern, \udp{u}r\st{\nat{e}}thra, ar\st{i}thmetic, ar\st{o}ma, \iot{e}r\st{\pln{o}}tic   \\ \hline

Unstressed complex vowels (rhotic)  & \textipa{[A\*r], [IK], [EK], [O\*r], [(j)UK], [aIK]} & funfair, centaur, dinosaur, l\pln{\i}near, reindeer, Eur\st{a}\sno{si}an, be\sel{}zoar.   \\ \hline 

Unstressed complex vowels (non-rhotic)  & \textipa{[A:], [i:], [I], [eI], [O:], [oU], [(j)u:], [aI]} & Bei\co{r}\st{\opq{u}}t, eu\co{r}\st{e}ka, Stearr\si{h}\st{\nat{e}}a, \stst{ur}b\st{a}ne (note that the second -ur in murmur has a different sound)   \\ \hline

Stressed single vowels (rhotic) & \textipa{[EK], [IK], [aIK], [O\*r], [(j)UK], [A\*r], [aUK], [3\*r]} & fare, mere, fire, more, cure, e\si{a}rly, far, her, first, for, fur, f\brd{\i{}e}rce, th\brd{e}re,  t\bbrd{\i{}e}r,  err, whirr, 
Mary, lor\sel{}ry (GA sounds like lorry), hur\sel{}ry (GA), fur\sel{}ry, bur\sel{}y (GA), fir\sel{}ry, j\si{o}urney, star, star\se{}ring, forest (GA), coral (GA), \nat{\i}r\si{o}n, c\si{o}\si{e}ur, w\rnd{a}r, w\cnt{o}rd (w\iot{o}rd would work as well), urine, y\si{o}u're (note that this case becomes one single segment), tyrant, d\iot{i}f\st{e}r\se{}ring / d\iot{i}f\st{\pln{e}}ring, \stst{i}r\st{\pln{o}}nic (\nat{\i}r\st{\pln{o}}nic would yield it non-rhotic), no\selr{}wh\brd{e}re (the alternative n\nat{o}wh\brd{e}re would make it non-rhotic), \stst{u}rb\st{a}ne, ergon\pln{\st{o}}mic (note that the `e' has a secondary stress.   \\ \hline

Stressed single vowels (non-rhotic) & \textipa{[eI], [I:], [aI], [oU], [(j)u:], [\ae], [E], [I], [6], [2], [A:], [O:], [aU], [wA:]} & marry (RP), m\cnt{a}rry (some American dialects), berry (RP, GA),  herring (RP, GA), mirror (RP, GA), lorry (RP), hurry (RP), b\cnt{u}\co{r}y (RP), w\iidp{a}rrant, fo\co{r}est (RP), co\co{r}al (RP), terr\cnt{\i}ble, territory (first non-rhotic, second rhotic in GA and unstressed in RP), d\nat{\i}r\st{e}ct, pi\sel{}rate / p\nat{\i}\co{r}ate, error, Ga\co{r}y, pa\co{r}ab\st{\pln{o}}lic, comp\st{a}\co{r}i\no{s}on, o\co{r}\iot{a}nge, mi\co{r}\iot{a}cle, si\co{r}up, Si\co{r}ius, p\nat{y}\co{r}om\st{a}niac, pha\co{r}ynx, hay\selr{}rick, re\serl{}run.   \\ \hline

Stressed double vowels (rhotic) & \textipa{[EK], [IK], [aIK], [O\*r], [(j)UK], [A\*r], [aUK], [3\*r], [waIK]} & Eir\cnt{e}, b\bbrd{ea}r, ear\sel{}ring, pray\si{e}r, may\si{o}r, Europe, \si{h}eir, their, flour, moor, deer, dear, w\bbrd{ea}r, aural, soar, c\si{h}\w{o}\nat{\i}r   \\ \hline

Stressed double vowels (non-rhotic) & \textipa{[eI], [I:], [aI], [oU], [(j)u:], [A:], [aU], [OI], [u:]} & l\oopq{au}\co{r}\iot{e}ate, show\se{}r\oopq{oo}m.   \\ \hline
\hline

\end{tabular}
} % small
\vspace{5mm}
\caption{Summary of rhotic and non-rhotic vowels before `r'.}
\label{tab:rhotic} 
\end{center}
\end{table}

\subsection{Distinguish Consonant Units in Groups using Digraphs}

Once finished with vowels, we start with consonants now. Even though there are many more consonants, the process is much simpler.
The first thing we have to do with consonants is to recognise the consonant units, which means to recognise the digraphs.
Given a consonant group such as `ph\no{th}' in `diphthong' we do not parse it from left to right, but we follow a priorised order of digraphs.
If there is a silent annotation, this does not avoid the formation of a digraph. For instance, `s\si{c}h' allows the formation of the digraph `sh'.

Said this, we identify digraphs following this order:

\begin{enumerate}
\item The digraphs `kn', `pn', `gn', `cn' become \textipa{[n]} at the beginning of a segment. Examples: `know', `knee', `gnaw',`gnome', `pneum\st{o}nia' `fore\serl{}kn\pln{o}\si{w}l\iot{e}dge'. It is maintained in `ackn\st{\pln{o}}\si{w}l\iot{e}dge' (although the `k' sound would be preserved because of the preceding `c'), `darkness', `bankn\nat{o}te', k\stst{a}\co{r}y\stst{o}pykn\st{o}\no{s}i\no{s}
\item The digraph `ph' becomes \textipa{[f]} at any position. Examples: `tr\nat{o}phy', `phil\st{\pln{o}}\no{s}ophy'. Some words use a separator to avoid applying this rule: up\serl{}hill. In other cases, an annotation may indicate a change of sound, as in St\nat{e}\vo{ph}en (note that the consonant group is still considered complex).
\item The digraph `ch' becomes \textipa{[tS]} at any position. Examples: `chin', `rich', `reaching', `D\oopq{eu}t\si{s}ch\selr{}mark'.
\item The digraph `sh' becomes \textipa{[S]} at any position. Examples: `shine', `v\pln{a}nish', `ashes', `s\si{c}hmooze'. Some words may use a separator or annotation to indicate an alternative pronunciation: `j\pln{e}\si{a}lou\no{s}h\oopq{oo}d', `thre\sno{s}h\nat{o}ld'
\item The digraph `ps' becomes \textipa{[s]} at the beginning of a segment. Examples: `ps\nat{y}c\si{h}\pln{o}logy', but `coll\st{a}pse'. Separators can be used to apply the rule:  `pre\serl{}ps\nat{y}c\si{h}\st{\pln{o}}tic' or `pr\nat{e}\si{p}s\nat{y}c\si{h}\st{\pln{o}}tic'.
\item The digraph `rh' becomes \textipa{[\*r]} at the beginning of a segment.  Examples: `rhyme', `rheumatism'.
\item The digraph `pt' becomes \textipa{[t]} at the beginning of a segment.  Examples: `pt\stst{e}\co{r}od\st{a}ctyl'.
\item The digraph `th' becomes \textipa{[D]} before vowel (a,e,i,o,u,y) and \textipa{[T]} in any other position. Examples with  \textipa{[D]}: `then', `that', `thus', `the', `this', `thy', `f\brd{a}ther', `bother', `le\si{a}ther', `we\si{a}ther'. Examples with \textipa{[T]}: `three', `mouth', `birthday'. Some words may use a separator or annotation to indicate an alternative pronunciation: `\no{th}ing', `wi\vo{th}', `T\si{h}\pln{o}mas', `eigh\co{th}', `\pln{a}d\pln{u}lt\serl{}h\oopq{oo}d', `rhy\vo{th}m' / `rhyt\ssch{hm}'.
\item The digraph `gg' becomes \textipa{[g]} elsewhere.  Examples: `bigger'. The other common sound must be annotated: \iot{e}x\st{a}\hvo{gg}er\stst{a}te. Note that. `suggest' (which is generally pronounced as `su\hvo{gg}est') has some peculiar pronunciation in some American accents: sug\selr{}gest
\item The digraph `ss' becomes \textipa{[s]} elsewhere.  Examples: `mess', `lossy'. 
\end{enumerate}

\noindent Note also the order. Only the pairs (`ps'-`sh', `ss'-`sh', `ps'-`ss', `pt'-`th', `gg'-`gn') can overlap, as in the interjection `pshaw', which is first converted into `p\textipa{[S]}aw', and then the rule `ps' cannot be applied and if is finally \textipa{[pS]}aw. The same happens with `troopship' (alhtthough `ps' is not at the beginning of a segment).

Once the previous conversions have been made, we only need to analyse letter by letter.

\subsection{Evaluate Consonant Units}

In tables \ref{tab:consonantsA}, \ref{tab:consonantsB} and \ref{tab:consonantsC} we show the conversion of the consonant units into IPA symbols.  
Many rules are just straightforward.
Many other rules are just scattered around but correspond to a simple rule. For instance, the annotations `\sno{}', `\svo{}', `\hno{}', `\hvo{}' typically correspond to the sounds  \textipa{[S]},  \textipa{[Z]},   \textipa{[tS]}, \textipa{[dZ]} (except `\sno{x}', which corresponds to \textipa{[kS]}), independently of the annotated consonant involved, as shown in table \ref{tab:sno}.

\begin{table}
\begin{center}
{\small
\begin{tabular}{|| c | c |  p{10cm}    ||}
\hline
\bf{Unit}                       & \bf{IPA}       & \bf{Examples}       \\ \hline \hline
{\em \sno{}} (except `\sno{x}')        & \textipa{[S]} 	& sp\pln{e}\sno{ci}al, o\sno{ce}an, cr\iot{e}\sno{sc}endo, con\sno{\group{sci}}ence, \sno{ch}amp{\st{a}}\si{g}ne, ma\sno{ch}\st{\brd{\i}}ne, \sno{s}ugar, \sno{s}ure, pen\sno{si}on, pa\sno{\group{ssi}}on, pre\sno{ss}ure, cen\sno{s}ure, ti\sno{ss}ue, cau\sno{ti}ous,   in\st{e}r\sno{ti}a \\\hline
{\em \svo{}}                    & \textipa{[Z]} 	&   l\clr{i}n\svo{g}erie, \svo{g}\opq{e}nr\cnt{e}, \svo{j}et\st{\brd{e}},  illu\svo{si}on, m\pln{e}\si{a}\svo{s}ure, \iot{e}qua\svo{ti}on, s\nnat{e\i}\svo{z}ure, \pln{a}bk\si{h}a\svo{zi}an, Bre\svo{zh}n\pln{e}\no{v}   \\\hline
{\em \hno{}}                    & \textipa{[tS]} 	& \hno{c}ello , \hno{Cz}ec\si{h}, ques\hno{ti}on,   na\hno{t}ure, cul\hno{t}ure  \\\hline
{\em \hvo{}}                    & \textipa{[dZ]} 	&   s\nat{o}l\hvo{di}er, gr\pln{a}\hvo{d}\opq{u}ate (noun) / gr\pln{a}\hvo{d}\opq{u}\stst{a}te (verb), proce\hvo{d}ure, \iot{e}xa\hvo{gg}er\stst{a}te, Bel\hvo{gi}an, gor\hvo{ge}ous, dun\hvo{ge}on, \hvo{Zh}o\si{u} \\\hline
{\em \sno{x}}                   & \textipa{[kS]} 	&   lu\sno{x}ury  \\\hline

\end{tabular}
} % small
\vspace{5mm}
\caption{Pronunciation of the annotations `\sno{}', `\svo{}', `\hno{}', `\hvo{}'}
\label{tab:sno} 
\end{center}
\end{table}

\begin{table}
\begin{center}
{\small
\begin{tabular}{|| p{2.5cm} | c | p{5cm}   | p{4cm}  ||}
\hline
\bf{Unit}                       & \bf{IPA}      & \bf{Examples}                 & \bf{Comments} \\ \hline \hline

{\em b}                             & \textipa{[b]}	& ab\clr{o}ve, bay, rib               & \\\hline

{\em c} (before e, i and y), {\em \no{c}}   & \textipa{[s]} & ice, cyan, scene, ac{\em c}ent, fl\pln{a}\si{c}cid                   & \\\hline

{\em c} elsewhere, \co{c}           & \textipa{[k]} & cat, sick, bis\co{c}\si{u}it, a\co{c}\si{h}e,  \co{c}eltic, a{\em c}cent, s\co{c}\si{h}ema, c\si{h}ord, drac\si{h}ma, Mic\si{h}a\si{e}l, \co{c}\si{h}a\pln{o}s, sc\si{h}ool                          & \\\hline

{\em  \sno{c}, \sno{ci}, \sno{ce}, \sno{cy}, \sno{sc}, \sno{sci}, \sno{sce}, \sno{scy} } & \textipa{[S]} & sp\pln{e}\sno{ci}al, o\sno{ce}an, cre\sno{sc}endo, con\sno{\group{sci}}ence &            \\\hline

{\em \hno{c}, \hno{ci}, \hno{ce}, \hno{cy}, \hno{cz} } & \textipa{[tS]} & \hno{c}ello , \hno{Cz}ec\si{h}                    & digraph `ch' doesn't reach this point. \\\hline 

{\em \sno{ch}}                      & \textipa{[S]} & \sno{ch}amp{\st{\em a}}\si{g}ne, ma\sno{ch}\st{\brd{\i}}ne & digraphs `sh' or `ch' don't reach this point. \\\hline 

{\em d}                             & \textipa{[d]} & dice, red, ladder            & \\\hline

{\em \no{d}}                        & \textipa{[t]} & rippe\no{d}, Ap\st{a}rt\si{h}ei\no{d} (RP, formal)                & \\\hline

{\em \hvo{d}, \hvo{di}, \hvo{de}, \hvo{dy}, \hvo{dj} } & \textipa{[dZ]} & s\nat{o}l\hvo{di}er, gr\pln{a}\hvo{d}\opq{u}ate (noun) / gr\pln{a}\hvo{d}\opq{u}\stst{a}te (verb), proc\st{e}\hvo{d}ure, Djib\st{\rnd{o}}\si{u}ti                    &  \\\hline 

{\em f, \co{gh}, \no{u}}            & \textipa{[f]} & fit, carf, co\si{u}\co{gh}, t\si{o}u\co{gh}, l\si{i}\pln{e}\no{u}t\st{\pln{e}}nant, tr\nat{o}\co{ph}y            & .\\\hline

{\em \vo{f}, \vo{ph}}                        & \textipa{[v]} & o\vo{f}, St\nat{e}\vo{ph}en           & digraph `ph' doesn't reach this point. \\\hline

{\em g} (before e, i and y), {\em \hvo{g}, \hvo{gi}, \hvo{ge}, \hvo{gy}, \hvo{gg}}  & \textipa{[dZ]} & gin, bridge, \iot{e}x\st{a}\hvo{gg}er\stst{a}te, \hvo{g}\nat{a}\si{o}l, ginger, Bel\hvo{gi}an, gor\hvo{ge}ous, dun\hvo{ge}on, Lo\no{s} Angel\cnt{e}s, adv\pln{a}nta\hvo{ge}ous (note the primary stress is on the last `a')         &  The digraph `gg' does not reach this point. \\\hline

{\em g} elsewhere, \co{g}           & \textipa{[g]}  & goat, mug, fin\co{g}er,  g\si{h}\nat{o}st, tar\co{g}\iot{e}t, \co{g}\si{h}etto, Spa\co{g}\si{h}\st{e}tti, \co{g}irl  & The digraph `gg' does not reach this point.       \\\hline

{\em \no{g}}                             & \textipa{[k]} &  lo\si{u}\no{g}\si{h}, \opq{u}\no{g}\si{h}           & \\\hline

{\em \svo{g}}                             & \textipa{[Z]} &  l\clr{i}n\svo{g}erie, \svo{g}\opq{e}nr\cnt{e}          & \\\hline

{\em \y{i}, \y{j}, \y{}}                             & \textipa{[j]} &  mill\y{i}on, H\stst{a}llel\st{u}\y{j}ah, fail\y{}ure          & \\\hline

{\em j, \hvo{ji}, \hvo{je}, \hvo{jy} } & \textipa{[dZ]} & jet, injure, F\brd{\i}\hvo{ji}an                    &  \\\hline 

{\em \co{j}}                             & \textipa{[h]} & fa\co{j}\st{\brd{\i}}ta            & \\\hline

{\em \svo{j}}                             & \textipa{[Z]} &  \svo{j}et\st{\brd{e}}           & \\\hline

{\em h} (except final), \no{h}                    & \textipa{[h]} & hen, \si{w}h\rnd{o}          &  \\\hline

{\em h} between vowel and consonant                  & \textipa{[]} & \brd{a}h, \brd{e}h, \pln{u}h, oh, y\pln{e}ah, Sarah, Noah, Ut\rnd{a}h, y\pln{e}ah, sh\brd{a}h           &  This rule is applied at a previous stage, so the vowels are interpreted as if the `h' were not there. \\\hline  %\todo{THIS SHOULD BE A RULE TO BE APPLIED BEFORE.}???

{\em \co{h}, \co{ch}}                             & \textipa{[x]} & lo\co{ch}  / lo\si{u}\si{g}\co{h}        & \\\hline

{\em k}                             & \textipa{[k]} & kit, black, baker         & \\\hline

\end{tabular}
} % small
\vspace{5mm}
\caption{Pronunciation of single consonants (part 1/3)}
\label{tab:consonantsA} 
\end{center}
\end{table}

\begin{table}
\begin{center}
{\small
\begin{tabular}{|| p{2.5cm} | c | p{3.8cm} |  p{6cm}  ||}
\hline
\bf{Unit}                       & \bf{IPA}      & \bf{Examples}               & \bf{Comments} \\ \hline \hline

{\em l}                             & \textipa{[l]} & lake, f\rnd{a}ll,         & \\\hline

{\em \no{l} (non-rhotic), \co{l}} (rhotic)                    & \textipa{[\*r]} & c\cnt{o}\no{l}\si{o}nel         & \\\hline

{\em \y{l}}                             & \textipa{[j]} &  t\stst{o}rt\st{\brd{\i}}\y{\group{ll}}a          & \\\hline

{\em m}                             & \textipa{[m]} & mum, lame, hym\si{n}       &  \\\hline

{\em n} (except before k, -ca, -co, -cu, -cw or -\co{c}), {\em \no{n}}                              & \textipa{[n]} & non, p\nat{\i}nt, pir\st{\pln{a}}n\si{h}a, la\no{s}\st{\pln{a}}\si{g}n\y{}a, in\ser{}cur, in\ser{}corpor\stst{a}te, un\serl{}k\nat{\i}nd, con\serl{}cave       &   digraph `ng' doesn't reach this point. \\\hline

{\em n} (before -k, -ca, -co, -cu, -cw or -\co{c}), {\em \co{n}}                        & \textipa{[N]} & donkey, \stst{a}nkyl\st{o}\no{s}i\no{s}, I\co{n}ca, tr\clr{u}nc\st{a}te     & digraph `ng' doesn't reach this point. \\\hline

{\em p}, {\em \no{gh}}                      & \textipa{[p]} & pet, cup, copper, hicc\cclr{ou}\no{gh}    &  digraphs `ph', `pt' don't reach this point. \\\hline

{\em q}                      & \textipa{[k]} & queen, Ir\st{\brd{a}}q  &    digraphs `qu' doesn't reach this point. \\\hline

{\em r, \co{r}}                      & \textipa{[\*r]} & rip, far, Mary, Pa\co{r}is  &    the appearance of `r', `rr' or `\co{r}' makes the position rhotic or non-rhotic and affects the vowels in many different ways, as seen in a separate table. \\\hline

{\em \no{s}}, {\em s} (starting `s'; `s' between consonant and vowel; any other case when preceded or followed by a voiceless consonant: c, f, h, k, p, t, x)      & \textipa{[s]}                &  sin, snake, slave, must, \rnd{a}lso, tense, horse, first, it's, pets, rates (note that the e is silent here), mess, lossy,  de\si{a}\no{th}s, ye\no{s},  hou\no{s}e, venu\no{s}, u\no{s}, la\no{s}\st{\pln{a}}\si{g}n\y{}a, \no{th}e\no{s}i\no{s}, wh\pln{\i}\no{s}\si{t}le, dress, possible, seriou\no{s}, seriou\no{s}ness, ba\no{s}eball, ba\no{s}eline, bi\serl{}sect &  A voiceless consonant is any of (c,f,h,k,p,t,x) or an annotated consonant or digraph whose sound is voiceless (e.g. th in `b\brd{a}ths'). Note: if `s' is alone in a segment, it becomes voiceless if the previous segment ends with a voiceless consonant group (e.g. it's). Note that the digraph `ss' has been processed before (and it is voiceless). \\\hline

{\em \vo{s}}, \vo{ss} {\em s} (between vowels; between vowel and voiced consonant; after vowel or voiced group at the end of a segment)                      & \textipa{[z]} & easy, hous\iot{e}s, l\rnd{o}se, pos\si{s}\st{e}ss (first `s' is voiced), wisdom, tr\pln{a}nsm\pln{\i}\sno{\group{ssi}}on (RP), Thursday, b\iot{u}{\em s}\si{i}n\iot{e}ss, r\brd{a}s\si{p}berry (RP), ras\si{p}b\pln{e}rry (GA), Disney, aut\st{i}sm, isn't, s\iidp{e\i}sm\st{\pln{o}}logy, as, \rnd{a}lways, coach\iot{e}s, lads, metres (note that the final `e' is silent, so we consider the `r'), meters, tales (note that the e is silent here), bridg\iot{e}s, cherries (note that the e is silent here), robs, bars, fans, Charles, Ann's, he's, sci\vo{ss}ors  & A voiced consonant is any of (b,d,g,j,l,m,n,r,v,w,y,z) or an annotated consonant or digraph whose sound is voiced (e.g. th in `cl\nat{o}thes'). This sound at the end of a segment may finally become voiceless if close to another segment which starts with a voiceless group. For instance, in `The cat is g\oopq{oo}d', we have that `is' makes its voiced sound, while in `The cat is str\nat{a}nge' it makes a voiceless sound. The annotation is not affected.  Note: if `s' is alone if a segment, it becomes voiced if the previous segment ends with a vowel or a voiced consonant group (e.g.: He's). \\\hline

{\em \svo{s}, \svo{si}, \svo{se}, \svo{sy} } & \textipa{[Z]} & illu\svo{si}on, m\pln{e}\si{a}\svo{s}ure       &              \\\hline

{\em \sno{s}, \sno{si}, \sno{se}, \sno{sy}, \sno{ss}, \sno{ssi}, \sno{sse}, \sno{ssy} } & \textipa{[S]} & \sno{s}ugar, \sno{s}ure, pen\sno{si}on, pa\sno{ssi}on, pre\sno{ss}ure, cen\sno{s}ure, ti\sno{ss}ue      &              \\\hline

\end{tabular}
} % small
\vspace{5mm}
\caption{Pronunciation of single consonants (part 2/3)}
\label{tab:consonantsB} 
\end{center}
\end{table}

\begin{table}
\begin{center}
{\small
\begin{tabular}{|| p{2.5cm} | p{2cm} | p{4cm} | p{4.5cm}  ||}
\hline
\bf{Unit}                       & \bf{IPA}      & \bf{Examples}              &  \bf{Comments} \\ \hline \hline

{\em t}                     & \textipa{[t]} & tame, cat, site, fatty, Ant\si{h}ony  &    Digraph `th' doesn't reach this point. \\\hline

{\em \co{t}}                     & \textipa{[R]} & compu\co{t}er, pr\iot{e}\co{t}\co{t}y, go\co{t}\co{t}a (some accents)  &    This is not used in formal transcription. . In fact, the rule is that this happens for American English when `t' is between vowels. The same could be defined for \co{d}, which sounds similarly in some accents for some words\\\hline

{\em \sno{t}, \sno{ti}, \sno{te}, \sno{ty} } & \textipa{[S]} & cau\sno{ti}ous,   iner\sno{ti}a    &              \\\hline 

{\em \hno{t}, \hno{ti}, \hno{te}, \hno{ty} } & \textipa{[tS]} & ques\hno{ti}on,   na\hno{t}ure, cul\hno{t}ure    &              \\\hline 

{\em \svo{t}, \svo{ti}, \svo{te}, \svo{ty} } & \textipa{[Z]} & \iot{e}qua\svo{ti}on    &             \\\hline 

{\em \co{th}}                     & \textipa{[tT]} & `eigh\co{th}'  &    digraph `th' doesn't reach this point. \\\hline

{\em v } & \textipa{[v]} & van, save, savvy                  &  \\\hline 

{\em \no{v} } & \textipa{[f]} & Bre\svo{zh}n\pln{e}\no{v}, l\iidp{e\i}tm\nat{o}t\brd{\i}\no{v}                  &  \\\hline 

{\em w, \w{w}, \w{}}                      & \textipa{[w]} & win, twin, \w{}\clr{o}nce   & Note that w in vowels have been processed previously, such as in `raw'. \\\hline

{\em \vo{w}}                      & \textipa{[v]} & rott\vo{w}\iidp{e\i}ler (although it is also simply pronounced as rottw\iidp{e\i}ler), \vo{W}\st{i}ttgen\sno{s}t\iidp{e\i}n &   Note that w in vowels have been processed previously, such as in `raw'. \\\hline

{\em \co{w}}                  & \textipa{[w]} or \textipa{[\*w]}, depending on the accent & when, why &   \\\hline

{\em \co{wh}}                  & \textipa{[\*w]} & \co{wh}en, \co{wh}y &   The annotator has indicated that she wants the sound to be pronounced as [hw]  \\\hline

{\em x} (before stressed vowel), {\em \vo{x}}               & \textipa{[gz]} & \iot{e}x\st{a}mple, \iot{e}xa\hvo{gg}er\stst{a}te &    \\\hline

{\em x} (any other position), {\em \no{x}}               & \textipa{[ks]} & extra, \iot{e}xtr\st{e}me, g\pln{a}laxy, taxi &   \\\hline

{\em \co{x}}               & \textipa{[z]} & \co{x}\st{y}loph\stst{o}ne, bur\si{e}\si{a}\rnd{u}\co{x} &    \\\hline

{\em \sno{x} } & \textipa{[kS]} & lu\sno{x}ury    &             \\\hline 

{\em y}                      & \textipa{[j]} & yale, ye\no{s} &   Note that y in vowels have been processed previously, such as in `ray'. \\\hline

{\em z}                 & \textipa{[z]} & crazy, zoo, buzz, \si{t}zar  & \\\hline

{\em \co{z}}                      & \textipa{[ts]} & p\brd{\i}\co{z}\si{z}a, Al\co{z}h\iidp{e\i}mer  & \\\hline

{\em \no{z}}                      & \textipa{[s]} & glit\no{z}y  &  \\\hline

{\em \svo{z}, \svo{zi}, \svo{ze}, \svo{zy}, \svo{zh} } & \textipa{[Z]} & s\nnat{e\i}\svo{z}ure, \pln{a}bk\si{h}\st{a}\svo{zi}an, Bre\svo{zh}n\pln{e}\no{v}      &             \\\hline 

{\em \hvo{z}, \hvo{zh} } & \textipa{[dZ]} & \hvo{Zh}o\si{u}       &             \\\hline

\end{tabular}
} % small
\vspace{5mm}
\caption{Pronunciation of single consonants (part 3/3)}
\label{tab:consonantsC} 
\end{center}
\end{table}

After these tables, we have converted everything into IPA.
Before going to postprocessing, let us see in table \ref{tab:qgc} a summary of how $c$, $g$, and $q$ are handled.

\begin{table}
\begin{center}
{\small
\begin{tabular}{|| c | c | p{4.5cm} | p{3cm}  ||}
\hline
\bf{Unit}                       & \bf{IPA}      & \bf{Examples} & \bf{Negative cases}             \\ \hline \hline

{\em qu + consonant, q + $\emptyset$}       & \textipa{[k]} & Ir\st{\stst{a}}q, Q\opq{u}r\st{\brd{a}}n & \\\hline
{\em qu + \{a,e,i,o,u,y\} }       & \textipa{[kw]} & queen, qu\opq{a}lity, quick & q\si{u}\si{a}\brd{y} \\\hline
{\em gu + \{a,o,u\} }       & \textipa{[gw]} & bi\serl{}l\st{i}ngual, gu\brd{a}no & cont\st{\pln{\i}}g\nat{u}ous, g\si{u}ard \\\hline
{\em gu + \{e,i,y\} }       & \textipa{[g]} & guess, guilty, guy & g\si{u}ard, dist\st{i}ng\w{u}ish \\\hline
{\em g + \{a,o,u\} or consonant }       & \textipa{[g]} & gas, gun, green, bag & \hvo{g}\nat{a}\si{o}l \\\hline
{\em g + \{e,i,y\} }       & \textipa{[dZ]} & gin, gene & \co{g}irl \\\hline
{\em c + \{a,o,u\} or consonant }       & \textipa{[k]} & can, cult, crazy, black, crystal & \\\hline
{\em c + \{e,i,y\} }       & \textipa{[s]} & centre, cyan & \co{c}elt \\\hline

\end{tabular}
} % small
\vspace{5mm}
\caption{Summary of sounds for q, g, c + vowel}
\label{tab:qgc} 
\end{center}
\end{table}

\subsection{Recompose the Words}

This step is just gluing all the units together to the level of word.

\subsection{Postprocess the IPA result} 

Finally, the last step performs some small adjustments in some cases. 

First, we perform the elimination of repeated consonants. For instance, words such as `nutty', `lorry', `cure', `scent' produce \textipa{[n2ttI]}, \textipa{[l6\*r\*rI]}, and \textipa{[ssEnt]}. Repeated consonants must be eliminated.

Second, the combination of some consonants can produce pairs which cannot be properly pronounced. The clearest case is the ending in -tre, as in `metre', which produces $\ra$ \textipa{[m\st{i}:t\*r]}, but should be \textipa{[m\st{i}:t@\*r]}, by introducing the schwa (an unstressed sound). 
Does this mean that `metre' should be annotated as met\sch{}re? 
Does this happen for other consonant combinations? Can we distinguish the right positions? (We do not have the problem for `central' or `centring').

If we first analyse the positions, the problem appears to be when `\#r' is not followed by a vowel. But consider the word `fibre-optics'. If we consider the two segments together and annotate it like this `fibr\si{e}optics', then the pronunciation is not correct. So the rule is confirmed. The problem appears when `\#r' is not followed by a vowel inside the segment.

Regarding the question of the combinations, it is very difficult (e.g. \textipa{[tT]} works, but \textipa{[Tt]} does not, while both \textipa{[ts]} works and \textipa{[st]} do, and of course \textipa{[bd]}, \textipa{[ft]}, etc.). We could analyse the approximately 400 combinations of pair of consonants to see whether they articulate well together, where many of them never happen. Instead, the idea is then to derive a very simple rule to cover the most frequent cases, and then annotate any possible exception. This is the philosophy taken for the process.

We distinguish two common situations:

\begin{itemize}
\item \textipa{[-l]}, \textipa{[-m]}, \textipa{[-n]}. Examples: able, girl, little, rhy\vo{th}m, aut\st{i}sm, isn't, re\si{a}lm, farm, fern. There is no need to insert a schwa, even though in some cases it seems that a schwa might be convenient (rhy\vo{th}m $\rightarrow$ rhyt\sch{hm} , aut\st{i}sm $\rightarrow$  aut\st{i}\ssch{sm},  isn't $\rightarrow$ i\ssch{sn}'t). 
\item \textipa{[-\*r]}. Examples: fibre, ogre, metre. We need to insert a schwa.
\end{itemize}

\noindent So the rule will be to introduce a schwa between a consonant + `r'/`rs' at the end of a segment, as in `metre' or `metres', which initially produces $\ra$ \textipa{[m\st{i}:tr]} but finally it is converted into \textipa{[m\st{i}:t@r]}. Note that this does not conflict with words such as `g\rnd{e}nr\cnt{e}', because the `r' is not at the endof the semgent.
In some other cases where it is also necessary, we will annotate it. For instance: `i\ssch{bn}' .

The opposite problem is whether we want to reduce idol, open, icon, t{\em i}ghten, f\brd{a}\no{s}\si{t}en, cottons, isn't,  etc. In this case, if the annotator wants to do it because in her accent `idol' and `idle' are the same, it is just enough to use the annotation \si{} for that (e.g. \nat{\i}d\si{o}l, instead of idol).

As a summary, we will only automatically deal with the `r' (which does not frequently happen with American spelling except for `acre' and some other few words), and the rest will be left to the annotator (as in aut\st{i}\ssch{sm}, or \nat{o}p\si{e}n ).

Now, we deal with different English varieties.
The following conversions are performed for GA:

\begin{center}
\noindent\textipa{[K]} $\rightarrow$  \textipa{[\*r]} (e.g. `fire', `beer') \\
\noindent\textipa{[3:]} $\rightarrow$  \textipa{[3]} (e.g. `fur') \\
\noindent\textipa{[6]} $\rightarrow$  \textipa{[A]} (e.g. `pot', `bother') \\
\noindent\textipa{[A:]} $\rightarrow$  \textipa{[A]} (e.g. `f\brd{a}ther') (note that we annotate `fast' and not `f\brd{a}st' for GA) \\
\noindent\textipa{[i:]} $\rightarrow$  \textipa{[i]} (e.g. `feed') (not all phonologists advocate for this phoneme change) \\
\noindent\textipa{[O:]} $\rightarrow$  \textipa{[O]} (e.g. `feed') (not all phonologists advocate for this phoneme change) \\
\noindent\textipa{[u:]} $\rightarrow$  \textipa{[u]} (e.g. `moon') (not all phonologists advocate for this phoneme change) \\
\end{center}

\noindent The following conversions are performed for RP:

\begin{center}
\noindent\textipa{[\*r]} $\rightarrow$  \textipa{[\*r]} (e.g. `for', if followed by a non-silent vowel, such as `for us', `carry' or `rat') \\
\noindent\textipa{[\*r]} $\rightarrow$  \textipa{[]} (e.g. `for', if not followed by a non-silent vowel, such as `for them', `forth' or `furs' \\
\noindent\textipa{[K]} $\rightarrow$  \textipa{[@\*r]} (e.g. `dear', if followed by a non-silent vowel, such as `dear Ann', `firing' or `dearest') \\
\noindent\textipa{[K]} $\rightarrow$  \textipa{[@]} (e.g. `dear', if not followed by a non-silent vowel, such as `dear John' or `moors') \\
\noindent\textipa{[oU]} $\rightarrow$  \textipa{[@U]} (e.g. `tone') \\
\end{center}

\noindent Other conversion rulesets can be established for other dialects (Australia, Scotland, etc.)
Note that the process only deals with these mergers at the end of the process. This is useful to distinguish between `flower' and `flour', since the first is \textipa{[flaU@\*r]} in GA and \textipa{[flaU@]} in RP, while the second is \textipa{[flaU\*r]} in GA and \textipa{[flaU@]} in RP.

Another issue is the pronunciation of the diphthong \textipa{[(j)u:]}, which is sometimes performed as \textipa{[ju:]} and sometimes as \textipa{[u:]}.
In tables \ref{tab:stressedvowelA}, \ref{tab:stressedvowelB} and \ref{tab:vowels} we denoted by \textipa{[(j)u:]} the correspondence for `eu/ew' and natural `u'. While `fume' and `few' always perform the sound \textipa{[ju:]} in all accents, and `june' and `jew' always perform the sound \textipa{[u:]} in all the accents, there are differences with other phonemes. In this final part of the process we eliminate the symbol (j) depending on the accent. For instance, if we choose RP, then it is only reduced in the following cases:

\begin{center}
\noindent\textipa{[rju:]} $\rightarrow$  \textipa{[ru:]} \\
\noindent\textipa{[lju:]} $\rightarrow$  \textipa{[lu:]} \\
\noindent\textipa{[Sju:]} $\rightarrow$  \textipa{[Su:]} \\
\noindent\textipa{[tSju:]} $\rightarrow$ \textipa{[tSu:]} \\
\noindent\textipa{[Zju:]} $\rightarrow$  \textipa{[Zu:]} \\
\noindent\textipa{[dZju:]} $\rightarrow$ \textipa{[dZu:]} \\
\end{center}

\noindent In case we want to force \textipa{[u:]} for a word we can use \brd{u}, \si{e}\brd{u}, \si{e}\brd{w}.
In case we want to force \textipa{[ju:]} for a word we can use \y{}u, \y{}eu, \y{}ew.

Finally, the elimination of repeated consonants is performed again.

With all this, we have the IPA pronunciation for each word. Note that the IPA produced will not be exactly a standard IPA because of the placement of stress. We place stress on vowels, while IPA place stress at the start of syllables.
Placing the stress with the IPA notation will not be performed, since we have not considered the notion of syllable, only vowel and consonant units.

\section{Annotation Rules} \label{sec:annotate}

After the previous section, we know how an annotated word must be interpreted (i.e. pronounced) in a precise way. Consequently, for any word a reader may have doubts about its pronunciation, the previous rules can be applied and the true pronunciation is obtained.

As discussed in the introduction, in order to read annotated English, someone (or something) must have annotated the text previously. As seen in Figure ºref{fig:inteannot}, we require two inputs for the introduction: the English word and its pronunciation, in order to produce an output: the annotated word. We can distinguish two issues here. First, going from the English word + pronunciation to the annotated word can be easy in some cases, but cumbersome in others. In fact, the problem is like finding a transformation, using deletions, insertions and substitutions, that makes an alignment between the original English word and its pronunciation. The problem is easier than the general computational problem since we only have three possible insertions (`{}\y{}', `\w{}', `\sch{}'). For instance, the word `house' and the pronunciation \textipa{[haUs]} easily suggest that we find a possible sound for `h' which matches the IPA  \textipa{[h-]}, we also find a possible sound for `s'  (\no{s}) which matches the IPA  \textipa{[-s]}, and we have that the digraph `ou' can be converted into \textipa{[-aU-]}. Finally, a deletion of the `e' makes the correspondence perfect. This gives us the correct annotation `hou\no{s}\si{e}'. Following the rules seen in the previous section, we can drop the last annotation and produce `hou\no{s}e'.

The previous process is relatively simple for a human being if she has a good knowledge of the interpretation rules. A more difficult problem is how to make that automatically, i.e., by a computer program. This would require a coding of the rules, several transformation / alignment techniques and some off-the-shelf algorithms, such as dynamic programming. This is a solvable technical problem if we have the word and the pronunciation\footnote{The pronunciation of every English word (including plurals and inflected verbs) can be obtained from several corpora and webpages.}. Given the small size of the words (each word is analysed independently), there are no relevant computational cost issues. The annotation process can be implemented  in such a way that it annotates very fast on any personal computer.
The only main concern to make this process automatic is the distinction of homographs which are not homophones (wind, live, use, read, lead, etc.). Some disambiguation techniques from computational linguistics would be required, which can resolve a high proportion of cases. The most difficult cases might be left for human supervision (perhaps one in each ten thousand words). 

In this section we are not addressing the automation of the annotation process. We leave the analysis of the automation out of this paper. In this section, we discuss on a related problem we have to solve first before any manual or automated annotation process can begin. The problem is that, given an English word and a pronunciation, several alternative (and correct) annotations are possible. For instance, we can annotate `he\si{a}d', `h\si{e}\cnt{a}d' and `h\pln{e}\si{a}d'.

An option would be to allow all of them (all of them can be interpreted and give the same, correct, result) and choose them randomly or by a lexicographical order. However, having different options depending on the annotator would cause confusion, since the same word with the same pronunciation would look like differently depending on the occasion. Consequently, we need to settle a `standard' choice, which gives preference for one annotation over the rest. This is also shown at the bottom of Figure \ref{fig:inteannot}.

\subsection{Annotation Preference Criteria for a Standard Annotation}

In order to define a standard to choose one annotation over the rest, we can consider the following kinds of criteria:

\begin{itemize}
\item Economy. The fewer annotations a word has the better. So, we should prefer `he\si{a}d' over `h\si{e}\cnt{a}d' and `h\pln{e}\si{a}d', and we should also prefer `hou\no{s}e' over `hou\no{s}\si{e}'. In fact, the rules in the previous section have been devised to avoid annotations when the (part of the) word follows the rule, so it would be stupid to include an annotatation which is not required.
\item Simplicity. Some annotations are simpler than others. For instance, `wer\si{e}' is simpler than `w\cnt{e}re'.
\item Explicitness.  Some annotations are more explicit than others. For instance, `w\cnt{e}re' is more explicit than `wer\si{e}'.
\item Frequency. Some annotations rely on more frequent symbols than others. For instance, `h\pln{e}\si{a}d' is preferrable over `h\si{e}\cnt{a}d'.
\item Etymology. Some annotations are more consistent with the etymology or word formation rules. For instance, `a\ser{}go' is preferrable over `ag\ser{}o'.
\end{itemize}

\noindent The problem is that the previous criteria are conflicting. For some words, applying one criterion implies going against some other criteria.
There are two options here: (1) to devise a priority of criteria, (2) to give cost weights to each criteria, compute the costs and choose the annotation with the lowest cost. Note that the second option subsumes the first if we define high differences between the costs. In any case, both options are fine whenever they make a choice and allow for a standard to be defined.

In what follows we propose a cost weighting approach. We exclude etymology from the criteria, since this would require additional information and would make the process difficult to automate. In fact, we give high costs to separators, since we do not want to see things such as `ag\ser{}o', `thirt\serl{}een' or `microph\selr{}one', because they are not etymological. So, the costs we propose are shown in Table \ref{tab:acosts}.

\newcommand{\stcost}{30}
\newcommand{\ststcost}{30}
\newcommand{\sicost}{31}
\newcommand{\rcost}{10}
\newcommand{\brdcost}{34}
\newcommand{\idpcost}{34}
\newcommand{\udpcost}{34}
\newcommand{\bbrdcost}{37}
\newcommand{\iidpcost}{37}
\newcommand{\uudpcost}{37}
\newcommand{\plncost}{42}
\newcommand{\natcost}{42}
\newcommand{\pplncost}{53}
\newcommand{\nnatcost}{53}
\newcommand{\restcost}{44}
\newcommand{\rrestcost}{55}
\newcommand{\secost}{65}
\newcommand{\selcost}{66}
\newcommand{\sercost}{66}
\newcommand{\selrcost}{92}
\newcommand{\serlcost}{92}
\newcommand{\introcost}{200}

\begin{table}
\begin{center}
{\small
\begin{tabular}{|| p{6cm}| c  | p{5cm}  ||}
\hline
\bf{Type of annotation}                                                                      & \bf{Cost}           & \bf{Comments}         \\ \hline \hline
Stress (`\st{a}' or `\stst{a}')                                                              &  \stcost{}          &                       \\\hline
Silent (`\si{a}')                                                                            &  \sicost{}          &                       \\\hline
Non-rhotic annotation (`co{r}')                                                              &  \rcost{}           &                       \\\hline
Broad, idiphthong or udiphthong annotation affecting one vowel                               &  \brdcost{}         &                       \\\hline
Broad, idiphthong or udiphthong annotation  affecting two vowels                             &  \bbrdcost{}        &                       \\\hline
Frequent Annotations affecting one vowel (plain, natural)                                    &  \plncost{}         &                       \\\hline
Double Frequent Annotations (plain, natural)                                                 &  \pplncost{}        &                       \\\hline
Non-Frequent Annotations (the rest)                                                          &  \restcost{}        &                       \\\hline
Double or Triple Non-Frequent Annotations (the rest)                                         &  \rrestcost{}       &                       \\\hline
Separation not affecting stress (`\se{}') not leaving a consonant group alone                &  \secost{}          &                       \\\hline
Simple separation affecting stress (`\sel{}', `\ser{}') not leaving a consonant group alone  &  \selcost{}         &                       \\\hline
Separation affecting stress (`\selr{}', `\serl{}')                                           &  \selrcost{}        &                       \\\hline
Introductions (`\w{}', `\y{}' or `\sch{}', with or without annotation, as in metre)          &  \introcost{}       &                       \\\hline
\hline
\end{tabular}
} % small
\vspace{5mm}
\caption{Annotation Costs}
\label{tab:acosts} 
\end{center}
\end{table}

For each annotation, we also reckon the position cost, which is an additional number we sums up the position number where all the annotations have taken place. This position is negative for deletions (`\si{a}') and positive for any other annotation. For instance, `\iot{e}xtr\st{e}me' has position cost $+1+5 = +6$, and `j\pln{e}\si{a}lou\no{s}' has position cost $+2-3+7 = +6$. The position cost is only used to solve ties when applying the costs in Table \ref{tab:acosts}. Generally, this means that with annotations with the same cost, we prefer the annotation to be leftmost unless it is a deletion, where we prefer it to be rightmost. For instance, `d\si{o}or' has position cost $-2$, while `do\si{o}r' has position cost $-3$ (hence `do\si{o}r' is preferrable).

We apply the costs in tables \ref{tab:exacost1}, \ref{tab:exacost2} and \ref{tab:exacost3} for some alternative annotations in the examples shown \ref{tab:acosts}.

\begin{table}
\begin{center}
{\small
\begin{tabular}{|| p{7cm} | p{6cm}  ||}
\hline
\bf{Type of annotation}                                                 & \bf{Comments}         \\ \hline \hline

y\si{o}u (\sicost{}), y\rnd{o}\si{u} (\restcost{}+\sicost{}), \si{y}\si{o}u (\sicost{}+\sicost{})                 &                       \\\hline

Ire\se{}land (\secost{}), \nat{I}r\si{e}land (\natcost{}+\sicost{})                           &                       \\\hline

Sp\pln{a}nish, Span\se{}ish (\secost{}) (\plncost{})                           &                       \\\hline

do\si{o}r (\sicost{}), d\si{o}or (\sicost{}), d\bbrd{oo}r (\bbrdcost{}), d\nnat{oo}r (\nnatcost{})                               & Position costs are $-2$ and $-3$ respectively. Hence `do\si{o}r' is preferrable.    \\\hline

Annot\st{a}t\iot{e}d (\stcost{}+\restcost{}), Annot\serl{}at\iot{e}d (\serlcost{} + \restcost{})  &                       \\\hline

bl\cclr{oo}d (\rrestcost{}), bl\clr{o}\si{o}d (\restcost{}+\sicost{}), bl\si{o}\clr{o}d (\sicost{} + \restcost{})  &                \\\hline

\iidp{ey}e (\iidpcost{}), \si{e}y\si{e} (\sicost{}+\sicost{}), \idp{e}\si{y}e (\idpcost{}+\sicost{})    &  \si{e}ye (\sicost{}) is not valid since it makes `yee'                \\\hline

t\bbrd{wo} (\bbrdcost{}), t\si{w}\rnd{o} (\sicost{}+\restcost{})    & tw\rnd{o} (\restcost{}) is invalid, since there is no semiconsonantic `w' sound                \\\hline

l\pln{e}vel (\plncost{}), lev\se{}el (\secost{}), lev\sel{}el (\selcost{})   &                \\\hline

br\bbrd{ea}k (\bbrdcost{}), br\brd{e}\si{a}k (\brdcost{}+\sicost{}), br\si{e}\nat{a}k (\sicost{}+\natcost{})   &                \\\hline

 br\si{e}aking (\sicost{}), br\bbrd{ea}king (\bbrdcost{}), br\brd{e}\si{a}king (\brdcost{}+\sicost{})   &                \\\hline

ke\si{y} (\sicost{}), k\nat{e}\si{y} (\natcost{}+\sicost{}), k\nnat{ey} (\nnatcost{}), k\si{e}\brd{y} (\sicost{} + \brdcost{})   &                 \\\hline

c\oopq{ou}\si{l}d (\restcost{}+\sicost{}), c\opq{o}\si{u}\si{l}d (\restcost{}+\sicost{}+\sicost{}), c\si{o}\opq{u}\si{l}d (\sicost{}+\restcost{}+\sicost{})      &               \\\hline

t\oopq{ou}r (\rrestcost{}), t\si{o}\brd{u}r (\sicost{} + \brdcost{}), t\opq{o}\si{u}r (\restcost{} + \sicost{}) &  \\\hline

b\si{u}ild (\sicost{}), b\iot{u}\si{i}ld (\restcost{}+\sicost{})      &               \\\hline

c\cclr{ou}s\cnt{i}n (\rrestcost{}+\restcost{}), c\si{o}\pln{u}s\cnt{i}n (\sicost{}+\plncost{}+\restcost{}), c\clr{o}\si{u}s\cnt{i}n (\restcost{}+\sicost{}+\restcost{}), c\si{o}\pln{u}s\cnt{i}n (\sicost{}+\plncost{}+\restcost{}) & \\\hline

b\cnt{u}\co{r}y (\restcost{}+\rcost{}), b\cnt{u}\se{}ry (\restcost{}+\secost{})  &   \\\hline

cling\se{}y (\secost{}), cli\co{n}\si{g}y (\restcost{}+\sicost{})  &  \\\hline

bi\se{}nd (\secost{}), b\nat{\i}nd (\natcost{}) & \\\hline

hav\si{e} (\sicost{}), hav\se{}e (\secost{}), h\pln{a}ve (\plncost{}) & \\\hline

microph\stst{o}ne (\ststcost{}), microph\nat{o}ne (\natcost{}), m\st{i}croph\stst{o}ne (\stcost{} + \ststcost{}), microph\selr{}one (\selrcost{}), micr\cnt{o}\selr{}phone (\restcost{}+\selrcost{}) & Note that some of these annotations are not equivalent (secondary stress or not) \\\hline

c\nat{o}m\si{b} (\natcost{}+\sicost{}), co\se{}m\si{b} (\secost{}+\sicost{}) & \\\hline

side\selr{}car (\selrcost{}), s\nat{\i}d\si{e}c\stst{a}r (\natcost{}+\sicost{}+\ststcost{}) & s\nat{\i}d\si{e}c\brd{a}r (\natcost{}+\sicost{}+\brdcost{}) does not make the `a' rhotic and it is invalid \\\hline

r\iot{e}c\st{e}\si{i}ved (\restcost{}+\stcost{}+\sicost{}), r\iot{e}c\nnat{\st{e}\i}ved (\restcost{}+\nnatcost{}+\stcost{})  & \\\hline

al\st{u}mn\nat{i} (\stcost{}+\secost{}) & al\st{u}mn\selr{}i (\stcost{}+\selrcost{}) is not valid, since the `i' is not stressed.\\\hline

underg\st{o} (\stcost{}), under\serl{}go (\serlcost{}), underg\serl{}o (\serlcost{}) & \\\hline

nightm\stst{a}re (\ststcost{})  & nightm\nat{a}re (\natcost{}) is not valid because the `a' should be rhotic. \\\hline

Birming\se{}\si{h}am (\secost{}+\sicost{})& Birmingh\se{}am (\secost{}) does not work.\\\hline

ag\st{o} (\stcost{}), a\ser{}go (\sercost{}), ag\ser{}o (\sercost{}) & \\\hline

fat\serl{}h\pln{e}\si{a}d\iot{e}d  & The separation is needed here \\\hline

t\pln{e}l\iot{e}ph\stst{o}ne (\plncost{}+\restcost{}+\ststcost{}), t\pln{e}l\iot{e}\selr{}phone (\plncost{}+\restcost{}+\selrcost{}).  & \\\hline

b\iot{e}l\bbrd{\st{\i{}}e}f (\restcost{}+\bbrdcost{}+\stcost{}), b\iot{e}l\si{i}\st{\nat{e}}f (\restcost{}+\sicost{}+\stcost{}+\natcost{}) &  \\\hline 
\hline
\end{tabular}
} % small
\vspace{5mm}
\caption{Examples of Annotation Costs (1/3)}
\label{tab:exacost1} 
\end{center}
\end{table}

\begin{table}
\begin{center}
{\small
\begin{tabular}{|| p{7cm} | p{6cm}  ||}
\hline
\bf{Type of annotation}                                                 & \bf{Comments}         \\ \hline \hline

y\oopq{ou}r (\rrestcost{}), y\si{o}\nat{u}r (\sicost{}+\natcost{}), y\rnd{o}\brd{u}r (\restcost{}+\brdcost{})    &  GA pronunciation \\\hline

pe\si{o}ple (\sicost{}), p\nnat{eo}ple (\nnatcost{})    &  \\\hline

thr\si{o}ugh (\sicost{}), thr\rnd{o}\si{u}gh (\restcost{}+\sicost{}), thr\si{o}\brd{u}gh (\sicost{}+\brdcost{})     &  \\\hline

y\si{o}\brd{u}th (\sicost{}+\brdcost{}), y\rnd{o}\si{u}th (\restcost{}+\sicost{}) &  \\\hline

fo\si{u}r (\sicost{}), f\bbrd{ou}r (\bbrdcost{})    &  \\\hline

h\si{e}ight (\sicost{}), h\iidp{e\i}ght (\restcost{})    &  \\\hline

\stst{o}utg\st{o}\se{}ing (\ststcost{}+\stcost{}+\secost{}), out\serl{}go\se{}ing (\serlcost{}+\secost{}), outg\serl{}o\se{}ing (\serlcost{}+\secost{})  , ou\serl{}tgo\se{}ing (\serlcost{}+\secost{})  &  \\\hline

eve\se{}ning (\secost{}), eve\sel{}ning (\selcost{}), \nat{e}v\si{e}ning (\natcost{}+\sicost{})    &  \\\hline

home\se{}less (\secost{}), home\sel{}less (\selcost{}), h\nat{o}m\si{e}less (\natcost{}+\sicost{})    &  \\\hline

bor\si{e}dom (\sicost{}), bore\se{}dom (\secost{}), bore\sel{}dom (\selcost{})    &  \\\hline

s\clr{o}m\si{e}\no{th}ing (\restcost{}+\sicost{}+\restcost{}), s\clr{o}me\se{}\no{th}ing (\restcost{}+\secost{}+\restcost{}), s\clr{o}m\si{e}th\se{}ing (\restcost{}+\sicost{}+\secost{})    &  \\\hline

\cnt{a}ny\no{th}ing (\restcost{}+\restcost{}), \cnt{a}ny\se{}\no{th}ing (\restcost{}+\secost{}+\restcost{}), \cnt{a}nyth\se{}ing (\restcost{}+\secost{})     &  \\\hline

s\clr{o}m\si{e}wh\brd{e}re (\restcost{}+\sicost{}+\brdcost{}), s\clr{o}me\se{}wh\brd{e}re (\restcost{}+\secost{}+\brdcost{})    &  \\\hline

no\se{}wh\brd{e}re (\secost{}+\brdcost{})    &  n\nat{o}wh\brd{e}re (\natcost{}+\brdcost{}) does not work, because `wh' is not at the beginning of a segment or after a consonant \\\hline

a\ser{}side (\sercost{}), a\no{s}\st{i}de (\restcost{}+\stcost{})    &  \\\hline

b\nat{\i}\ser{}sect (\natcost{}+\sercost{}), b\nat{\i}\no{s}\st{e}ct (\natcost{}+\restcost{}+\stcost{})    &  \\\hline

a\ser{}while (\sercost{})    &  awh\st{i}le (\stcost{}) doesn't work (`aw' digraph). \\\hline

\rnd{a}lth\st{o}\si{u}gh (\restcost{}+\stcost{}+\sicost{}), \rnd{a}l\ser{}tho\si{u}gh (\restcost{}+\sercost{}+\sicost{})  &   \\\hline

p\nat{\i}\co{r}ate (\natcost{}+\rcost{}), pi\se{}rate (\secost{})    &  `pirate' does not work in RP, since the `i' is not rhotic here. \\\hline

pa\co{r}ab\pln{\st{o}}lic (\restcost{}+\plncost{}+\stcost{}), p\stst{a}\se{}rab\pln{\st{o}}lic (\ststcost{}+\secost{}+\restcost{}+\plncost{}+\stcost{})   & \\\hline

t\bbrd{\i{}e}r (\bbrdcost{}), t\pln{\i}er {\plncost{}}    &  t\bbrd{\i{}e}r is valid for GA and RP, while t\pln{\i}er only for RP. \\\hline

f\brd{a}ther (\brdcost{}), f\rnd{a}ther (\restcost{})    &  f\rnd{a}ther is only useful for some American accents merging father and bother. \\\hline

Denm\stst{a}rk (\ststcost{}), Denm\brd{a}rk (\brdcost{}), Denm\brd{a}\si{r}k (\brdcost{}+\sicost{})    &  `Denm\brd{a}rk' does not work in RP, since the `a' needs stress to become rhotic. `Denm\brd{a}\si{r}k' does not work for GA \\\hline

d\nat{\i}r\st{e}ct (\natcost{}+\stcost{}), d\nat{\i}\ser{}rect (\natcost{}+\sercost{})    &  `di\serl{}rect' does not work, since `i' is not stressed. \\\hline

ear\si{r}ing (\sicost{}), ear\se{}ring (\secost{})   & The double `r' must be broken to ensure it is rhotic in RP \\\hline

tar\co{g}\iot{e}t (\restcost{}+\restcost{}), targ\se{}\iot{e}t (\secost{}+\restcost)   &  \\\hline

hou\co{s}e (\restcost{}), hou\se{}se (\secost{})   &  \\\hline

\nat{e}qual (\natcost{}), e\se{}qual (\secost{})   &  \\\hline

m\pln{a}cro (\plncost{}), mac\se{}ro (\secost{})   &  \\\hline

turb\nat{o}j\stst{e}t (\natcost{}+\ststcost{}), turbo\selr{}jet (\selrcost{})    &  \\\hline

in\ser{}cur (\sercost{}), i\no{n}c\st{u}r (\restcost{}+\stcost)  &  \\\hline

clo\no{s}\si{e}\stst{u}p (\restcost{}+\sicost{}+\ststcost{}), clo\no{s}e\selr{}up (\restcost{}+\selrcost{})  &  \\\hline

sci\vo{ss}ors (\rrestcost{}), sc\pln{\i}s\si{s}ors (\plncost{}+\sicost{})  &  \\\hline

% (\cost{}),  (\cost{})    &  \\\hline

\hline
\end{tabular}
} % small
\vspace{5mm}
\caption{Examples of Annotation Costs (2/3)}
\label{tab:exacost2} 
\end{center}
\end{table}

\begin{table}
\begin{center}
{\small
\begin{tabular}{|| p{7cm} | p{6cm}  ||}
\hline
\bf{Type of annotation}                                                 & \bf{Comments}         \\ \hline \hline

\pln{e}ver (\plncost{}), ev\se{}er (\secost{}), ev\si{e}r (\sicost{}+\introcost{})    &  The last option requires the introduction of schwa \\\hline

ev\cnt{i}l (\restcost{}), \nat{e}v\si{i}l (\natcost{}+\sicost)    &   \\\hline

dev\si{i}l (\sicost{}), d\pln{e}v\cnt{i}l (\plncost{}+\restcost{})     &   \\\hline

rhy\vo{th}m (\restcost{}), rhyth\sch{}m (\introcost{})     &   \\\hline

\iot{e}l\pln{\i}x\cnt{\i}r (\restcost{}+\plncost{}+\restcost{}), \iot{e}l\pln{\i}x\si{i}r (\restcost{}+\plncost{}+\sicost{}+\introcost{})     &   \\\hline

si\si{e}v\si{e} (\sicost{}+\sicost{}), s\pln{\i}\si{e}ve (\plncost{}+\sicost{})      &   \\\hline

be\si{a}uty (\sicost{}), b\si{e}\si{a}uty (\sicost{}+\sicost{})      &   \\\hline

fail\idp{u}re (\restcost{}), fail\y{}ure (\introcost{})      &   \\\hline

eur\nat{o}s\co{c}\st{e}ptic (\natcost{}+\restcost{}+\stcost{}), euro\serl{}s\co{c}eptic (\serlcost{}+\restcost{}), eur\nat{o}sc\serl{}eptic (\natcost{}+\serlcost{}),       &   \\\hline

th\stst{i}rt\st{e}en (\ststcost{}+\stcost{}), thir\serl{}teen (\serlcost{}), thirt\serl{}een (\serlcost{})      &   \\\hline

nine\serl{}teen (\serlcost{}), n\stst{\nat{\i}}n\si{e}t\st{e}en (\ststcost{}+\natcost{}+\sicost{}+\stcost{})   &   \\\hline

f\stst{o}r\si{e}\si{k}n\st{\pln{o}}\si{w}l\iot{e}dge (\ststcost{}+\sicost{}+\sicost{}+\stcost{}+\restcost{}+\sicost{}+\restcost{}), 
fore\serl{}kn\pln{o}\si{w}l\iot{e}dge (\serlcost{}+\plncost{}+\sicost{}+\restcost{})      &   \\\hline

Australo\serl{}pi\no{th}ecu\no{s} (\serlcost{}+\restcost{}+\restcost{}), \stst{A}ustral\nat{o}p\st{i}\no{th}ecu\no{s} (\ststcost{}+\natcost{}+\stcost{}+\restcost{}+\restcost{}) & \\\hline

frame\selr{}w\cnt{o}rk (\serlcost{}+\restcost{}), frame\se{}w\stst{\cnt{o}}rk (\secost{}+\ststcost{}+\restcost{}), fr\nat{a}m\si{e}w\stst{\cnt{o}}rk (\natcost{}+\sicost{}+\ststcost{}+\restcost{})  & \\\hline

thr\si{o}ugh\st{o}ut (\sicost{}+\stcost{}),  thr\si{o}\brd{u}gh\st{o}ut (\sicost{}+\brdcost{}+\stcost{}),  thr\rnd{o}\si{u}gh\st{o}ut (\restcost{}+\sicost{}+\stcost{}), thr\rnd{o}\si{u}gh\ser{}out (\restcost{}+\sicost{}+\sercost{}) & \\\hline

arr\st{\nat{a}}n\hvo{g}\si{e}ment (\natcost{}+\stcost{}+\restcost{}+\sicost{}), arr\st{\nat{a}}nge\se{}ment (\natcost{}+\stcost{}+\secost{})   & \\\hline

% (\cost{}),  (\cost{})    &  \\\hline

\hline
\end{tabular}
} % small
\vspace{5mm}
\caption{Examples of Annotation Costs (3/3)}
\label{tab:exacost3} 
\end{center}
\end{table}

\section{Some Examples}\label{sec:examples}

In this section we include several examples taken from different areas, in order to see how annotations look like in real text and also to observe the proportion of annotation in general texts vs. formal texts.

\subsection{200 Most Common Words}

\noindent 1--50:

the
o\vo{f}
t\rnd{o}
and
a
in
is
it
y\si{o}u
that
he
w\opq{a}s
for
on
ar\si{e}
wi\vo{th}
as
I
his
they
be
at
\w{}\clr{o}ne
hav\si{e}
thi\no{s}
from
or
had
by
hot
w\cnt{o}rd
but
wh\opq{a}t
s\clr{o}me
we
can
out
\clr{o}ther
wer\si{e}
\rnd{a}ll
th\brd{e}re
when
up
use (verb, the noun would be `u\no{s}e')
yo\si{u}r
h\uudp{ow}
s\cnt{a}\si{i}d
an
each
she

\vspace{5mm}

\noindent 51--100:

which
d\rnd{o}
their
time
if
will
way
ab\st{o}ut
m\cnt{a}ny
then
them
write
w\oopq{ou}\si{l}d
like
so
these
her
long
make
\no{th}ing
see
him
t\bbrd{wo}
has
l\oopq{oo}k
more
day
c\oopq{ou}\si{l}d
go
c\clr{o}me
did
number
sound
no
m\nat{o}st
pe\si{o}ple
my
over
know
w\rnd{a}ter
than
c\rnd{a}ll
first
\si{w}h\rnd{o}
may
d\uudp{ow}n
side
been
n\uudp{ow}
f\nat{\i}nd

\vspace{5mm}

\noindent 101--150:

\cnt{a}ny
new
w\cnt{o}rk
part
take
\co{g}et
place
made
liv\si{e}
wh\brd{e}re
after
back
little
\nat{o}nly
round
man
year
came
show
\pln{e}v\si{e}ry
g\oopq{oo}d
me
giv\si{e}
our
under
name
ve\co{r}y
thr\si{o}ugh
just
form
sentence
gr\bbrd{ea}t
think
say
help
low
line
differ
turn
cause
much
mean
b\iot{e}f\st{o}re
m\rnd{o}ve
right
boy
\nat{o}ld
too
same
tell

\vspace{5mm}

\noindent 151--200:

d\clr{o}es
set
three
w\opq{a}nt
air
well
\rnd{a}lso
play
sm\rnd{a}ll
end
p\opq{u}t
home
read (the past participle is `re\si{a}d')
hand
port
large
spell
add
even
land
here
must
big
high
such
follow
act
why
ask
men
ch\nat{a}nge
went
light
k\nat{\i}nd
off
need
hou\no{s}e
pic\hno{t}ure
try
u\no{s}
ag\st{\cnt{a}}\si{i}n
\pln{a}nimal
point
m\clr{o}ther
w\cnt{o}rld
near
b\si{u}ild
self
e\si{a}rth
f\brd{a}ther

\subsection{Miscellanea}

\noindent Numeral Numbers:

\w{}\clr{o}ne
t\bbrd{wo}
three
fo\si{u}r
five
six
s\pln{e}ven
eight
nine
ten
\iot{e}l\st{\pln{e}}ven
twelve
th\stst{i}rt\st{e}en
f\stst{o}\si{u}rt\st{e}en
f\stst{i}ft\st{e}en
s\stst{i}xt\st{e}en
s\st{\pln{e}}vent\st{e}en,
\stst{e}ight\st{e}en
nine\serl{}teen
twenty
thirty
forty
fifty
sixty
s\pln{e}venty
eighty
nine\sel{}ty
hundr\iot{e}d
\no{th}ousand
mill\y{i}on

\vspace{5mm}

\noindent Ordinal Numbers:

first
s\pln{e}cond
\no{th}ird
fo\si{u}rth
fifth
sixth
s\pln{e}venth
eigh\co{th}
n\nat{\i}nth
tenth
\iot{e}l\st{\pln{e}}venth
twelfth
th\stst{i}rt\st{e}enth
...
twentieth
thirtieth
hundre\si{d}th
thousan\si{d}th
mill\y{i}onth

\vspace{5mm}

\noindent Days o\vo{f} the week:

M\clr{o}nday
T\nnat{ue}sday
We\si{d}n\si{e}sday
\no{Th}ursday
Friday
S\pln{a}turday
Sunday

\vspace{5mm}

\noindent C\clr{o}lours:

Black
white
red
blue
green
yellow
pink
violet
gray/grey
purple
o\co{r}\iot{a}nge
br\uudp{ow}n
cyan
mag\st{e}nta
turqu\st{o}ise. 

\vspace{5mm}

\noindent Solar system:

Sun
Moon
E\si{a}rth
Merc\udp{u}ry
Venu\no{s}
Mars
Jupiter
S\pln{a}turn
Uranu\no{s}
Pluto  

\vspace{5mm}

\noindent Places:

Europe
Am\st{e}\co{r}ica
\pln{A}frica
A\sno{si}a
O\sno{c}\iot{e}\st{\brd{a}}nia
\pln{A}nt\st{a}rctica
\nat{U}n\st{i}t\iot{e}d States o\vo{f} Am\st{}e\co{r}ica
\iot{E}ngland
Br\pln{\i}ta\si{i}n
Wales
Scotland
Ire\sel{}land
\oopq{Au}str\st{a}lia
New Zealand
India
C\pln{a}nada
Spain
Fr\brd{a}nce
Germany
Denm\stst{a}rk
Holland
\pln{I}taly
China
Braz\st{i}l  
Jap\st{a}n
Mexic\stst{o}
Ru\sno{\group{ssi}}a
P\brd{a}kist\st{\brd{a}}n
L\clr{o}ndon
New York
Birming\se{}\si{h}am
\pln{E}dinbur\oopq{gh}
Pa\co{r}is
Rome
Berl\st{i}n
Barc\iot{e}l\st{o}na

\vspace{5mm}

\noindent 

\subsection{A Literary Work. A Tale of Two Cities}

A Tale o\vo{f} t\bbrd{wo} C\pln{\i}ties by Charles Dick\iot{e}ns

\vspace{5mm}

\noindent B\oopq{oo}k the First, Chapter I: The Period

\vspace{5mm}

It w\opq{a}s the best o\vo{f} times, it w\opq{a}s  the w\cnt{o}rst o\vo{f} times, it w\opq{a}s  the age o\vo{f} wisdom, it w\opq{a}s the age o\vo{f} foolishness, it w\opq{a}s  the epoc\si{h} o\vo{f} b\iot{e}l\bbrd{\st{\i}e}f, it w\opq{a}s the  epoc\si{h} o\vo{f} incred\st{u}lity, it w\opq{a}s  the season o\vo{f} Light, it w\opq{a}s the season o\vo{f} Darkness, it w\opq{a}s  the spring o\vo{f} hope, 
it w\opq{a}s  the winter o\vo{f} d\iot{e}sp\st{a}ir, we had \pln{e}v\si{e}ry\no{th}ing b\iot{e}f\st{o}re u\no{s}, we had n\clr{o}\no{th}ing b\iot{e}f\st{o}re u\no{s}, we wer\si{e} \rnd{a}ll go\sel{}ing d\nat{\i}r\st{}ect t\rnd{o} H\pln{e}\si{a}ven, we wer\si{e} \rnd{a}ll go\sel{}ing d\nat{\i}r\st{}ect the \clr{o}ther way -- in short, the period w\opq{a}s so far like the pr\pln{e}sent period, that s\clr{o}me o\vo{f} its noisiest auth\st{o}\co{r}ities ins\st{i}st\iot{e}d on its be\sel{}ing r\iot{e}c\st{e}\si{i}ved, for g\oopq{oo}d or for ev\cnt{i}l, in the s\brd{u}p\st{e}rlative d\iot{e}gr\st{e}e o\vo{f} comp\st{a}\co{r}i\no{s}on \nat{o}nly.

      Th\brd{e}re w\opq{a}s a king wi\vo{th} a large jaw and a queen with a plain face, on the throne o\vo{f} \iot{E}ngland; th\brd{e}re wer\si{e} a king with a large jaw and a queen wi\vo{th} a fair face, on the throne o\vo{f} Fr\brd{a}nce. In b\nat{o}th c\si{o}untries it w\opq{a}s clearer than crystal t\rnd{o} the lords o\vo{f} the State pr\iot{e}s\st{e}rves o\vo{f} loaves and fish\iot{e}s, that \no{th}ings in g\pln{e}neral wer\si{e} settled for \pln{e}ver.
      
      It w\opq{a}s the year o\vo{f} Our Lord \w{}\clr{o}ne \no{th}ousand s\pln{e}ven hundred and s\pln{e}venty-five. Spi\co{r}it\udp{u}al r\pln{e}vela\sno{ti}ons wer\si{e} conc\st{e}d\iot{e}d t\rnd{o} \iot{E}ngland at that favoured period, as at thi\no{s}. Mrs. Southc\pln{o}tt had recently att\st{a}ined her five-and-twentieth blessed birthday, o\vo{f} \si{w}h\rnd{o}m a proph\st{\pln{e}}tic private in the Life G\si{u}ards had he\co{r}ald\iot{e}d the subl\st{i}me app\st{e}arance by ann\st{o}uncing that arr\st{\nat{a}}nge\se{}ment (\natcost{}+\stcost{}+\secost{}) wer\si{e} made for the sw\opq{a}llowing up o\vo{f} L\clr{o}ndon and Westm\stst{i}nster. Even the Cock-lane g\si{h}\nat{o}st had been laid \nat{o}nly a round d\clr{o}zen o\vo{f} years, after rapping out its mess\iot{a}ges, as the spi\co{r}its o\vo{f} thi\no{s} ve\co{r}y year l\brd{a}st p\brd{a}st (s\brd{u}perna\hno{t}urally d\iot{e}fi\sno{ci}ent in originality) rapped out theirs. Mere messag\iot{e}s in the e\si{a}rthly order o\vo{f} \iot{e}v\st{e}nts had late\se{}ly c\clr{o}me t\rnd{o} the \iot{E}nglish Cr\uudp{ow}n and Pe\si{o}ple, from a congr\pln{e}ss of Br\pln{\i}tish subj\iot{e}cts in Ame\co{r}ica: which, str\nat{a}nge t\rnd{o} r\iot{e}l\st{a}te, hav\si{e} pr\rnd{o}ved more imp\st{o}rtant t\rnd{o} the human race than \cnt{a}ny communica\sno{ti}ons yet r\iot{e}c\st{e}\si{i}ved thr\si{o}ugh \cnt{a}ny o\vo{f} the chick\iot{e}ns o\vo{f} the Cock-lane brood. Fr\brd{a}nce, less favoured on the \si{w}hole as t\rnd{o} matters spi\co{r}it\udp{u}al than her sister o\vo{f} the sh\bbrd{\i{}e}ld and trident, r\nat{o}lled wi\vo{th} \iot{e}xc\st{e}eding smoo\vo{th}ness d\uudp{ow}n hill, making paper m\clr{o}ney and spending it. Under the guidance of her C\si{h}ris\hno{ti}an p\brd{a}stors, she entert\st{a}ined h\stst{e}rs\st{e}lf, b\iot{e}\ser{}sides, wi\vo{th} such h\nat{u}m\st{a}ne ach\bbrd{\st{\i{}}e}v\si{e}ments as sentencing a y\si{o}\brd{u}th  t\rnd{o} hav\si{e} his hands cut off, his t\clr{o}ngue torn out wi\vo{th} pincers, and his b\pln{o}dy burned al\st{i}ve, b\iot{e}c\st{a}use he had not kneeled d\uudp{ow}n in the rain t\rnd{o} d\rnd{o} \si{h}\pln{o}nour t\rnd{o} a dirty proce\sno{\group{ssi}}on o\vo{f} m\clr{o}nks which p\brd{a}sse\no{d} with\st{i}n his view, at a distance o\vo{f} s\clr{o}me fifty or sixty yards. It is like\se{}ly \iot{e}n\si{o}\st{u}\co{gh} that, root\iot{e}d in the w\oopq{oo}ds of Fr\brd{a}nce and Norway, th\brd{e}re wer\si{e} growing trees, when that sufferer w\opq{a}s p\opq{u}t t\rnd{o} de\si{a}th, \rnd{a}lr\st{\pln{e}}\si{a}dy marked by the W\oopq{oo}dman, Fate, to c\clr{o}me d\uudp{ow}n and be sawn int\rnd{o} boards, t\rnd{o} make a certa\si{i}n m\rnd{o}vable frame\selr{}w\cnt{o}rk wi\vo{th} a sack and a knife in it, terr\cnt{i}ble in history. It is like\se{}ly \iot{e}n\si{o}u\co{gh} that in the r\si{o}u\co{gh} outhou\no{s}\iot{e}s o\vo{f} s\clr{o}me tillers o\vo{f} the h\pln{e}\si{a}vy lands adj\st{a}cent t\rnd{o} Pa\co{r}is, th\brd{e}re wer\si{e} sheltered from the we\si{a}ther that ve\co{r}y day, rude carts, b\iot{e}sp\st{a}ttered wi\vo{th} rustic mire, snuffed ab\st{o}ut by pigs, and roost\iot{e}d in by p\nnat{ou}ltry, which the Farmer, De\si{a}th, had \rnd{a}lr\st{\pln{e}}\si{a}dy set ap\st{a}rt t\rnd{o} be his tumbr\cnt{i}ls o\vo{f} the R\pln{e}volu\sno{ti}on. But that W\oopq{oo}dman and that Farmer, tho\si{u}gh they w\cnt{o}rk \pln{u}nc\st{e}a\no{s}ingly, w\cnt{o}rk silently, and no \w{}\clr{o}ne he\si{a}rd them as they went ab\st{o}ut wi\vo{th} muffled tre\si{a}d: the r\brd{a}ther, forasm\st{u}ch as t\rnd{o} entert\st{a}in \cnt{a}ny susp\pln{\i}\sno{ci}on that they wer\si{e} aw\st{a}ke, w\opq{a}s t\rnd{o} be \nat{a}\no{th}\iot{e}\stst{i}stical and traitorous.

\subsection{Songs, Poems and Sayings}

\subsubsection{Rock-a-bye Baby}:

Rock-a-bye baby, on the treet\pln{o}p,

When the wind blows, the cradle will rock,

When the bough br\bbrd{ea}ks, the cradle will f\rnd{a}ll,

And d\uudp{ow}n will c\clr{o}me baby, cradle and \rnd{a}ll.

\subsubsection{\rnd{A}ll the pr\iot{e}tty hors\iot{e}s}:

Hush-a-bye, d\nat{o}n't y\si{o}u cry,

Go t\rnd{o} sleepy y\si{o}u little baby.

When y\si{o}u wake, y\si{o}u shall hav\si{e} cake,

And \rnd{a}ll the pr\iot{e}tty little hors\iot{e}s.

Blacks and bays, dapples and greys,

Go t\rnd{o} sleepy y\si{o}u little baby,

Hush-a-bye, d\nat{o}n't y\si{o}u cry,

Go t\rnd{o} sleepy little baby.

Hush-a-bye, d\nat{o}n't y\si{o}u cry,

Go t\rnd{o} sleepy little baby,

When y\si{o}u wake, y\si{o}u shall hav\si{e},

\rnd{A}ll the pr\iot{e}tty hors\iot{e}s.

Way d\uudp{ow}n yonder, d\uudp{ow}n  in the m\pln{e}\si{a}dow,

Th\brd{e}re's a p\oopq{oo}r\footnote{`po\si{o}r' in RP.} wee little l\pln{a}m\si{b}y.

The bees and the butterfl\stst{i}es pickin' at its \iidp{ey}es,

The p\oopq{oo}r\footnote{`po\si{o}r' in RP.} wee \no{th}ing cried for her mammy.

Hush-a-bye, d\nat{o}n't y\si{o}u cry,

Go t\rnd{o} sleepy little baby.

When y\si{o}u wake, y\si{o}u shall hav\si{e} cake,

And \rnd{a}ll the pr\iot{e}tty little hors\iot{e}s.

\subsection{A Formal Text. Human Rights}

Univ\st{e}rsal D\pln{e}clara\sno{ti}on o\vo{f} Human Rights

Ad\st{o}pt\iot{e}d and procl\st{a}imed by G\pln{e}neral Ass\st{e}mbly r\pln{e}solu\sno{ti}on 217 A (III) o\vo{f} 10 D\iot{e}c\st{e}mber 1948
On D\iot{e}c\st{e}mber 10, 1948 the G\pln{e}neral Ass\st{e}mbly o\vo{f} the \stst{U}n\st{i}t\iot{e}d Na\sno{ti}ons ad\st{o}pt\iot{e}d and procl\st{a}imed  the Univ\st{e}rsal D\pln{e}clara\sno{ti}on o\vo{f} Human Rights, the f\opq{u}ll text of which app\st{e}ars in the following pag\iot{e}s. Following thi\no{s} hist\st{o}\co{r}ic act the Ass\st{e}mbly c\rnd{a}lled up\st{o}n \rnd{a}ll Member c\si{o}untries t\rnd{o} p\pln{u}blic\stst{i}ze the text o\vo{f} the D\pln{e}clara\sno{ti}on and t\rnd{o} cause it t\rnd{o} be diss\st{\pln{e}}min\stst{a}t\iot{e}d, displ\st{a}yed, re\si{a}d and \iot{e}xp\st{o}und\iot{e}d principally in sc\si{h}ools and \clr{o}ther {\pln{e}duca\sno{ti}onal institu\sno{ti}ons, with\st{o}ut distinc\sno{ti}on ba\no{s}ed on the pol\st{\pln{\i}}tical status o\vo{f} c\si{o}untries or territories."

\subsection{A Scientific Text}

\nonl{}First section of the article `Australopithecus' (retrieved from the English Wikipedia on August 25th, 2010).\nonr{}

Australo\serl{}pi\no{th}ecu\no{s} (L\pln{a}tin {\em \nonl{}australis\nonr{}} ``s\si{o}uthern'', Greek $\pi{}\iota\theta\eta\kappa\omicron\sigma$ {\em \nonl{}pithekos\nonr{}} ``ape") is a genu\no{s} o\vo{f} h\pln{o}minids that ar\si{e} n\uudp{ow} \iot{e}xt\st{i}nct. From the \pln{e}vidence gathered by p\stst{\pln{a}}l\iot{a}\si{e}\pln{o}nt\st{\pln{o}}logists and arc\si{h}\iot{a}e\serl{}\pln{o}logists, it app\st{e}ars that the Australo\serl{}pi\no{th}ecu\no{s} genu\no{s} \iot{e}v\st{o}lved in eastern \pln{A}frica ar\st{o}und 4 mill\y{i}on years ag\st{o} b\iot{e}f\st{o}re spr\pln{e}\si{a}ding thr\si{o}ugh\st{o}ut  the continent and \iot{e}v\st{e}nt\udp{u}ally bec\pln{o}ming \iot{e}xt\st{i}nct 2 mill\y{i}on years ag\st{o}. 

%\nonl{}During this time period various different forms of australopiths existed, including Australopithecus anamensis, A. af\st{a}r\serl{}en\no{s}i\no{s}, A. sediba, and A. africanus. There is still some debate amongst academics whether certain African hominid species of this time, such as A. robustus and A. boisei, constitute members of the same genus; if so, they would be considered to be robust australopiths whilst the others would be considered gracile australopiths. However, if these species do indeed constitute their own genus, then they may be given their own name, the Paranthropus.

%It is widely held by archaeologists and palaeontologists that the australopiths played a significant part in human evolution and it was one of the australopith species that eventually evolved into the Homo genus in Africa around 2 million years ago, which contained within it species like Homo habilis, H. ergaster and eventually the modern human species, H. sapiens sapiens.\nonr{}

\section{Conclusions}\label{sec:conclusion}

This paper has presented \annl{}Annot\st{a}t\iot{e}d \iot{E}nglish\annr{}, a proposal of diacritical annotations which converts English into a language whose pronunciation can be unambiguously inferred from its spelling (plus annotations). The novel principle, which distinguishes this from any spelling reform, is that we set as an unbreakable principle that no single letter of English can be modified.
Since English has virtually no diacritics (only in some loanwords, such as résumé, crème, piñata, etc., whose diacritics would be eliminated prior to the annotation process) we can use diacritics to annotate the pronunciation. \annl{}Annot\st{a}t\iot{e}d \iot{E}nglish\annr{} differs from a simple pronunciation without respelling (partially existing in some old spelling books, dictionaries and specialised books) in the fact that we introduce a set of rules which incorporate most of the usual pronunciation rules in the English language. In this way we not only minimise the number of annotations, but they appear when the reader has the perception that the part of the word which is annotated is an exception to a rule.

The proposal includes a set of rules and many annotation symbols. Rules include how vowels, vowel digraphs, semiconsonants, consonants, and consonant digraphs must be pronounced. A set of about 20 annotation symbols has been necessary, especially because there are some characters in English (most especially vowels) which can have about 10 different sounds (without considering rhotic sounds). A first impression might be that the set of rules and the set of annotations are large and complex. We have to make some comments on this.

\begin{itemize}
\item Rules: most rules we have introduced here (double-consonant rule to distinguish between plain and natural vowels, vowel digraphs, consonant pronunciation, final e, final ed, rhotic sounds, etc.) are just the same (or very similar) rules every speaker of English knows. This document just places everything together. For an average Ejnglish speaker, a short leaflet or schematic annotation table can be constructed as a mnemonic for beginners to the annotation system.
\item Symbols: we do not need to know all the symbols precisely in order to guess the right sound. For instance, if we see `w\cnt{o}rd', we can infer that it is not like `lord', even ignoring what the sound of `\cnt{o}' is. In fact, when a new word is read, the doubt is typically between two possibilities: the regular and the irregular way. The annotation just sorts this out. 
\end{itemize}

\noindent Given an annotated text, a reader can take advantage from annotations at several degrees. They can be ignored, and the text can be read as any other plain text. It can be used to distinguish regular cases from exceptions, or to place the stress in some words. Many annotations are intuitive or just obvious, such as the deletion symbol, as in `de\si{b}t', so even the first time an annotated text is used, there are some basic understanding. More advanced (or interested) readers will soon make the small effort to take a look at the rules or to infer them by reading. In the end, if we have been able to infer thousands of pronunciation exceptions in the English language, we can also manage with a set of rules.

Having said this, there is of course an interesting discussion about the trade-off between number of rules and percentage of annotations.
One of the criteria we have followed is that the rules should be tuned to favour the most common words. That is the reason-why we have preferred a voiced `th' and `s' in some situations (instead of a voiceless one as the by-default case for every position). Even though we need two rules for `th' and `s', it saves many annotations in frequent words.
In the appendix we discuss some other rules we did not finally include, because we thought (and think) that the set of rules must be simple and easy to learn by average people.
Nonetheless, the exclusion or inclusion of rules can be done after a proper frequency analysis (this is what we have done to a greater or lesser degree for the introduction of our rules). In any case, the analysis of any old or new rule must take into account the frequency of words\footnote{A very good resource is http://en.wiktionary.org/wiki/Wiktionary:Frequency\_lists}, and very especially the most frequent words.
According to \cite{Fry-et-al1993}, the first 25 most common words make up about one-third of all printed material in English. The first 100 make up about one-half of all written material, and the first 300 make up about sixty-five percent of all written material in English.

In Table \ref{tab:statistics} and Table \ref{tab:statistics2}  we show some preliminary statistics (computed from the first five paragraphs for the entry `bed' in wikipedia in different languages) where we see that the proportion of annotations is high. Nevertheless there are some European languages using the Roman alphabet with a higher annotation/diacritics ratio. This means two things. First, English spelling is not so irregular, provided a consistent set of rules is defined for it. Second, we should not obsess about reducing the annotation ratios by introducing many more rules, since the values are comparable to other languages which use diacritics and their speakers are relatively happy with them. Additionally, the annotation process will be automatic, so we should not care about {\em writing} in \annl{}Annot\st{a}t\iot{e}d \iot{E}nglish\annr{}.

\begin{table}
\begin{center}
{\small
\begin{tabular}{|| c | c  ||}
\hline
\bf{Indicator}                                                 & \bf{Value}         \\ \hline \hline
Annotations per Letter                  & 1/12                 \\\hline
Annotations per Word                    & 1/3                 \\\hline
Annotated Word per Word                 & 1/4                 \\\hline
\hline
\end{tabular}
} % small
\vspace{5mm}
\caption{Some preliminary statistics for Annotated English.}
\label{tab:statistics} 
\end{center}
\end{table}

\begin{table}
\begin{center}
{\small
\begin{tabular}{|| c | c  ||}
\hline
\bf{Annotations/Diacritics per Letter}                                                 & \bf{Value}         \\ \hline \hline
Annotated English                 & 1/12                 \\\hline
Czech                    & 1/9                 \\\hline
Polish                 & 1/12                 \\\hline
Turkish                 & 1/15                 \\\hline
Swedish                 & 1/20                 \\\hline
Danish                  & 1/25                 \\\hline
French                 & 1/30                 \\\hline
Spanish                 & 1/40                 \\\hline
\hline
\end{tabular}
} % small
\vspace{5mm}
\caption{Some preliminary statistics for Annotated English compared to the diacritical systems used in other languages.}
\label{tab:statistics2} 
\end{center}
\end{table}

Another criticism against \annl{}Annot\st{a}t\iot{e}d \iot{E}nglish\annr{} is that if we get used to it, then we will be unable to read in plain English. First of all, if we get used and feel comfortable with something, that means that it is useful, which is the main goal of this proposal. According to current computer technology, it should not be a problem to annotate documents on demand. In any case, if one thinks that some people get a strong habit to \annl{}Annot\st{a}t\iot{e}d \iot{E}nglish\annr{} that it seems impossible to deal with plain English language again, we can take a look at the experience with other languages. For instance, in Spanish, a text without (or with wrong) diacritics can irritate many, but it is still well read and understood by all.

One of the issues we have not addressed in detail is how to annotate words which are pronounced differently in several dialects (some of them very common, such as `from' , `of', `real'). As we said in the introduction, the annotation for a text must choose one dialect. But it is important to be consistent. The entire document must be annotated with the same dialect and it should also be consistent with the spelling (e.g. it would be strange to see `fast t\brd{u}mour' or `f\brd{a}st tumor'). Of course different annotations can be used in a novel or a conversation when people with different accents participate. In some cases, an annotation is valid for more dialects than a more specific one. For instance, `t\bbrd{\i{}e}r' is valid for both GA and RP, while `t\pln{\i}er' is only valid for RP.

Typography for \annl{}Annot\st{a}t\iot{e}d \iot{E}nglish\annr{} is of course a matter of taste. There are of course many other choices for the symbols. Our criterion has been to use frequent symbols that can be easily recognised, and to avoid symbols which can be confused with others. The distinction between upper annotations (vowels) and bottom annotations (consonants) is also the result of our criterion for clarity.
It is important to highlight that any annotation covering two or more letters can be reduced to an annotation for one letter plus deletions. This makes typography much easier, since we can use single letter accents in text processors and editors, as well as in environments where we cannot play freely with the symbols as in \LaTeX{} (the text processor we have used to produce this document).

Finally, even though \annl{}Annot\st{a}t\iot{e}d \iot{E}nglish\annr{} is not a spelling reform, it could be used as a basis for it, or at least for a spelling reform based to reduce the number of annotations (e.g. `ar\si{e}' $\rightarrow$ `ar') or the most awkward ones (`M\clr{a}g\si{h}reb' $\rightarrow$ M\pln{u}greb), in such a way that the words that require the rarest annotation symbols could be modified first. Nonetheless, we will not pursue this idea further, since we are not convinced that a spelling reform is a good thing for the English language, as we have mentioned in the introduction.

The appendices include some ideas on rules we excluded, but also some ideas on the use of diacritical accents for homographs, as well as some notes on \LaTeX{}. Nonetheless we have plenty of work to be done before going back to these issues. Let us just enumerate some of the possible future work:

\begin{itemize}
\item Construct an automatic interpreter (a reader) for \annl{}Annot\st{a}t\iot{e}d \iot{E}nglish\annr{}, taking an annotated text (using, e.g. a notation such as the  \LaTeX{} commands used here or an XML approach) and converting it into an IPA representation.
\item Construct an automatic annotator. This is a more ambitious project, that would require dynamic programming techniques (to choose the most efficient annotation according to the costs) and also some disambiguation techniques. It also requires access to a corpus of IPA pronunciations for thousand of words (there are some which are freely accessible on-line).
\item Analysis of the rules by the reckoning of statistics for each rule (ratio of annotations using it or not using it). This task can be boosted if we have an annotator. An interesting approach for this analysis could be made by using the Minimum Message Length (MML) principle \cite{Wallace-Boulton68}\cite{Wallace05}, where an MML coding would be used to transmit the pronunciation of an English text given its spelling. In other words, if sender and receiver share an English text and the receiver does not have any notion about English pronunciation, how can we devise a code such that we minimise the message including the pronunciation information? MML coding does not care about intuitive simplicity (only information efficiency) and  it is very sensitive to the size of the documents (for infinite documents, everything would be embedded into rules) but it is an interesting idea as a principle.
\item Incorporate \annl{}Annot\st{a}t\iot{e}d \iot{E}nglish\annr{} in an English for foreigners programme, or in a pilot programme using material following the annotations.
\item Incorporate \annl{}Annot\st{a}t\iot{e}d \iot{E}nglish\annr{} in a spelling learning programme (e.g. for children), or in a pilot programme using material following the annotations.
\end{itemize}

\noindent Only after these tasks we will be able to evaluate the real usefulness and impact of this proposal.

\section*{Acknowledgements} 
I would like to thank some useful and encouraging comments from Sergio España, Carlos Monserrat, Cèsar Ferri, Andrés Terrasa and David L. Dowe.
%\end{acknowledgements}

\newpage

\appendix

\section{Appendix. Alternative Rules and Additional Symbols}

In this appendix we discuss some rules that could be included or excluded. We also comment on the reasons why we finally took the decision of excluding or including them, respectively. This appendix is only recommended for people who wants to find explanations for some of the choices or wants to open a discussion on some of the rules.

\subsection{The $n$ Most Common Words}

A very effective way of reducing annotations is to insert the $n$ most common words into the pronunciation (interpretation) rules. We are against this option in principle, since the good thing about having annotations on common words is that they help learn the pronunciation of other, less frequent, words with the same annotation. 

For instance, the five most common words according to \cite{1000words} are: (the, of, to, and, a). If we eliminate annotations for the five of them, we have significant reduction in the annotation ratio, as we discussed in the conclusions.

Perhaps we can treat differently those words which change from a strong from and a weak form. In fact, the first most common words have this property, as shown in Table \ref{tab:five}. The use of a table such as this as a rule (keeping these five words without annotation) would reduce the ratio of annotations from 1/12 letters to 1/13 letters (approx).

\begin{table}
\begin{center}
{\small
\begin{tabular}{|| c | p{4cm}  | p{7cm}  ||}
\hline
\bf{Word}                                             & \bf{Strong}      & \bf{Weak}         \\ \hline \hline
the & the (\textipa{[Di:]}) (before vowel) & (unstressed) the (\textipa{[D@]})   \\\hline
of  & o\vo{f} (\textipa{[6v]}), \clr{o}\vo{f} (\textipa{[2v]}),  & (unstressed) o\vo{f} (\textipa{[@v]}), (unstressed) of / o\si{f}(\textipa{[@f]/[@]}) (before voiceless consonant)   \\\hline
to  & t\rnd{o} (\textipa{[tu:]}), t\opq{o} (\textipa{[tU]}) & (unstressed) to (\textipa{[t@]}), t\opq{o} (\textipa{[tU]}), (unstressed) \co{t}o (\textipa{[R@]})   \\\hline
and & and (\textipa{[\ae{}nd]}) & (unstressed)  and (\textipa{[@nd]}), (unstressed) an\si{d} (\textipa{[@n]}), \si{a}n\si{d} (\textipa{[n]})   \\\hline
a   & a (\textipa{[eI]}) & (unstressed) a (\textipa{[@]})   \\\hline
\hline
\end{tabular}
} % small
\vspace{5mm}
\caption{Strong and weak forms for five most common words}
\label{tab:five} 
\end{center}
\end{table}

\subsection{The ed/es Rule}

The ed/es rule could be removed, simplified or extended. For instance, it could be extended to cover the regular case where `ed' and `es' are pronounced \textipa{[Id]} and \textipa{[Iz]}, which is in the rules of past participle formation and plural formation. For instance, `beated' and `bridges' would be left unannotated. We have not included this (even it is well-known by speakers) because we want to stress the distinction between `loaves' and `bridg\iot{e}s', and `beat\iot{e}d' and `stuffe\no{d}'.

Another issue could be to convert `ed' into 't' when preceded by a voiceless consonant group (to avoid the annotation in `stuffe\no{d}'). This would be consistent with what we do with the `s', where we do not need annotations for `loaves' and `rats' even though the `s' are voiced and voiceless respectively.

We consider the combination we present here more consistent, but we understand that this is of course open for further discussion and analysis.

\subsection{The Rules for Natural and Plain Positions}

Given the rules for natural and plain positions, it is easy to realise that many more natural positions require a plain annotation than otherwise. In fact, the vowels in natural position in technical or new words are rarely pronounced with the natural sound.

An interesting study would be to simplify the rule and consider all positions plain, except when stressed at the end of a segment. For instance, words such as `be' and `go' would be considered natural positions, but `plane', `code', etc., would be considered plain positions, so requiring an annotation. Although this looks contrary to the traditional English spelling and pronunciation rules, it is competitive in terms of annotations. For instance, Table \ref{tab:tradeoff2} shows a text with the annotation system we have presented in this document on the left, and it shows the same text with the modification of the natural position rule on the right.

\begin{table}
\begin{center}
{\small
\begin{tabular}{|| p{6cm} || p{6cm}    ||}
\hline
\bf{Annotated text with the rules used in this document} & \bf{Annotated text with the same rules but the natural position rule} \\ \hline \hline

It w\opq{a}s the best o\vo{f} times, it w\opq{a}s  the w\cnt{o}rst o\vo{f} times, it w\opq{a}s  the age o\vo{f} wisdom, it w\opq{a}s the age o\vo{f} foolishness, it w\opq{a}s  the epoc\si{h} o\vo{f} b\iot{e}l\bbrd{\st{\i}e}f, it w\opq{a}s the  epoc\si{h} o\vo{f} incred\st{u}lity, it w\opq{a}s  the season o\vo{f} Light, it w\opq{a}s the season o\vo{f} Darkness, it w\opq{a}s  the spring o\vo{f} hope, 
it w\opq{a}s  the winter o\vo{f} d\iot{e}sp\st{a}ir, we had \pln{e}v\si{e}ry\no{th}ing b\iot{e}f\st{o}re u\no{s}, we had n\clr{o}\no{th}ing b\iot{e}f\st{o}re u\no{s}, we wer\si{e} \rnd{a}ll go\sel{}ing d\nat{\i}r\st{}ect t\rnd{o} H\pln{e}\si{a}ven, we wer\si{e} \rnd{a}ll go\sel{}ing d\nat{\i}r\st{}ect the \clr{o}ther way -- in short, the period w\opq{a}s so far like the pr\pln{e}sent period, that s\clr{o}me o\vo{f} its noisiest auth\st{o}\co{r}ities ins\st{i}st\iot{e}d on its be\sel{}ing r\iot{e}c\st{e}\si{i}ved, for g\oopq{oo}d or for ev\cnt{i}l, in the s\brd{u}p\st{e}rlative d\iot{e}gr\st{e}e o\vo{f} comp\st{a}\co{r}i\no{s}on \nat{o}nly.
&
It w\opq{a}s the best o\vo{f} t\nat{\i}mes, it w\opq{a}s  the w\cnt{o}rst o\vo{f} t\nat{\i}mes, it w\opq{a}s  the \nat{a}ge o\vo{f} wisdom, it w\opq{a}s the \nat{a}ge o\vo{f} foolishness, it w\opq{a}s  the \nat{e}poc\si{h} o\vo{f} b\iot{e}l\bbrd{\st{\i}e}f, it w\opq{a}s the  \nat{e}poc\si{h} o\vo{f} incred\st{\nat{u}}lity, it w\opq{a}s  the season o\vo{f} Light, it w\opq{a}s the season o\vo{f} Darkness, it w\opq{a}s  the spring o\vo{f} h\nat{o}pe, 
it w\opq{a}s  the winter o\vo{f} d\iot{e}sp\st{a}ir, we had ev\si{e}ry\no{th}ing b\iot{e}f\st{o}re u\no{s}, we had n\clr{o}\no{th}ing b\iot{e}f\st{o}re u\no{s}, we were \rnd{a}ll go\sel{}ing d\nat{\i}r\st{}ect t\rnd{o} He\si{a}ven, we were \rnd{a}ll go\sel{}ing d\nat{\i}r\st{}ect the \clr{o}ther way -- in short, the p\nat{e}riod w\opq{a}s so far l\nat{\i}ke the present p\nat{e}riod, that s\clr{o}me o\vo{f} its noisiest auth\st{o}\co{r}ities ins\st{i}st\iot{e}d on its be\sel{}ing r\iot{e}c\nnat{\st{e}\i}ved, for g\oopq{oo}d or for \nat{e}v\cnt{i}l, in the s\brd{u}p\st{e}rlative d\iot{e}gr\st{e}e o\vo{f} comp\st{a}\co{r}i\no{s}on \nat{o}nly. \\ \hline \hline
\end{tabular}
} % small
\vspace{5mm}
\caption{Difference between an annotated text with the rules explained in this document (left) and with a modification on the natural position rule (right)}
\label{tab:tradeoff2} 
\end{center}
\end{table}

While in the previous text the affected annotations are 11 against and 5 in favour, in other kinds of texts, the difference may be more balanced. A more complete study of frequencies should be done. If the statistical study gives a more balanced situation there would be reasons to modify the rule (the new rule is easier and an annotation symbol could be removed) and reasons to keep it as it is (the rules for natural and plain position match the implicit or explicit rules we use about the English language).

A possible compromise would be to reconsider the situation of the groups \#l and \#r as single consonants. It is sometimes difficult to realise that 
words such as `p\pln{u}blic\stst{i}ze' and `D\pln{e}clara\sno{ti}on o\vo{f}' require annotations to make the plain sound.

\subsection{The Pronunciation of `s' after a Vowel}

The rule for `s' understands this letter as one which oscillates between [s] and [z]. For the [z] sound we have `z', and for the [s] sound we have `ce', `ci', `ss'. So `s' is considered a transition consonant having both sounds depending on the context, including other words nearby (such as `This is bad' or `This is terrible'). Our rules try to accommodate the voiced sound between vowels in common words (`these', `easy', `cause', `music', etc) and also at the end of the word following a vowel or a voiced consonant (`as', `is', and many plurals). It is also common in combinations `s' + voiced group (`Oslo', `\nat{a}\no{th}\iot{e}ism', etc.).

However, these rules have many exceptions. For instance, it is typically voiceless after a vowel in these endings (-ous, -itis, -isis, -us, as well as very common words such as `this' and `us'), many common words make it voiceless between vowels (`hou\no{s}e', `cea\no{s}e', `ba\no{s}i\no{s}') and most technical words (`pa\co{r}a\no{s}\stst{i}te').

A further sophistication would be to make it voiced between vowels (or at the end of the segment after a vowel) when the vowel before the `s' has the stress. So it would work well for `these', `easy', `cause', `music', but also for endings such as `-ous', `-itis', partially `-isis', -us. But we would require to fix the plurals (and they are not always after e, such as lamas, minis, gurus).

Summing up, we are closer to consider `s' voiceless for every situation than making the rule more complex.

\subsection{Annotating `r' or the Vowel to Distinguish Rhotic from Non-Rhotic}

We have decided to annotate the `r' to make it non-rhotic using the symbol \co{r}. Another option we considered at an early stage was to make it non-rhotic when the vowel was annotated. For instance, `lorry' would become `l\pln{o}rry', and `sy\co{r}up' would become `s\pln{y}rup'. However, there are problems to distinguish between firing and pirate (the first is rhotic and the second is not). Annotating `p\nat{\i}rate' and not `firing' would be an obscure way to transmit the idea. So we finally decided to consider the \co{r}. Another issue is to consider it equivalent to a double `rr', which forces us to write `p\nat{\i}\co{r}ate', but avoids the \pln{y} in `sy\co{r}up', which is, by far, more common. So we finally saw that it was clear and economic in terms of annotations.

\subsection{Use of Introduction and Replacement Symbols for `y' and `w'}

A possibility is to eliminate the annotations `\y{}' and `\w{}'.

The first can be easily eliminated since we have already introduced the annotations for `cell\udp{u}lar' and `fail\idp{u}re', instead of `cell\y{\opq{u}}lar' and `fail\y{}ure'. We could do similarly with the following: mill\y{i}on $\rightarrow$ mill\iidp{\i{o}}n.
The reason why we have not implemented this is because we already have the symbol and the sound (e.g. torti\y{ll}a), so it does not solve anything (it is merely an aesthetic thing). However, the same reason could be used to eliminate the diphthong annotation  for `cell\udp{u}lar' and `fail\idp{u}re'.

For `\w{}', it is more difficult to eliminate this annotation completely, because we have words such as peng\w{u}in and \w{}\clr{o}ne. The sound of \w{}\clr{o}ne is closer to the sound of f\clr{o}ie, but not identical. This suggests a possible switch, and to annotate \idp{o}ne (\idp{o} = \textipa{[w2]}) and f\si{o}\udp{\i}e (\udp{i} = \textipa{[wA:]}). Nonetheless, we think these combinations would be more difficult to learn and are less intuitive (`\idp{o}ne' is simpler than `\w{}\clr{o}ne', but it hides that we have the same sound as in `m\clr{o}ney', and also the strange consonantic sound `w'.

\subsection{One (or no) Diphthong Symbol}

We have two special symbols for diphthongs: `\idp{}' and `\udp{}'. We initially considered none of them, and used `\clr{}' and `\opq{}' to cover the exception sounds, but it was too confusing to have `l\clr{o}ve' and `n\cclr{ow}'. 

Our first idea was to make combinations, such as in `backgr\clr{o}\opq{u}nd', `n\clr{o}\opq{w}', `t\clr{a}\opq{u}', `L\clr{a}\opq{o}s', etc.
But this is not accurate in phonologic terms.
So we considered the introduction of a new symbol for diphthongs, as follows:

\noindent Wide Diphthong. Single letter ({\tt \textbackslash{dip\{a\}}}): \dip{a}. Double letter ({\tt \textbackslash{ddip\{ae\}}}): \ddip{ae}. Before {\tt i}, it goes like this: ({\tt \textbackslash{dip\{\textbackslash{}i\}}}): \dip{\i}.

Examples: backgr\dip{ou}nd (better annotated as back\selr{}ground), n\dip{ow}, t\dip{au}, L\dip{ao}s, f\dip{oi}e, Fr\dip{eu}d, fr\dip{au}, \dip{ey}e, D\pln{a}l\dip{\st{a}i}. 

Table \ref{tab:diphthong} shows how it works.

\begin{table}
\begin{center}
{\small
\begin{tabular}{|| c | c | c | p{5cm} | p{3cm}  ||}
\hline
\bf{Unit} & \bf{IPA} & rhotic & \bf{Examples} & \bf{Comments} \\ \hline \hline

{\em \dip{a}, \dip{a\i}, \dip{ay}, \dip{ay}, \dip{e\i}, \dip{ey}}  & \textipa{[aI]}   & \textipa{[aI@r]} & \dip{ey}e, \dip{e\i}ther, \dip{E\i}nst\dip{e\i}n, Fa\si{h}\co{r}enh\dip{e\i}t, D\pln{a}l\dip{\st{a}\i}, m\dip{ae}stro, p\dip{a}\st{\pln{e}}lla, Ha\serl{}w\st{\dip{a\i}}i & Alt: \iidp{ey}e / \si{e}y\si{e}, \iidp{e\i}ther / \si{e}\nat{\i}ther, D\pln{a}l\si{a}\st{\nat{\i}}, m\iidp{ae}stro \\\hline
{\em \dip{au}, \dip{aw}, \dip{ao}, \dip{ou}, \dip{ow}}  & \textipa{[aU]}   & \textipa{[aU@r]} & n\dip{ow}, t\dip{au}, L\dip{ao}s, northb\dip{ou}nd (although `north\serl{}bound' is preferred) & Alt: n\uudp{ow}, t\uudp{au}, L\uudp{ao}s / n\clr{o}\opq{w}, t\clr{a}\opq{u}, L\clr{a}\opq{o}s \\\hline
{\em \dip{e}, \dip{eu}, \dip{ew}}  & \textipa{[Oi]}  & \textipa{[OI@r]]} & Fr\dip{eu}dian  & Alt: Fr\oopq{eu}dian /  Fr\opq{e}\iot{u}dian\\\hline
{\em \dip{o}, \dip{o\i}, \dip{oy}}  & \textipa{[wA:]}   & \textipa{[wA:r]} & f\dip{o\i}e, p\st{\pln{a}}t\dip{o\i}\si{s}, mem\dip{o\i}r. Note:  & Alt: f\oopq{o\i}e, c\si{h}\hvo{o}\nat{\i}r  / f\hvo{o}\clr{i}e \\\hline \hline
\end{tabular}
} %small
\vspace{5mm}
\caption{Pronunciation of the diphthong annotation}
\label{tab:diphthong} 
\end{center}
\end{table}

However, this was against the rule that the annotation for a digraph should be equivalent to the annotation of the first vowel + the elimination of the second, and we did not have that  Fr\ddip{eu}dian =  Fr\dip{e}\si{u}dian if we want \ddip{ey}e = \dip{e}\si{y}e.
So we finally decided to introduce two diphthong symbols.

\subsection{The wa-, qua-, wo-  Rule}

The sound for `want', `was', `qua', etc., is so common that it may suggest a rule.
But the variety of sounds (water, wane, whale, wall, want, war, ware, wallet, wander) makes it somehow impractical.
In any case it should be merged with the qua- rule (quantity, quarrel, but quark, quarter, quake), so considering the `wa' combination.

The rule would be (as a two-vowel unit), such as this:

\begin{itemize}
\item  ``wa in plain position" $\rightarrow$ \textipa{[w6]}  : want, was, what, watch, wad, wallet, wander, wallaby, warrior, quality, quantity, quality, quarrel, squatter  (exception: wh{a}ck, w\rnd{a}ter, wall, wag, wax, waggon)
\item ``wa in natural position (or with other vowels)" $\rightarrow$ \textipa{[weI]}  : wane, whale, ware, waist  (exception: qu\opq{a}drant)
\item ``wa in plain position (rhotic)" $\rightarrow$ \textipa{[wO:r]}  : war, quarter, warn, warm.
\end{itemize}

This could also include the wo- rules, which typical sounds opaque: woman, womb, wolf, wound, would, ... , an in rhotic position with word, work, worse, worm, with many exceptions as well (women, sword, worry, won, wonder)

We discarded this because we wanted to make the natural and plain positions limited for the natural and plain sounds. Otherwise, it would have been much more difficult.

\subsection{The -al- / -all, -ol / -oll Rule}

The sounds for `a' and `o' before `l' are typically affected in a systematic way.
Again, a variety of sounds makes it impractical: all, wall, call, walk, talk, wallet, alter, ally, algae, shall. 
In any case, the rule could also include the -ol- / -oll rule (control, roll, cold, bold, fold, sold, gold, told, poll) which are frequently natural and not plain), but (doll, follow, holly, alcohol, ...).

\subsection{The -ing- Rule}

It would be useful in many cases to consider -ing- a block, to avoid separations in `go\sel{}ing', `be\sel{}ing', `age\sel{}ing', etc., which do not take place for `sk\brd{\i}ing', `d\rnd{o}ing'.

But we would have some exceptions: Laing, boing (both with a diphthong before then ng sound), vainglorious (vain, glorious), and we would have problems with `fatiguing' (fatigu, ing), or `bilingual' (bil,ing,ual), depending on the order of the rules. Perhaps, we would also have problems with `ing\st{e}nious' and similar words.
It would also create confusion with `point', `joint'.
Leaving it explicit as we have done helps to avoid a typical confusion for foreign people to read `going' as `goi\sel{}ng'.

Furthermore, another reason to dismiss this rule is that  we did not like to consider trigraphs in our system.

\subsection{The gu- Rule}

An alternative would be to consider the `u' (after `g') silent in any case. It would go well for `g\si{u}est', `g\si{u}ilt', `vogue', `guard' and `g\si{u}arantee'.
Another option would be to consider the u as w for every case. It would help for `penguin', `distinguish', etc.

Given the frequencies, the similarity with the -ce, -ci and -ge, -gi rules, we finally decided to distinguish `gue' and `gui' from the rest.

\subsection{Stress and Separators}

Initially, there were no rules for stress. The relation between secondary stress and primary stress in the way that a secondary stress is assuming two vowel units on the left can, of course, be reconsidered. We only consider this on the left, but not on the right. The reason is that this happens (almost always) on the left, but on right it only happens sometimes (and it depends on the dialect, as the word `territory').

The connection with the sounds `\sno{}', `\svo{}', `\hno{}' and `\hvo{}' is certainly innovative, but it is almost always true. 

There are other possible rules, such as that the primary stress is always on the second vowel unit from the right if the word ends with a `c', such as `automatic'. But we should consider other suffixes such as `ical', `isis', etc.

Separators are a delicate thing. Separators affecting stress imply many things at the same time. They can be handy, but they can also be difficult to understand.

\subsection{Reduced Vowels}

Any sound can be unstressed using the appropriate annotation. For a single vowel, the most common unstressed sounds are \textipa{[@]} for a, e, o and u, \textipa{[oU]} for o, \textipa{[i]} for e (\iot{e}) and i, and \textipa{[U]/[jU]/[j@]} for u. Many of these sounds can be annotated.

The problem appears when the same letter may have different sounds depending on the reader's speed or emphasis. For instance, emission and omission can sound differently for many dialects if pronounced carefully, but they become indistinct in some dialects if read quickly. There are different situations, depending on the speed and the dialect.

lent{\em i}l   \textipa{[i]/[@]}

{\em o}mission   \textipa{[oU]/[@]}

beautif{\em u}l  \textipa{[U]/[@]}

cell{\em u}lar  \textipa{[jU]/[j@]}

{\em e}mission   \textipa{[i]/[@]}

amm{\em u}nition  \textipa{[jU]/[j@]}

fail{\em u}re   \textipa{[j@]}

fig{\em u}re   \textipa{[j@]/[@]}

Merc{\em u}ry  \textipa{[jU]/[j@]}

Chocol{\em a}te     \textipa{[i]/[@]}

ten{\em u}re     \textipa{[j@]/[jU@r]}

tel{\em e}phone   \textipa{[i]/[@]}

fut{\em i}le      \textipa{[aI]/[@]}

fort{\em u}ne      \textipa{[@]}

\noindent In some cases, this is related with the word being rhotic or not, and also to the cases in section \ref{tion}.
At the bottom of table \ref{tab:ipa2}, we can see that even the IPA has some symbols for these situations (\textipa{[j@]}, \textipa{[8]}, \textipa{[0]}, \textipa{[1]}), because they do not only oscillate but they can even be a different phoneme.

A possibility would be to include symbols that could account for these variants, so annotating a word in such a way that it would be valid for different pronunciations of the word and even for different dialects. A proposal we considered is:

\noindent Reduced. Single letter ({\tt \textbackslash{red\{a\}}}): \red{a}. Double letter ({\tt \textbackslash{rred\{ae\}}}): \rred{ae} is not used.

Examples: \red{e}m\st{i}s\sno{si}on, \red{o}m\st{i}s\sno{si}on, curr\st{\pln{\i}}c\red{u}lum, amm\red{u}n\st{\pln{\i}}\sno{ti}on

However, this is somehow against the philosophy of producing one pronunciation given an annotated word.
Consequently, in cases where the difference is caused by speed, we choose the pronunciation when the word is read slowly and carefully. In cases where the difference is originated by dialect, we have to choose one of them.

\subsection{Unstressed di-gi/ji-ci/sci/ssi-si-sti-ti-xi-zi Followed by Vowel and du-ju-ssu-su-tu-xu-zu }\label{tion}

There is an implicitly (or explicitly) well-known rule in English about the pronunciation of words such as passion, vision, nation, pressure, leisure, nature, etc. We have of course analysed a possible rule form them. Table \ref{tab:passion} summarise the possibilities.
Table \ref{tab:passion} is restricted to unstressed combinations. For stressed cases, the sound is not altered, as in science, june  (but sure and sugar). But in some dialects, dune sounds \textipa{[dZ]}, tune sounds (\textipa{[tS]}. 

In the table we also show that some cases would be excluded by previous rules (denoted by *), and some cases have several pronunciations (denoted by +).

\begin{table}
\begin{center}
{\small
\begin{tabular}{|| p{3cm} | c | p{3cm}  | p{2.7cm} | p{1.8cm} ||}
\hline
\bf{Combination}   & \bf{IPA}            & \bf{Examples}                  & \bf{Exceptions}                      & \bf{Excluded} \\ \hline \hline
di + a/e/o         & \textipa{[dZ@]}     & soldier                        & media, insidious, radial, mediate, accordion & sodium \\ \hline 
gi/ji + a/e/o      & \textipa{[dZ@]}     & religion, Belgian, contagious  & vestigial, collegiate                & orgies* \\ \hline 
ci/sci/ssi + a/e/o & \textipa{[S@]}      & conscience, passion, precious  &                                      & calcium, cesium, fancies* \\ \hline
consonant + si + a/e/o & \textipa{[S@]}  & tension, excursion, pension    &                                      & Celsius+, pansies* \\ \hline
vowel + si + a/e/o & \textipa{[Z@]}      & Asian, euthanasia, vision, illusion      & cosiest, amnesia+, Cartesian+      & gimnasium \\ \hline
sti + a/e/o        & \textipa{[stS@]}    & question+, Christian & celestial+                           & \\ \hline 
ti + a/e/o          & \textipa{[S@]}     & nation, inertia, cautious,  substantial, dictionary, abortion, patience, initial & Equation (\textipa{[Z@]}), initiate (the ver is \textipa{[SI@]} and the noun is \textipa{[SII]}, but it is never \textipa{[S@]}), prettiest, mighties  & pentium, beauties* \\ \hline 
xi + a/e/o         & \textipa{[kS@]}     & crucifixion, complexion, anxious & axiom, anorexia, asphyxiate                  &  \\ \hline 
zi + a/e/o         & \textipa{[Z@]}      &                                   & Abkhazian, trapezial            &  \\ \hline 
du + a/e/o/r@/rable/ring         & \textipa{[dZ@]/[dZU@]}& procedure, graduate, graduation            & arduous+, procedural+, gradual+, verdure+        & residuum \\ \hline 
ju + a/e/o/r@/rable/ring         & \textipa{[dZ@]/[dZU@]}& injure, perjure                         &         &  \\ \hline 
ssu + a/e/o/r@/rable/ring         & \textipa{[dZ@]/[dZU@]}& tissue+, issue+, pressure     &         &  \\ \hline 
consonant + su + a/e/o/r@/rable/ring  & \textipa{[S@]/[SU@]}& censure     & sensual, commensurate         &  \\ \hline 
vowel + su + a/e/o/r@/rable/ring  & \textipa{[Z@]/[ZU@]} & casual, usual, measurable, measuring, leisure      &          &  \\ \hline 
stu + a/e/o/r@/rable/ring  & \textipa{[stS@]/[stSU@]} & gesture      &          & perturbation \\ \hline 
tu + a/e/o/r@/rable/ring  & \textipa{[tS@]/[tSU@]} & nature, century, mutual, cultural, Portuguese, eventual+, tortuous, saturation+, saturated+, spatula+      &  fatuous+, situation+      &  \\ \hline 
xu + a/e/o/r@/rable/ring  & \textipa{[kS@]/[kSU@]} & luxury       &  sexual+        &  \\ \hline 
zu + a/e/o/r@/rable/ring  & \textipa{[Z@]/[ZU@]} & seizure       &          &  \\ \hline 
\hline
\end{tabular}
} % small
\vspace{5mm}
\caption{Unstressed di/gi-ji/ci-sci-ssi/si/sti/ti/xi/zi  and du/ju/ssu/su/tu/xu/zu}
\label{tab:passion} 
\end{center}
\end{table}

We also have cases where -ge-, -ce-, -te- sounds -ci-, -ti-, and we have positive cases such as dungeon, surgeon, ocean, crustacean, violaceous, but also negative cases such ossean, Piscean, Odysseus, and other cases just because -ea- is taken as a cluster. 

Apart from all these exceptions and special cases (and the length of the table to describe all the combinations), we would need annotations to say that the effect does not take place. This means that we would require annotations for the normal t, for the normal d, for the normal x, for the normal z. This is the main reason why we did not adopted this. Because it could create cases where we would not be able to find an annotation to undo the rule (perhaps the separation symbols, but this would be very awkward, like in `gimn\st{\nat{a}}\vo{s}\sel{}ium' or might\sel{}iest.

Related to that we have to mention the endings -lure, -nion, -nious, -lia, -nia, etc. In some cases they require a sound \textipa{[j]} (as in failure, million and onion), and in other cases they do not.

\subsection{Diacritic Use of Annotations for Homographs (Homophone and Not-homophone)}

% http://en.wikipedia.org/wiki/List_of_English_homographs
There is a possibility to annotate a word which would not require an annotation in order to distinguish it from its homograph.
In the cases where the homographs are not homophone, it is clear that both words will have a different annotation (such as wind and w\nat{\i}nd, or w\oopq{ou}nd and w\uudp{ou}nd). In the first case, we can get confused if we get used to annotated text and we eventually read non-annotated text. Any appearance of the word `wind' will be assimilated with the first variant, even though in non-annotated English it could be both of them.
We think this is slight problem since the reader must realise the context. The problem can be more serious if we just read some words (a slogan, a sign, etc.). If a few words have no annotations, it might be an annotated text, but this possibility is quite unlikely. So we will not care about this.

Another different situation is when two different words are homographs and homophones. In this case, the annotation would be the same. For instance, we have `run' for the infinitive and for the past participle. Since none of them have an annotation, we could write `r\pln{u}n' for the participle to distinguish it from the infinitive. This technique is used in other languages, such as Spanish, where it is called a `diacritical accent'. It is an interesting idea, but we will not develop it at this moment of the proposal.

\section{Appendix. Some \LaTeX{} Stuff. Packages and Symbol Construction}

This appendix is intended to people who wants to use \LaTeX{}. As it is possible that some annotators would produce \LaTeX{} text which could be compiled into PDFs using \LaTeX{}, this may be useful to people who are not regular users of \LaTeX{}.

The good thing about using \LaTeX{} or any other command-based or mark-up language is that we can define commands or labels for several annotations and then modify their graphical representation. For instance the annotated word `show\se{}r\oopq{oo}m' is coded in \LaTeX{} as follows:

\begin{verbatim}
show\se{}r\oopq{oo}m
\end{verbatim}

\noindent This can be easily converted into XML as:

\begin{verbatim}
show<se/>r<oopq>oo<oopq/>m
\end{verbatim}

\noindent And represented graphically with a proper style sheet.

\subsection{Packages used and new symbols we created}

This section describes the packages we need in \LaTeX{}. Here they are:

\begin{verbatim}
\usepackage{amssymb}   % many mathematical symbols
\usepackage{amsmath}   % many mathematical symbols
\usepackage{yhmath}    % \widetriangle
\usepackage{tipa}      % many
\usepackage{phonetic}  % \textipa and related symbols
\usepackage{extraipa}  % \subdoublevert
\usepackage{mathabx}   % \widecheck
\usepackage{stmaryrd}  % \Ydown
\usepackage{harpoon}   % \overleftharpdown, \overleftharp
\usepackage{mathtools} % \overbracket
\usepackage{accents}   % fabrication of new accents (\accentset, \underaccent)
\end{verbatim}

\noindent When using multiple packages we may encounter incompatibilities because a package tries to redefine a command, which has been already defined.
For instance, when using the LNCS documentclass, we may get a warning when loading package `amsmath':

\begin{verbatim}
Unable to redefine math accent \vec
\end{verbatim}

\noindent This can trigger further errors on other packages, such as in accents.sty, with something like ``Argument of \textbackslash{vec} has and extra }", because the right redefinition did not take place.
One way of solving this is surrounding the `documentclass' declaration as follows:

\begin{verbatim}
\let\accentvec\vec
\documentclass[fleqn]{llncs}
\let\spvec\vec
\let\vec\accentvec
\end{verbatim}

\subsection{Manual Installation of Packages}

Package ``extraipa'' is not standard, so it has to be installed manually.
The mathabx fonts can be found here:

\begin{verbatim}
http://www.ctan.org/tex-archive/fonts/mathabx/
\end{verbatim}

\noindent We have to downloaded mathabx.zip from there and unzip it.
Inside that document there are:

\begin{itemize}
\item 54 files in a folder named `source'; all 54 of those files had the extension .mf; 
\item 4 files in a folder named `texinputs': mathabx.dcl, mathabx.sty, mathabx.tex, testmac.tex
\item Other 4 files outside these folders. They are examples. You can ignore them.
\end{itemize}

\noindent Then, the installation depends on the \LaTeX{} system you are using.
Here I will describe that for MIKTEX for Windows, but it is similar for other distributions.

Go to where your miktex distribution is located, and then look for the fonts' source folder.
Typically something like this:

\begin{verbatim}
c:\Program Files\MikTeX 2.7\fonts\source\public
\end{verbatim}

\noindent Create a new folder called `mathabx' and put all 54 of the .mf files in there.

Then go to to the folder

\begin{verbatim}
c:\Program Files\MikTeX 2.7\tex\generic\misc
\end{verbatim}

\noindent Create a new folder called `mathabx' and put the other four files (mathabx.dcl, mathabx.sty, mathabx.tex, testmac.tex) in there.

Then, on the Windows' `start' go  to Miktex2.7 $\rightarrow$ Settings.
On the window that comes up, click the buttons ``Refresh FNDB" and then ``Update Formats". 

That's it. You should compile fine with this.

\subsection{Creating Commands}

The basic commands in text mode are created as follows. For instance, the separator is just created as:

\begin{verbatim}
\newcommand{\se}[1]{\textraising{#1}}
\end{verbatim}

\noindent We can join two symbols into one, just putting them together. For instance, the separator ``\selr" is created as follows:

\begin{verbatim}
\newcommand{\selr}[1]{\sel{}\textsubplus{}}
\end{verbatim}

\noindent Some accents needed to use the math environment to cover two letters.
For instance, this is how we constructed a $tilde$ symbol covering two letters:

\begin{verbatim}
\newcommand{\nnat}[1]{$\widetilde{\mbox{#1}}$}
\end{verbatim}

\noindent For the $breve$ we (in the $mathabx$ package) we did similarly.
\begin{verbatim}
\newcommand{\oopq}[1]{$\widecheck{\mbox{#1}}$}
\end{verbatim}

\noindent For the cross beneath the letter, we managed to create a new symbol called {\tt \textbackslash{textsubcross}} from the package $tipa$.

\begin{verbatim}
 \newcommand{\textsubcross}[1]{\tipaloweraccent{24}{#1}}
\end{verbatim}

We used it to define the {\tt \textbackslash{si}} command.

\begin{verbatim}  
\newcommand{\si}[1]{\textsubcross{#1}}
\end{verbatim}

\subsection{List of Command Definition}

Copying the package loading instructions and the following command definitions, we can compile any annotated English text with \LaTeX{} notation.

\begin{verbatim}

\newcommand{\textsubcross}[1]{\tipaloweraccent{24}{#1}}


% SHORTCUTS

% Annotated
\newcommand{\annl}[1]{\textopencorner{#1}}   %{$\llcorner{}{\mbox{#1}}$}
\newcommand{\annr}[1]{\textcorner{#1}}   %{$\lrcorner{}{\mbox{#1}}$}

% Non-annotated
\newcommand{\nonl}[1]{$\llcorner{}{\mbox{#1}}$}   %{\textopencorner{#1}}
\newcommand{\nonr}[1]{$\lrcorner{}{\mbox{#1}}$}   %{\textcorner{#1}}


% Silent, stress and separators (beneath)  
% Hint: all start with 's'
\newcommand{\si}[1]{\textsubcross{#1}}
\newcommand{\st}[1]{\textsyllabic{#1}}
\newcommand{\stst}[1]{\subdoublevert{#1}}
\newcommand{\se}[1]{\textraising{#1}}
\newcommand{\sel}[1]{\textadvancing{#1}}
\newcommand{\ser}[1]{\textretracting{#1}}
\newcommand{\selr}[1]{\sel{}\textsubplus{}}
\newcommand{\serl}[1]{\textsubplus{}\ser{}}


% Consonants (beneath)
% Hint: all end with 'o'
\newcommand{\co}[1]{\textsubring{#1}}
\newcommand{\io}[1]{\d{#1}}
\newcommand{\vo}[1]{\textsubwedge{#1}}
\newcommand{\no}[1]{\textsubcircum{#1}}

\newcommand{\svo}[1]{\textinvsubbridge{#1}}
\newcommand{\sno}[1]{\textsubbridge{#1}}

\newcommand{\hvo}[1]{\b{\textinvsubbridge{#1}}}
\newcommand{\hno}[1]{\b{\textsubbridge{#1}}}


% Semiconsonant (w)
\newcommand{\w}[1]{\textsubw{#1}}
% Semiconsonant (y)
\newcommand{\y}[1]{$\underaccent{\Ydown}{\mbox{#1}}$}


% Vowel Introductions
\newcommand{\sch}[1]{$\accentset{\backepsilon}{\mbox{#1\upbar{}}}$}  % package textcomp
\newcommand{\ssch}[1]{$\accentset{\backepsilon}{\mbox{#1}}$}  % package textcomp


% Vowel Annotations
% Hint: all with three letters
\newcommand{\pln}[1]{\textvbaraccent{#1}}
\newcommand{\ppln}[1]{\textvbaraccent{#1}}

\newcommand{\nat}[1]{\~{#1}}
\newcommand{\nnat}[1]{$\widetilde{\mbox{#1}}$}

\newcommand{\brd}[1]{\={#1}}
\newcommand{\bbrd}[1]{$\overline{\mbox{#1}}$}

\newcommand{\rnd}[1]{\r{#1}}
\newcommand{\rrnd}[1]{\r{#1}}

\newcommand{\clr}[1]{\^{#1}}
\newcommand{\cclr}[1]{$\widehat{\mbox{#1}}$}

\newcommand{\opq}[1]{\v{#1}}
\newcommand{\oopq}[1]{$\widecheck{\mbox{#1}}$}

\newcommand{\dip}[1]{$\widetriangle{\mbox{#1}}$}  % Problems
\newcommand{\ddip}[1]{$\widetriangle{\mbox{#1}}$}

\newcommand{\idp}[1]{\`{#1}}
\newcommand{\iidp}[1]{$\overleftharpdown{\mbox{#1}}$}

\newcommand{\udp}[1]{\'{#1}}
\newcommand{\uudp}[1]{$\overleftharp{\mbox{#1}}$}

\newcommand{\iot}[1]{\.{#1}}

\newcommand{\cnt}[1]{\"{#1}}
\newcommand{\ccnt}[1]{\"{#1}}

\newcommand{\red}[1]{\crtilde{#1}}
\newcommand{\rred}[1]{\crtilde{#1}}

% Grouping Command
\newcommand{\group}[1]{\underline{#1}}

\end{verbatim}

{\small 
\bibliographystyle{plain}

\bibliography{biblio}

}

\end{document}